%% file: icml_main.tex
\documentclass{article}

\usepackage{microtype}
\usepackage{graphicx}
\usepackage{caption}
\usepackage{booktabs} % for professional tables

\usepackage{hyperref}

\usepackage[accepted]{icml2024}

\usepackage{amsmath}
\usepackage{amssymb}
\usepackage{mathtools}
\usepackage{amsthm}

\usepackage[capitalize,noabbrev]{cleveref}

\theoremstyle{plain}

\theoremstyle{definition}

\theoremstyle{remark}

\usepackage[textsize=tiny]{todonotes}

\usepackage{wrapfig}
\usepackage{listings}
\usepackage{comment}
\usepackage{adjustbox}
\usepackage{xspace}
\usepackage{fancyvrb}
\usepackage[subtle]{savetrees}
\usepackage{soul}
\usepackage{titletoc}

\newcommand{\ours}{WAVES\xspace}

\input{include/000_premeable.tex}

\icmltitlerunning{WAVES: Benchmarking the Robustness of Image Watermarks}

\begin{document}

\twocolumn[
\icmltitle{WAVES: Benchmarking the Robustness of Image Watermarks}



\icmlsetsymbol{equal}{*}

\begin{icmlauthorlist}
\icmlauthor{Bang An}{equal,yyy}
\icmlauthor{Mucong Ding}{equal,yyy}
\icmlauthor{Tahseen Rabbani}{equal,yyy}
\icmlauthor{Aakriti Agrawal}{yyy}
\icmlauthor{Yuancheng Xu}{yyy}
\icmlauthor{Chenghao Deng}{yyy}
\icmlauthor{Sicheng Zhu}{yyy}
\icmlauthor{Abdirisak Mohamed}{yyy,comp}
\icmlauthor{Yuxin Wen}{yyy}
\icmlauthor{Tom Goldstein}{yyy}
\icmlauthor{Furong Huang}{yyy}
\end{icmlauthorlist}

\icmlaffiliation{yyy}{University of Maryland, College Park}
\icmlaffiliation{comp}{SAP Labs, LLC}

\icmlcorrespondingauthor{Bang Ang}{bangan@umd.edu}
\icmlcorrespondingauthor{Mucong Ding}{mcding@umd.edu}
\icmlcorrespondingauthor{Tahseen Rabbani}{trabbani@umd.edu}

\icmlkeywords{Machine Learning, ICML}

\vskip 0.3in
]



\printAffiliationsAndNotice{\icmlEqualContribution}  % leave blank if no need to mention equal contribution

\begin{abstract}
\input{icml_sec/0_abstract}   
\end{abstract}

\input{icml_sec/1_intro}

\input{icml_sec/2_background}
\input{icml_sec/3_standardizedEval}

\input{icml_sec/4_attacks}

\input{icml_sec/5_results}

\input{icml_sec/7_impact}

% \section*{Software and Data}

% If a paper is accepted, we strongly encourage the publication of software and data with the
% camera-ready version of the paper whenever appropriate. This can be
% done by including a URL in the camera-ready copy. However, \textbf{do not}
% include URLs that reveal your institution or identity in your
% submission for review. Instead, provide an anonymous URL or upload
% the material as ``Supplementary Material'' into the OpenReview reviewing
% system. Note that reviewers are not required to look at this material
% when writing their review.

%\section*{Acknowledgements}

% \textbf{Do not} include acknowledgements in the initial version of
% the paper submitted for blind review.

% If a paper is accepted, the final camera-ready version can (and
% probably should) include acknowledgements. In this case, please
% place such acknowledgements in an unnumbered section at the
% end of the paper. Typically, this will include thanks to reviewers
% who gave useful comments, to colleagues who contributed to the ideas,
% and to funding agencies and corporate sponsors that provided financial


\bibliography{icml_main}
\bibliographystyle{icml2024}

%%%%%%%%%%%%%%%%%%%%%%%%%%%%%%%%%%%%%%%%%%%%%%%%%%%%%%%%%%%%%%%%%%%%%%%%%%%%%%%
%%%%%%%%%%%%%%%%%%%%%%%%%%%%%%%%%%%%%%%%%%%%%%%%%%%%%%%%%%%%%%%%%%%%%%%%%%%%%%%
% APPENDIX
%%%%%%%%%%%%%%%%%%%%%%%%%%%%%%%%%%%%%%%%%%%%%%%%%%%%%%%%%%%%%%%%%%%%%%%%%%%%%%%
%%%%%%%%%%%%%%%%%%%%%%%%%%%%%%%%%%%%%%%%%%%%%%%%%%%%%%%%%%%%%%%%%%%%%%%%%%%%%%%
\clearpage
\onecolumn
{\centering
    \Large
    \textbf{WAVES: Benchmarking the Robustness of Image Watermarks}\\
    \vspace{0.5em}Supplementary Material \\
    \vspace{1.0em}
}

% appendix ToC
\startcontents[appendices]
\setcounter{tocdepth}{2}
\printcontents[appendices]{l}{1}{}
% \tableofcontents

\clearpage
\appendix
\input{icml_sec/X_suppl}
%%%%%%%%%%%%%%%%%%%%%%%%%%%%%%%%%%%%%%%%%%%%%%%%%%%%%%%%%%%%%%%%%%%%%%%%%%%%%%%
%%%%%%%%%%%%%%%%%%%%%%%%%%%%%%%%%%%%%%%%%%%%%%%%%%%%%%%%%%%%%%%%%%%%%%%%%%%%%%%

\end{document}

%% file: icml_sec/0_abstract.tex
In the burgeoning age of generative AI, watermarks act as identifiers of provenance and artificial content. We present WAVES (\textbf{W}atermark \textbf{A}nalysis \textbf{v}ia \textbf{E}nhanced \textbf{S}tress-testing), a benchmark for assessing image watermark robustness, overcoming the limitations of current evaluation methods. WAVES integrates detection and identification tasks and establishes a standardized evaluation protocol comprised of a diverse range of stress tests. The attacks in WAVES range from traditional image distortions to advanced, novel variations of diffusive, and adversarial attacks. 
Our evaluation examines two pivotal dimensions: the degree of image quality degradation and the efficacy of watermark detection after attacks. 
Our novel, comprehensive evaluation reveals previously undetected vulnerabilities of several modern watermarking algorithms. We envision WAVES as a toolkit for the future development of robust watermarks. 
The project is available at \href{https://wavesbench.github.io/}{https://wavesbench.github.io/}.

%% file: icml_sec/1_intro.tex
\section{Introduction}
\label{sec:intro}
Recent and pivotal advancements in text-to-image diffusion models \citep{ho2020denoising, dhariwal2021diffusion, rombach2022high} have garnered the attention of the AI community and the general public. Open-source models such as Stable Diffusion and proprietary models such as the Dall$\cdot$E family and Midjourney 
have enabled users to produce images that are of human-produced quality. Consequently, there has been a strong push in the AI/ML community to develop reliable algorithms for detecting AI-generated content and determining its source \citep{executive2023order}. 
One avenue for maintaining the provenance of generative content is by embedding \textit{watermarks}. A watermark is a signal encoded onto an image to signify its source or ownership \citep{al2007combined, zhu2018hidden, zhang2019robust, tancik2020stegastamp, fernandez2023stable, wen2023tree}. To avoid degradation of image quality, an invisible watermark is desired. 
Many such watermarks are robust to common image manipulations \citep{lukas2023leveraging, zhao2023invisible, wen2023tree, fernandez2023stable}, and adversarial efforts to remove the watermark are complicated by the difficulty of decoding/extracting the message without private knowledge of the watermarking scheme \citep{tancik2020stegastamp, fernandez2023stable}.
Despite this difficulty, various watermark removal schemes can still be effective \citep{zhao2023invisible, saberi2023robustness}. 
However, a lack of standardized evaluations in existing literature (i.e., inconsistent image quality measures, statistical parameters, and types of attacks) has resulted in an incomplete picture of the vulnerabilities and robustness of these algorithms in the real world.

\begin{figure}[!htbp]
    \centering
    \vspace{-0.5em}
    \includegraphics[width=\linewidth]{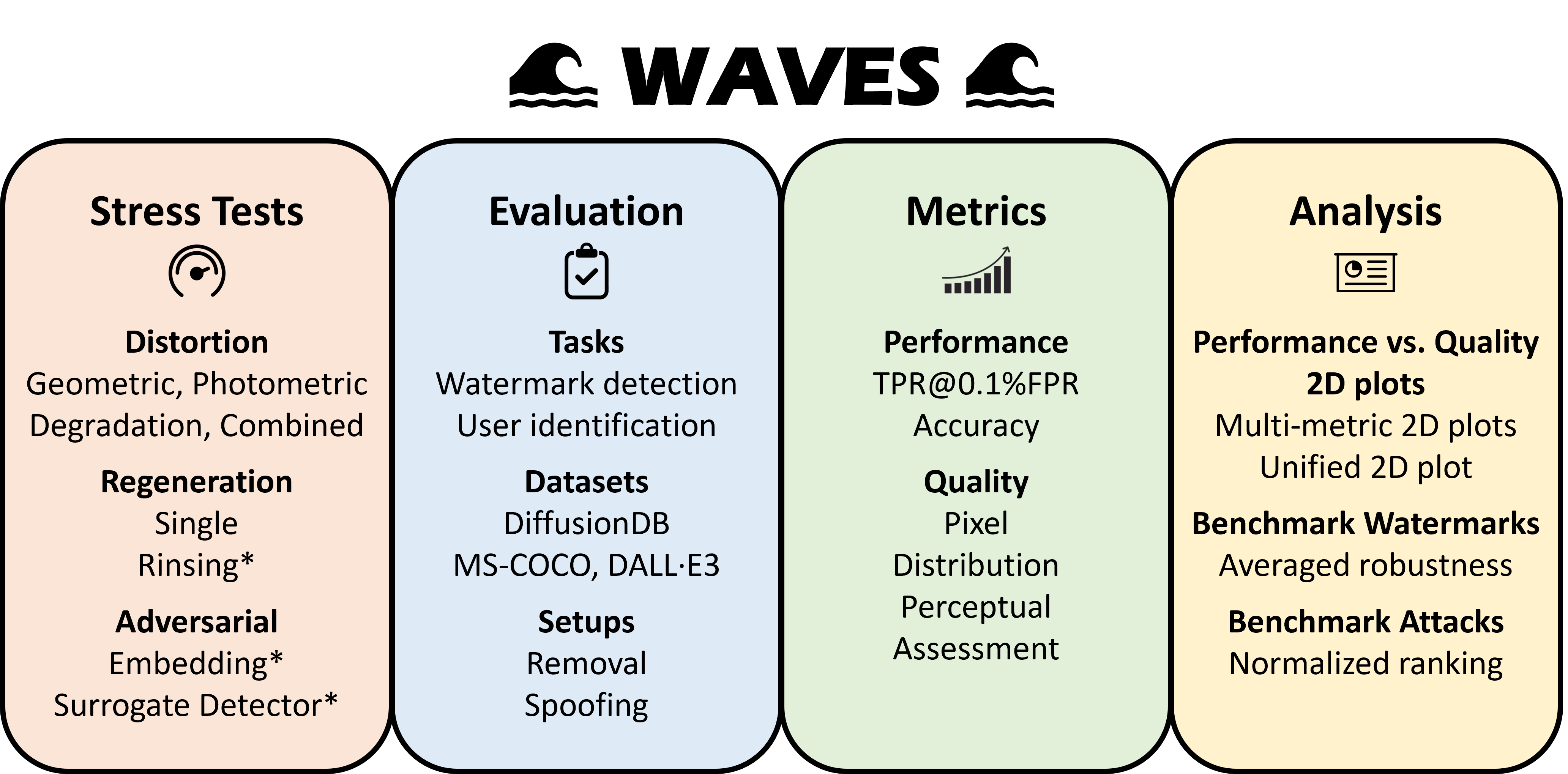}
    \vspace{-1.5em}
    \caption{\textbf{\ours} establishes a standardized evaluation framework that encompasses a comprehensive suite of stress tests including both existing and newly proposed stronger attacks (denoted by $^*$).
    }\label{fig:waves-sunbrust}
    \vspace{-0em}
\end{figure}

We present \ours (\textbf{W}atermark \textbf{A}nalysis \textbf{v}ia \textbf{E}nhanced \textbf{S}tress-testing), a benchmark for assessing watermark robustness, overcoming the limitations of current evaluation methods. \ours consists of a comprehensive variety of novel \& realistic attacks, including classical image distortions, image regeneration, and adversarial attacks. 
In an effort to stress-test existing/future watermarks, we propose several new attacks such as adversarial embedding attacks, and new variants of existing attacks such as multi-regeneration attacks. 

\input{table/eval_compare}
\ours focuses on the sensitivity and robustness of watermark detection, measured by the true positive rate (TPR) at 0.1\% false positive rate (FPR), and in the meantime, studies the severity of image degradations needed to decrease this sensitivity with multiple quality metrics. 
\ours develops a series of Performance vs. Quality 2D plots varying over several prominent image similarity metrics, which are then aggregated in a heuristically novel manner to paint an overall picture of watermark robustness and attack potency.

We extensively evaluate the security of three prominent watermarking algorithms, Stable Signature, Tree-Ring, and StegStamp, respectively representing three major techniques for embedding an invisible signature. 
\ours effectively reveals weaknesses in them and discovers previously undetected vulnerabilities. For example, watermarking algorithms using publicly available VAEs can have their watermarks effectively removed with minimal image manipulation. DALL$\cdot$E3's usage of an open-source KL-VAE underscores the need for unique VAEs in such systems. 

Our \textbf{contributions} are summarized as follows:
\begin{parenum}
\item In practical scenarios where false alarms incur high costs, our evaluation metric for watermark detection prioritizes the True Positive Rate (TPR) at a stringent False Positive Rate (FPR) threshold, specifically 0.1\%. This focus addresses the inadequacies of alternative metrics such as the $p$-value and Area Under the Receiver Operating Characteristic (AUROC).
\item Additionally, our metric incorporates image quality alongside TPR@0.1\% FPR. This integration acknowledges the necessity of maintaining a balance between reducing the accuracy of watermark detection and the practical utility of the image in practical scenarios.
\item We introduce a comprehensive taxonomy of attacks that encompasses classical distortions (blurring, rotation, cropping, etc.) and powerful, novel variations of regeneration and adversarial attacks, against watermarks. 
\item We standardize the evaluation of watermark robustness, allowing us to rank attacks and watermarks. We formalize the watermark \textit{detection} and user \textit{identification} problems and evaluate the robustness under both scenarios. 
\item Our benchmark uncovers several especially harmful attacks for popular watermarks, some of which are first introduced in this work, underscoring the need for refinement of existing watermarking algorithms and systems. 
\ours contributes as a toolkit to examine the watermark robustness and helps future development of robust watermarks.
\end{parenum}

%% file: table/eval_compare.tex
\begin{table*}[t]
\centering
\caption{\label{tab:eval_compare}\textbf{Comparison of robustness evaluations with existing works.} For \textit{categories of attacks}, D, R, and A denote distortions, image regeneration, and adversarial attacks. 
\textit{Joint test} means whether the performance and quality are jointly tested under a range of attack strengths. Our benchmark is the most comprehensive one, with a large scale of attacks, data, metrics, and more realistic evaluation setups.}
\begin{adjustbox}{max width=\textwidth}
\renewcommand{\arraystretch}{0.9}
\begin{threeparttable}
\begin{tabular}{ccccccccc}
\toprule
\multicolumn{1}{c}{Research} & \multicolumn{1}{c}{Num. of} & \multicolumn{1}{c}{Categories} & \multicolumn{1}{c}{Num. of} & \multicolumn{1}{c}{Sample Size} & \multicolumn{1}{c}{Non-watermarked} & \multicolumn{1}{c}{Performance} & \multicolumn{1}{c}{Num. of} & \multicolumn{1}{c}{Joint} \\
\multicolumn{1}{c}{Work} & \multicolumn{1}{c}{Attacks} & \multicolumn{1}{c}{of Attacks} & \multicolumn{1}{c}{Datasets} & \multicolumn{1}{c}{per Dataset} & \multicolumn{1}{c}{Image Source} & \multicolumn{1}{c}{Metric} & \multicolumn{1}{c}{Quality Metrics} & \multicolumn{1}{c}{Test} \\
\midrule
StegaStamp Watermark\tnote{1}       & 5                   & D                      &   1                 &   1000                      &  ---                      & bit accuracy                     &  3                                 &  \xmark                                 \\
Stable Signature Watermark\tnote{2} &  12                  & D, R                      &  1                  &   5000                      & ---                       &  bit accuracy                        &   3                                & \xmark                                  \\
TreeRing Watermark\tnote{3}         &   6                 &  D                     & 2                   &  1000                       &  generate by same model                     &   TPR@1\%FPR               &    2                               &  \xmark                                 \\
Regeneration Attack\tnote{4}        &  10                  &  D, R                     &  2                  & 500                        &   ---                     &   bit accuracy                        &    3                               &  \xmark                                 \\
Surrogate Model Attack\tnote{5}     &  2                  &  R, A                     &  1                  &  2500                       &   real images                    &   AUROC                        &    0                               &  \xmark                                 \\
Adaptive Attack\tnote{6}            &  10                  &  D, A                     & 1                   & 1000                        &  real images                      &  TPR@1\%FPR                         &    3                               &   \xmark                               \\
\midrule
\textbf{WAVES (ours)}               &  26                  & D, R, A                      & 3                   &   5000                      &  real images                      &  TPR@0.1\%FPR                         &     8                                   &  \cmark                    \\
\bottomrule
\end{tabular}
\begin{tablenotes}
\item[1] \citet{tancik2020stegastamp}.
\item[2] \citet{fernandez2023stable}.
\item[3] \citet{wen2023tree}.
\item[4] \citet{zhao2023invisible}.
\item[5] \citet{saberi2023robustness}.
\item[6] \citet{lukas2023leveraging}.
\end{tablenotes}
\end{threeparttable}
\end{adjustbox}
\end{table*}

%% file: icml_sec/2_background.tex
\section{Image Watermarks}\label{sec:related}
We briefly review invisible watermarks and defer detailed discussions to Appendix~\ref{app:related}.
Generally, there are two types of watermarking methods. 
\textbf{(1) Post-processing watermarks} embed watermarks after image generation.
\textit{(1a) Frequency-domain methods} like DWT, DCT~\citep{cox2007digital}, and DWTDCT~\citep{al2007combined} modify images in transform domains. 
\textit{(1b) Deep encoder-decoder methods} such as HiDDeN~\citep{zhu2018hidden}, RivaGAN~\citep{zhang2019robust}, and \underline{StegaStamp}~\citep{tancik2020stegastamp} use trained neural networks for embedding and decoding watermarks. 
Post-processing watermarks are model-agnostic but can introduce human-visible artifacts, compromising image quality.
\textbf{(2) In-processing watermarks} integrate watermarking into the image generation process, substantially eliminating visible artifacts. 
\textit{(2a) Whole model modifications} embed watermarks by training the entire generative models on watermarked images \citep{yu2021artificial,zeng2023securing, lukas2023ptw}. 
\textit{(2b) Partial model modifications} such as \underline{Stable Signature}~\citep{fernandez2023stable} only fine-tune the decoder of the latent-diffusion model.
\textit{(2c) Random seed modification} watermarks like \underline{Tree-Ring}~\citep{wen2023tree} embed watermarks into the initial noise vector of diffusion models which can be retrieved at detection time.

Robustness is an essential property of watermarks especially since there is an incentive to remove watermarks. 
Besides natural image distortions, some watermarks are shown to be vulnerable to regeneration through diffusion models or VAEs \citet{zhao2023invisible, saberi2023robustness}, and adversarial attacks \citet{lukas2023leveraging, saberi2023robustness}. 
However, some unrealistic attacks and inconsistent robustness evaluations across different studies have muddled the understanding of watermark robustness, obscuring the true vulnerabilities of these methods. 
Therefore, this paper provides a standardized and comprehensive benchmark, encompassing a set of realistic and strong attacks. Our benchmark enables apple-to-apple comparison of watermarks as well as attacks, which helps standardize and accelerate the studies of robust watermarks.

%% file: icml_sec/3_standardizedEval.tex
\begin{figure*}[!htbp]
    \hfill
      \begin{subfigure}{\textwidth}
        \centering
        \includegraphics[width=\linewidth]{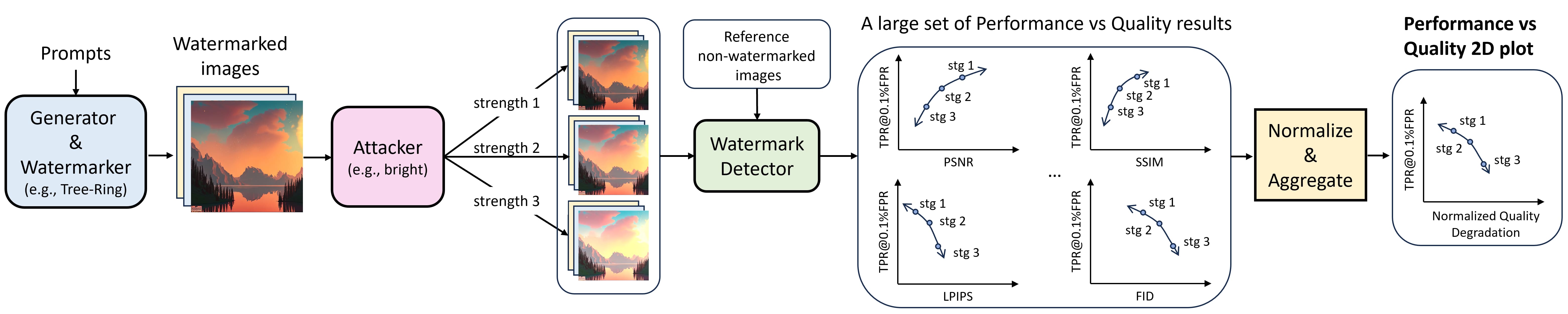}
        \caption{Evaluation of a single attack on a watermarking method. We first attack watermarked images over a variety of strengths (also labeled 'stg'). Then, we evaluate the detection performance (TPR@0.1\%FPR) and a collection of image quality metrics such as PSNR and plot a set of performance vs. quality plots. By normalizing and aggregating these quality metrics, we derive a consolidated 2D plot that represents the overall performance vs. quality for the evaluation.}
      \end{subfigure}%
    \hfill
      \begin{subfigure}{\textwidth}
        \centering
        \includegraphics[width=\linewidth]{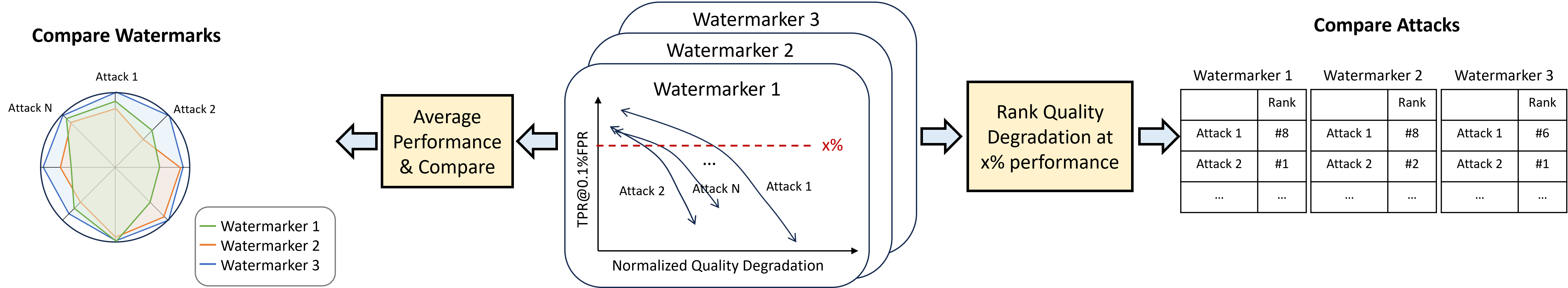}
        \caption{Benchmarking watermarks and attacks. For each watermark, we plot all attacks on a unified performance vs. quality 2D plot to facilitate a detailed comparison. Based on this, we provide two additional analytical perspectives. We compare watermarks' robustness through the averaged performance under different attacks. We evaluate attacks' potency by ranking the quality at a specific performance threshold.}
      \end{subfigure}%
      \vspace{-1em}
     \caption{Evaluation workflow. }
      \label{fig:eval-workflow}
      \vspace{-1em}
\end{figure*}

\section{Standardized Evaluation through \ours}\label{sec:eval-metric}

\begin{table*}[t]
\centering
\caption{\label{tab:summary_attacks}A taxonomy of all the attacks in our stress-testing set. Novel attacks proposed by \ours are marked with $^*$.}
\begin{adjustbox}{max width=\textwidth}
\renewcommand{\arraystretch}{1.1}
\begin{threeparttable}
\begin{tabular}{cccc}
\toprule
\textbf{Category} & \textbf{Subcategory (prefix)} & \textbf{Description} & \textbf{Attack Names (suffix)} \\
\midrule
\multirow{2}{*}{Distortion} & Single (Dist-) & Single distortion & -Rotation, -RCrop, -Erase, -Bright, -Contrast, -Blur, -Noise, -JPEG \\ \cline{2-4} 
& Combination (DistCom-) & Combination of a type of distortions & -Geo, -Photo, -Deg, -All \\
\midrule
\multirow{2}{*}{Regeneration} & Single (Regen-) & A single VAE or diffusion regeneration & -Diff, -DiffP\tnote{1}~, -VAE, -KLVAE\tnote{2} \\ \cline{2-4} 
& Rinsing$^*$ (Rinse-) & A multi-diffusion regeneration & -2xDiff, -4xDiff \\
\midrule
\multirow{3}{*}{Adversarial} & Embedding (grey-box)$^*$ (AdvEmbG-)\tnote{3} & Use the same VAE & -KLVAE8 \\ \cline{2-4} 
& Embedding (black-box)$^*$ (AdvEmbB-) & Use other encoders & -RN18, -CLIP, -KLVAE16, -SdxlVAE \\ \cline{2-4} 
& Surrogate detector attack$^*$ (AdvCLS-)\tnote{4} & Train a watermark detector & -UnWM\&WM, -Real\&WM, -WM1\&WM2 \\
\bottomrule
\end{tabular}
\begin{tablenotes}
\item[1] DiffP requires user prompts.
\item[2] KLVAE with bottleneck size 8 is grey-box.
\item[3] AdvEmbG is grey-box.
\item[4] AdvCLS needs data and training.
\end{tablenotes}
\end{threeparttable}
\end{adjustbox}
\end{table*}

\subsection{Standardized Evaluation Workflow and Metrics}\label{subsec:std_eval}
As shown in Table~\ref{tab:eval_compare}, our benchmark, \ours, stands out by considering three diverse datasets, incorporating 26 diverse attacks across three categories, and employing 8 quality metrics. These distinguish our work as the most extensive and realistic setup to date for watermark robustness evaluation. For more details on evaluation workflow, setups, metrics, and more analyses, see Appendix~\ref{app:eval_details}.

\textbf{Applications and formulation of invisible image watermarks.} Invisible image watermarks, originally for protecting creators' intellectual property, have expanded into broader applications like \textbf{AI Detection} — identifying AI-generated images~\citep{saberi2023robustness}, and \textbf{User Identification} — tracking the source of an image to its creator~\citep{fernandez2023stable}. We are interested in message-based approaches, where a unique, invisible identifier is embedded into an image. which may be recovered by the content creator at any time to establish provenance. The choice of message varies across methods, with Tree-Ring using random complex Gaussians and others like Stable Signature employing binary strings.

\textbf{Evaluation Workflow.} The trade-off between watermark performance and image quality, especially when watermark attacks lead to image distortions, is critical. We introduce \textit{Performance vs. Quality 2D plots} for a comprehensive comparison, a novel perspective over the typical performance-centric analyses. The evaluation process involves comparing watermarked images with a diverse set of real and AI-generated reference images to produce the performance vs. quality 2D plots, and processing or aggregating the 2D plots to compare attacks and watermarks, as depicted in Figure~\ref{fig:eval-workflow}.

\textbf{Performance Metrics in AI Detection and User Identification.}
\ours prioritizes fairness and comprehensiveness by using evaluation metrics that are independent of the choice of statistical tests and $p$-value thresholds, in contrast to some prior practices such as~\citep{fernandez2023stable}.
AI detection in \ours is akin to binary classification, utilizing ROC curve-based metrics.
Given the significant impact of false positives in mislabeling non-watermarked images, strict control over the false positive rate (FPR) is crucial. Therefore, rather than AUROC (since a high AUROC score does not necessarily imply a high true positive rate (TPR) at low FPR levels), \ours focuses on TPR@$x\%$FPR, specifically at a challenging low FPR threshold of $0.1\%$, extending recent studies such as \citep{wen2023tree} with a larger dataset and a more stringent FPR criterion.
User identification is approached as multi-class classification, and we measure performance by the accuracy of correct image assignments to users.

\textbf{Implementing Diverse Image Quality Metrics:} Recognizing that no single metric can fully capture the aspects of generated images, we use a range of image quality metrics and propose a normalized, aggregated metric for evaluating watermark and attack methods. \ours integrates over 8 metrics in 4 categories: \textit{(1)} \textbf{Image similarities}, including Peak Signal-to-Noise Ratio (PSNR), Structural Similarity Index (SSIM), and Normalized Mutual Information (NMI), which assess the pixel-wise accuracy after attacks; \textit{(2)} \textbf{Distribution distances} such as Frechet Inception Distance (FID)~\citep{heusel2017gans} and a variant based on CLIP feature space (CLIP-FID)~\citep{kynkaanniemi2022role}; \textit{(3)} \textbf{Perception-based metrics} like Learned Perceptual Image Patch Similarity (LPIPS) \citep{zhang2018perceptual}; \textit{(4)} \textbf{Image quality assessments} including aesthetics and artifacts scores~\citep{xu2023imagereward}, which quantify the changes in aesthetic and artifact features.

\textbf{Normalization and Aggregation of Image Quality Metrics:} Addressing the distinct characteristics of various image quality metrics, \ours proposes {\textit{a normalized and aggregated quality metric}} for a unified measure of image quality degradation and comprehensive scoring of attack or watermark methods. We define the normalized scale for each metric by assigning the 10\% quantile value over all attacked images (across 26 attack methods, three watermark methods, and three datasets) as the 0.1 point, and the 90\% quantile as the 0.9 point. \textit{Normalized quality metrics are always ranked in ascending order of image degradation.} This normalization ensures equivalent significance across different metrics, defined by their quantiles in a large set of attacked watermarked images. Normalized metrics are aggregated and extensively utilized in Section~\ref{sec:results} for Performance vs Quality plots, watermark radar plots, and attack leaderboards.

%% file: icml_sec/4_attacks.tex
\subsection{Stress-testing Watermarks}
\label{sec:attack}

We evaluate the robustness of watermarks with a wide range of attacks detailed in this section and summarized in Table~\ref{tab:summary_attacks} and Table \ref{tab:attacker_know}. Figure \ref{many-attacks} demonstrates the visual effects.

\textbf{Distortion Attacks.}
Watermarked images often face distortions such as compression and cropping during internet transmission, necessitating watermarks that can endure common alterations. 
However, most studies only test resilience against singular or extreme distortions.
In \ours, we establish the following distortions within an acceptable quality threshold as our baselines. 
\textbf{Geometric distortions}: rotation, resized-crop, and erasing;
\textbf{Photometric distortions}: adjustments in brightness and contrast;
\textbf{Degradation distortions}: Gaussian blur, Gaussian noise, and JPEG compression;
\textbf{Combo distortions}: combinations of geometric, photometric, and degradation distortions, both individually and collectively. 
Detailed setups for each are provided in the Appendix~\ref{app:distortion}.

\begin{figure}[t]
    \centering
    \includegraphics[trim={0 20px 0 0}, width=\linewidth]{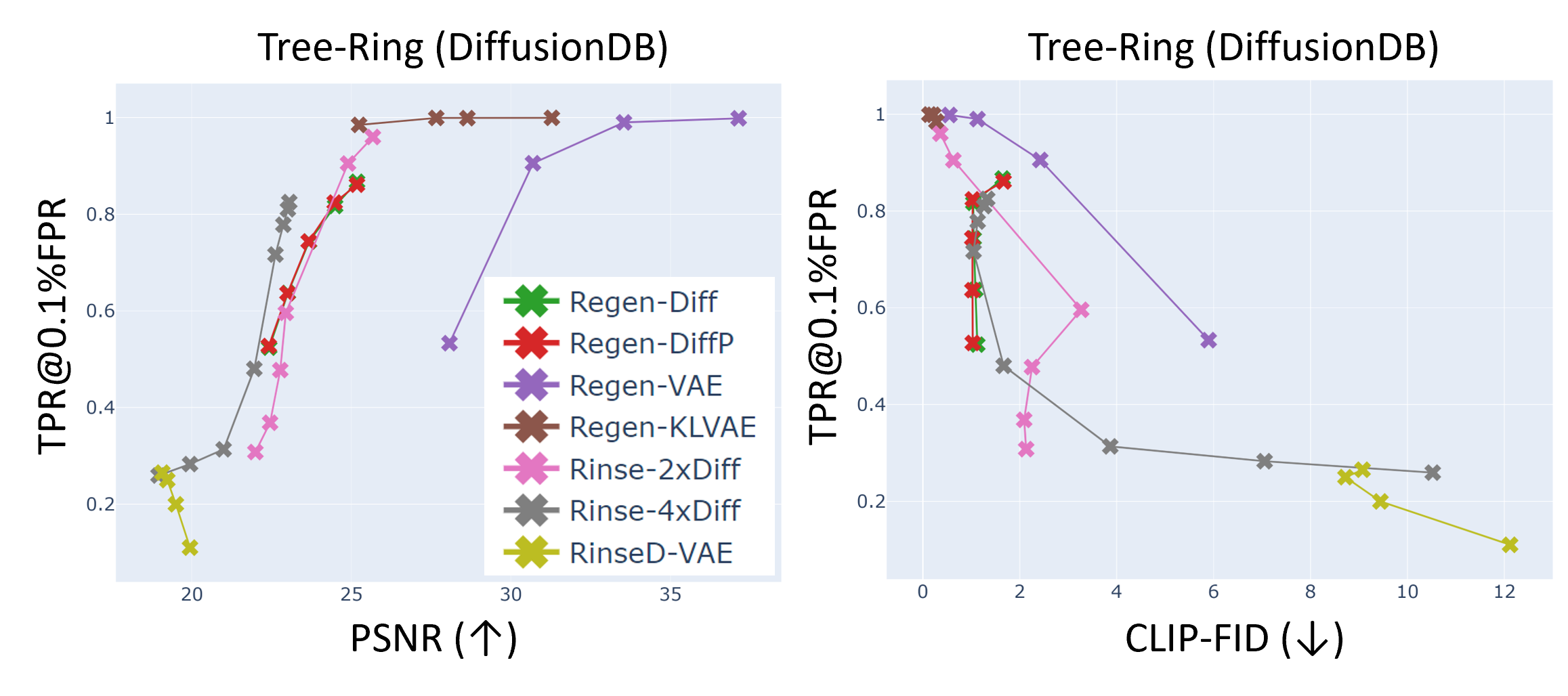}
    \vspace{-2em}
\caption{Regeneration attacks on Tree-Ringk. Regen-Diff is a single diffusive regeneration and Rinse-[N]xDiff is a rinsing one with $N$ repeated diffusions, with the number of noising steps as attack strength. Regen-VAE uses a pre-trained VAE with quality factor as strength and Regen-KLVAE uses pre-trained KL-VAEs with bottleneck size as strength. RinseD-VAE applies a VAE as a denoiser after Rinse-4xDiff.} 
    \label{multi-regen}
    \vspace{-1.3em}
\end{figure}

\begin{figure*}[!htbp]
\vspace{-0.5em}
    \centering
    \includegraphics[trim={0 20px 0 0}, width=\linewidth]{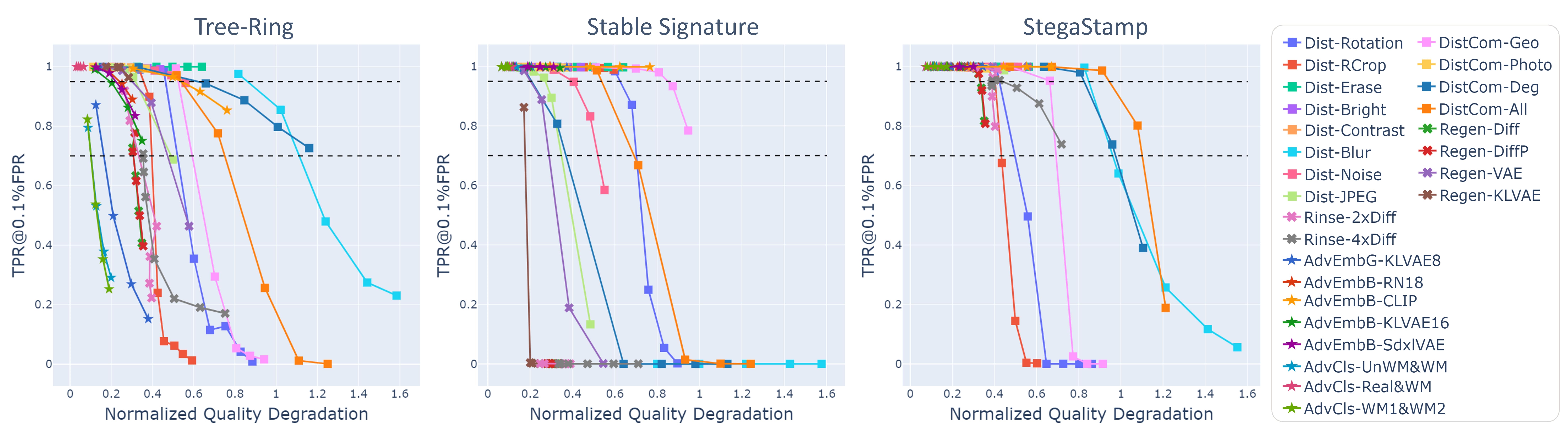}
    \vspace{-1em}
    \caption{Unified performance vs. quality degradation 2D plots under detection setup. We evaluate each watermarking method under various attacks. Two dashed lines show the thresholds used for ranking attacks. }
    \label{fig:2d_plots}
\end{figure*}

\textbf{Regeneration Attacks}, employing diffusion models or VAEs \cite{saberi2023robustness, zhao2023invisible}, aim at altering an image's latent representation by noising and then denoising an image.
Different from existing works that only perform a \textbf{Single regeneration}, we also investigate \textbf{Rinsing regenerations}, where an image undergoes multiple cycles of noising and denoising through a pre-trained diffusion model. 
Furthermore, we introduce two additional variations: prompted regeneration and mixed regeneration (rinse + VAE denoising). 
To simulate a realistic attack, we use a lower version diffusion model than the one used to generate watermarked images. All such attacks are detailed in Appendix \ref{regeneration}.
As shown in \Cref{multi-regen}, in contrast with the conclusions of \citet{zhao2023invisible}, the Tree-Ring watermark is not robust against regeneration attacks. In particular, a single regeneration such as Regen-Diff and Regen-VAE can significantly harm the TPR@0.1\%FPR while maintaining reasonable CLIP-FID. Rinsing regenerations significantly lower the TPR@0.1\%FPR at the cost of markedly decreased image quality. A 2x rinsing regeneration (Regen-2xDiff) strikes a balance between both low-TPR@0.1\%FPR and high image quality. 
In regards to the Stable Signature, Figure \ref{fig:2d_plots} and Table~\ref{tab:leaderboard_detect} concur with the analysis of \citet{zhao2023invisible} -- regeneration attacks are completely destructive and rinsing regenerations reiterate this phenomenon. The StegaStamp is mildly affected by regenerations, and only by diffusive attacks, including our novel rinsing and prompted regenerations.

\textbf{Adversarial Attacks.}
Deep neural networks are vulnerable to adversarial examples, \citep{ilyas2019adversarial, chakraborty2018adversarial}. 
In \ours, we explore watermark robustness against two types of adversarial attacks.

\textbf{(A) Embedding Attacks.}
Watermark detection can be thwarted by perturbations on image embedding. Such attacks have been used against Multimodal Large Language Models like GPT-4V \citep{dong2023robust} and shown good transferability \citep{inkawhich2019feature}. We examine if attacks on off-the-shelf embedding models can transfer to watermark detectors.
Given an encoder $f: \mathcal{X} \rightarrow \mathcal{Z}$ mapping images to latent features, we craft an adversarial image $x_{adv}$ to diverge its embedding from the original watermarked image $x$, within an $l_{\infty}$ perturbation ball limit: 
%\begin{align*}
    $\max_{x_{adv}} \|f(x_{adv}) - f(x)\|_2, \  \text{s.t. } \|x_{adv}-x\|_{\infty} \leq \epsilon. $ %\label{eq:adv_emb}
%\end{align*}
We approximately solve this using the PGD~\citep{madry2017towards} algorithm (see details in Appendix~\ref{app:adv_emb}), and see if the adversarial image transfers to real watermark detectors.

We evaluate five off-the-shelf encoders.
\textbf{AdvEmbB-RN18} uses a pre-trained ResNet18~\citep{he2016deep}, targeting the pre-logit feature layer.
\textbf{AdvEmbB-CLIP} employs CLIP's~\citep{radford2021learning} image encoder. 
\textbf{AdvEmbG-KLVAE8} utilizes the encoder of KL-VAE (f8) which is used in the victim latent diffusion model. This is a grey-box setting but reflects the use of public VAEs in proprietary models (for example, DALLE$\cdot 3$ uses a public KL-VAE according to \url{https://cdn.openai.com/papers/dall-e-3.pdf}). 
Further, we do ablation studies on KL-VAE (f16), which has a different architecture but is trained on the same data, and on SDXL-VAE~\citep{podell2023sdxl}, an enhanced version of KL-VAE (f8). They are black-box attacks and are labeled \textbf{AdvEmbB-KLVAE16} and \textbf{AdvEmbB-SdxlVAE}.

\begin{figure}
\vspace{-0.5em}
    \centering
    \includegraphics[trim={0 20px 0 0}, width=\linewidth]{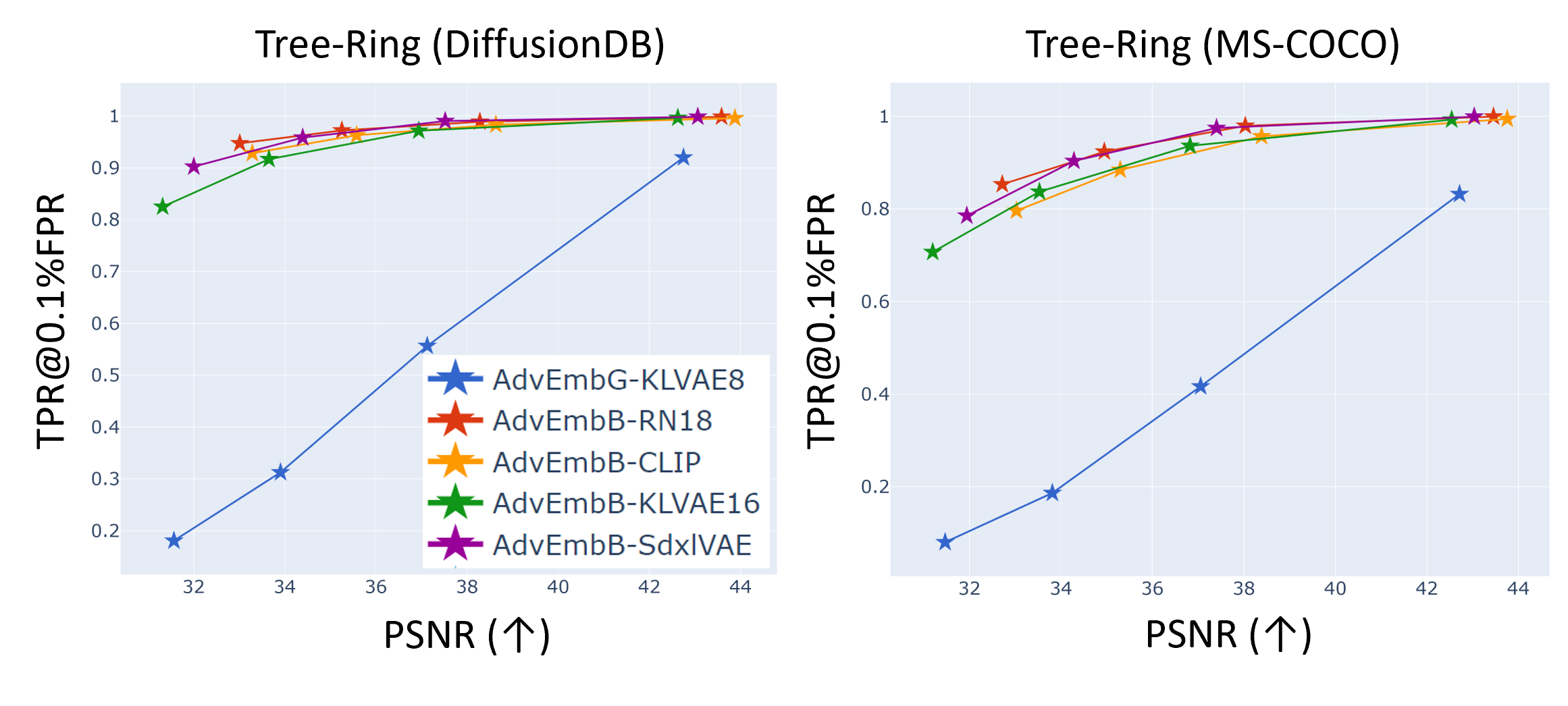}
    \vspace{-2em}
    \caption{Adversarial embedding attacks target Tree-Ring at strengths of \{2/255, 4/255, 6/255, 8/255\}. Tree-Ring shows vulnerability to embedding attacks, especially when the adversary can access the VAE being used.}
    \vspace{-2em}
    \label{fig:adv_emb}
\end{figure}

As shown in \Cref{fig:adv_emb}, Tree-Ring is vulnerable to embedding attacks, particularly under the grey-box condition where TPR@0.1\%FPR can drop to nearly zero, effectively removing most watermarks. 
This is because the detection process of Tree-Ring first maps the image to the latent representation through the encoder of KL-VAE (f8), then conducts inverse DDIM to retrieve the watermark. 
The embedding attack changes the latent representation severely; therefore, watermark retrieval becomes very difficult. 
Using similar yet distinct VAEs, attack effectiveness diminishes but still manages to remove some watermarks, with KL-VAE (f16), trained on the same images, demonstrating the highest transferability.
CLIP-based attacks also achieve some success, especially on natural images like MS-COCO, likely due to CLIP being trained on natural images akin to those in MS-COCO, enhancing the transferability.
Conversely, Stable Signature and StegaStamp demonstrate robustness against embedding attacks (\cref{fig:2d_plots}), likely because their detectors are trained independently from generative models, differing significantly from standard classifiers and VAEs. Hence, our attacks fail to effectively transfer to their detectors.

\begin{figure}[t]
% \vspace{-1em}
      \centering
      \includegraphics[width=\linewidth]{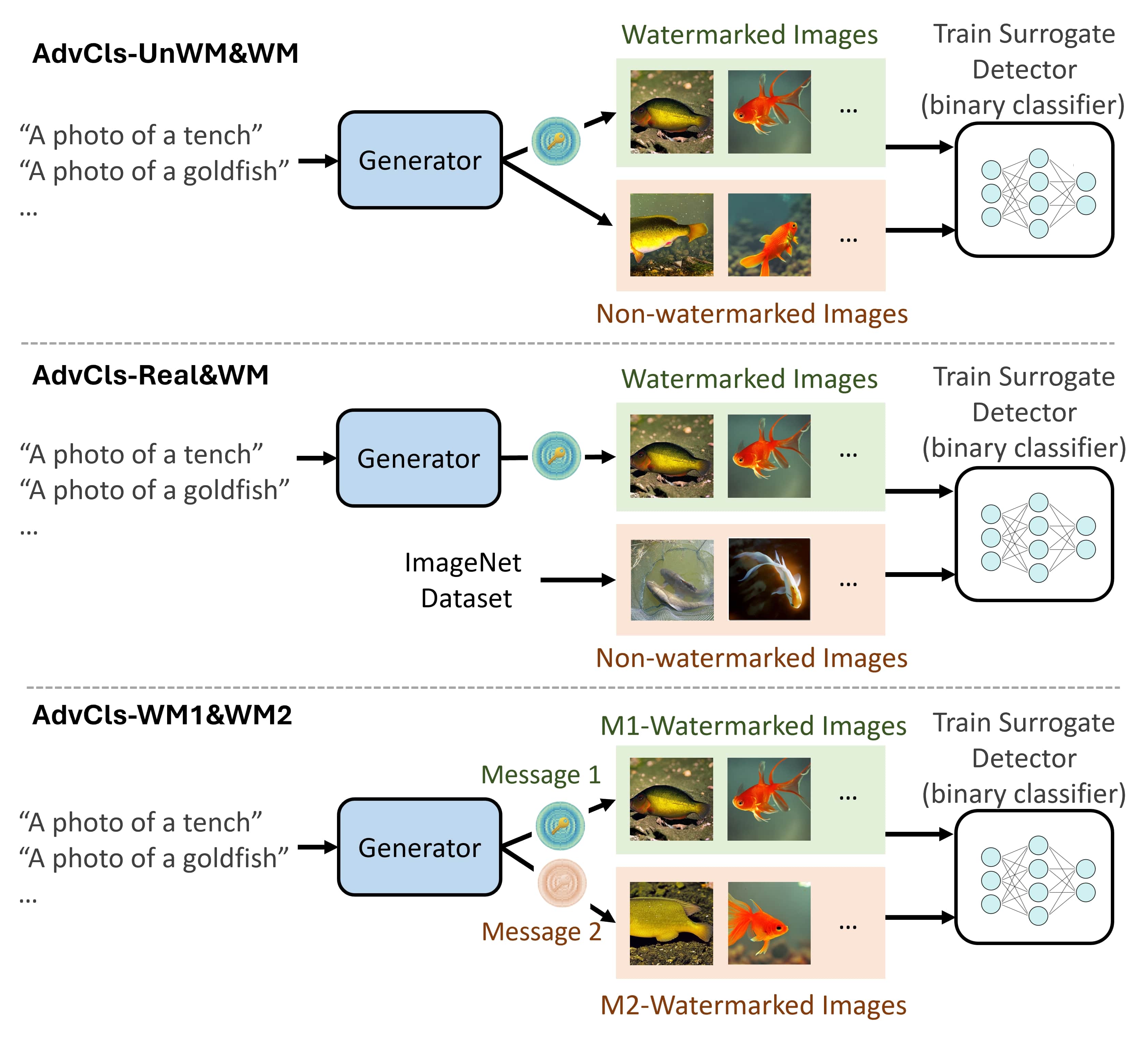}
      \vspace{-2em}
      \caption{Three settings for training the surrogate detector. The Generator is the victim generator under attack. We externalize the watermarking process for simplicity, but it could be in-processing watermarks. After training the surrogate detectors, the adversary performs PGD attacks on them to flip the labels.}
      \label{fig:illu_adv}
      \vspace{-1.5em}
  \end{figure}

\textbf{(B) Surrogate Detector Attacks.}
Watermark detection hinges on a detector that decodes and verifies messages from watermarked images. 
Adversaries might acquire numerous watermarked and non-watermarked images to train a surrogate detector, and transfer attacks on it to the actual watermark detector. \Cref{fig:illu_adv} explores our various settings.

\textbf{AdvCls-UnWM\&WM} trains a surrogate detector with both watermarked and non-watermarked images from the victim generative model, as per \citet{saberi2023robustness}. Note that this is an unrealistic setting for proprietary models since all their outputs are assumed to be watermarked. 
\textbf{AdvCls-Real\&WM} trains the surrogate watermark detector with watermarked and non-watermarked images, where non-watermarked images are sampled from the ImageNet dataset (not from the generative model). This approach is more applicable to proprietary models.
\textbf{AdvCls-WM1\&WM2} only uses watermarked images. 
It actually trains a surrogate watermark message classifier to distinguish two users.
Suppose the system assigns a particular message to each user for identification purposes, the adversary can collect the training data from two users' outputs, with an identical set of prompts. Adversarial attacks on this surrogate model aim at user misidentification.
All surrogate detectors are fine-tuned on ResNet18. 
We use ImageNet text prompts ``A photo of a \{\textit{class name}\}'' to generate training images (see details in Appendix~\ref{app:adv_cls}). 

With the trained surrogate detector $f: \mathcal{X} \rightarrow \mathcal{Y} $, where $\mathcal{Y}=\{0,1\}$, adversaries launch targeted attacks. The goal is to craft an adversarial image $x_{adv}$ from an original image $x$ so that $f$ incorrectly predicts the target label $y_{target}$ (i.e., wrong label), minimizing the following with cross-entropy loss:
% \begin{align*}
    $\min_{x_{adv}}L(f(x_{adv}), y_{target}), \  \text{s.t.}\  \|x_{adv}-x\|_{\infty} \leq \epsilon.$
% \end{align*}
It enables adversaries to erase watermarks from marked images or implant them into clean images in the first two settings, and to disrupt user identification as well as watermark detection in the third setting. We solve it with the PGD algorithm.

\begin{wrapfigure}{r}{0.5\linewidth}
    \centering
    \vspace{-1em}
    \includegraphics[width=\linewidth]{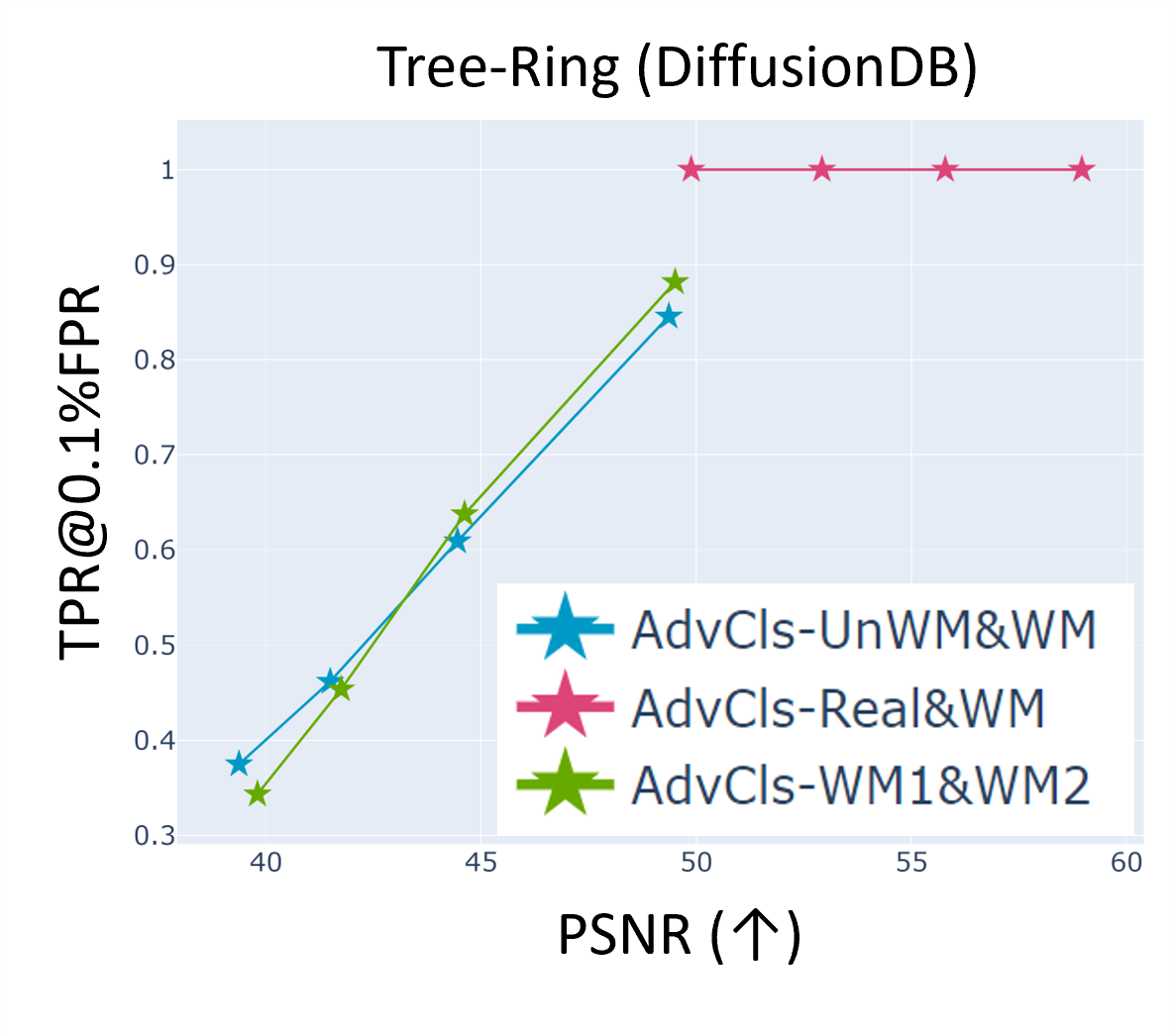}
    \vspace{-2.7em}
    \caption{Adversarial surrogate detector attacks on Tree-Ring.} 
    \vspace{-0.8em}
    \label{fig:adv_su}
\end{wrapfigure}
Figure~\ref{fig:adv_su} shows Tree-Ring's vulnerability to surrogate detector-based attacks. 
In \textbf{AdvCls-UnWM\&WM}, the adversary accessing non-watermarked images has good transferability and removes watermarks effectively. However, it fails to add watermarks to clean images (spoofing attack), as detailed in \Cref{fig:adv_spoof}.
The reason behind this is explored in Appendix~\ref{app:adv_results}, where we find the attacker disrupts the entire latent space, not just the watermark (as shown in \Cref{fig:visual_unwm_wm}). Conversely, the spoofing attack fails to embed the precise watermark. 
\textbf{AdvCls-Real\&WM} attack fails entirely, likely due to the surrogate model appearing to differentiate real from generated images, using broader features than the watermark.
The newly proposed \textbf{AdvCls-WM1\&WM2} successfully attacks Tree-Ring using only watermarked images. 
Like the first scenario, the surrogate model fails to precisely locate watermarks but learns the mapping to the latent feature space, allowing a PGD attack to remove the watermark by disturbing the entire latent space (see \Cref{fig:visual_wm1_wm2}). 
In user identification tasks (\cref{fig:adv_ident}), the attack doesn't consistently mislead the detector into misidentifying User1's watermarked images as User2's (targeted misidentification). Instead, imprecise perturbations often lead to incorrect attribution of User1's images to others.

\cref{fig:2d_plots} shows that Stable Signature and StegaStamp are robust to these attacks. 
Even with high surrogate classifier accuracy in AdvCls-UnWM\&WM, adversarial examples fail to transfer to the true detector, possibly due to reliance on different features than those used by the true detector.

%% file: icml_sec/5_results.tex
\section{Benchmarking Results and Analysis}
\label{sec:results}

We extensively evaluate the security of three prominent watermarking algorithms (according to Appendix~\ref{app:selection}), Stable Signature, Tree-Ring, and StegaStamp, respectively representing three major watermarking types: in-processing via model modification, in-processing via random seed modification, and post-processing. 
We conduct thorough evaluations with images from DiffusionDB \cite{wang2022diffusiondb}, MS-COCO \cite{lin2014microsoft}, and the DALL$\cdot$E3 datasets; see Appendix~\ref{app:dataset} for details. Note that our evaluation process can be applied to any watermark (as shown in Appendix~\ref{app:add_watermark}).

\textbf{Performance vs. Quality 2D plots.} We evaluate 3 watermarking methods under 26 attacks, and report results across 3 datasets in \Cref{fig:all_diffusiondb_1} to \Cref{fig:all_dalle_2}. The quality of images post-attack is evaluated using 8 metrics and the detection performance is measured by TPR@0.1\%FPR. 
\Cref{fig:tree-ring-heat} shows that different quality metrics yield a similar ranking of attacks. 
Consequently, we aggregate these metrics into a single, unified quality metric — \textit{Normalized Quality Degradation}, with lower scores indicating lesser quality degradation caused by attacks. 
Furthermore, we aggregate the results across three distinct datasets, and derive the  unified Performance vs. Quality degradation 2D plots in \cref{fig:2d_plots}, visualizing
the unified evaluation results for each watermarking method against each attack. 
We defer the aggregation details to Appendix~\ref{app:eval_details}.
Based on these unified 2D plots, we benchmark watermarks and attacks in the following sections.

\subsection{Benchmarking Watermark Robustness }
\label{sec:eval_wm}
\begin{figure*}
\centering
\vspace{-0.5em}
\hfill
\begin{subfigure}{0.5\textwidth}
\centering
    \includegraphics[width=0.8\linewidth]{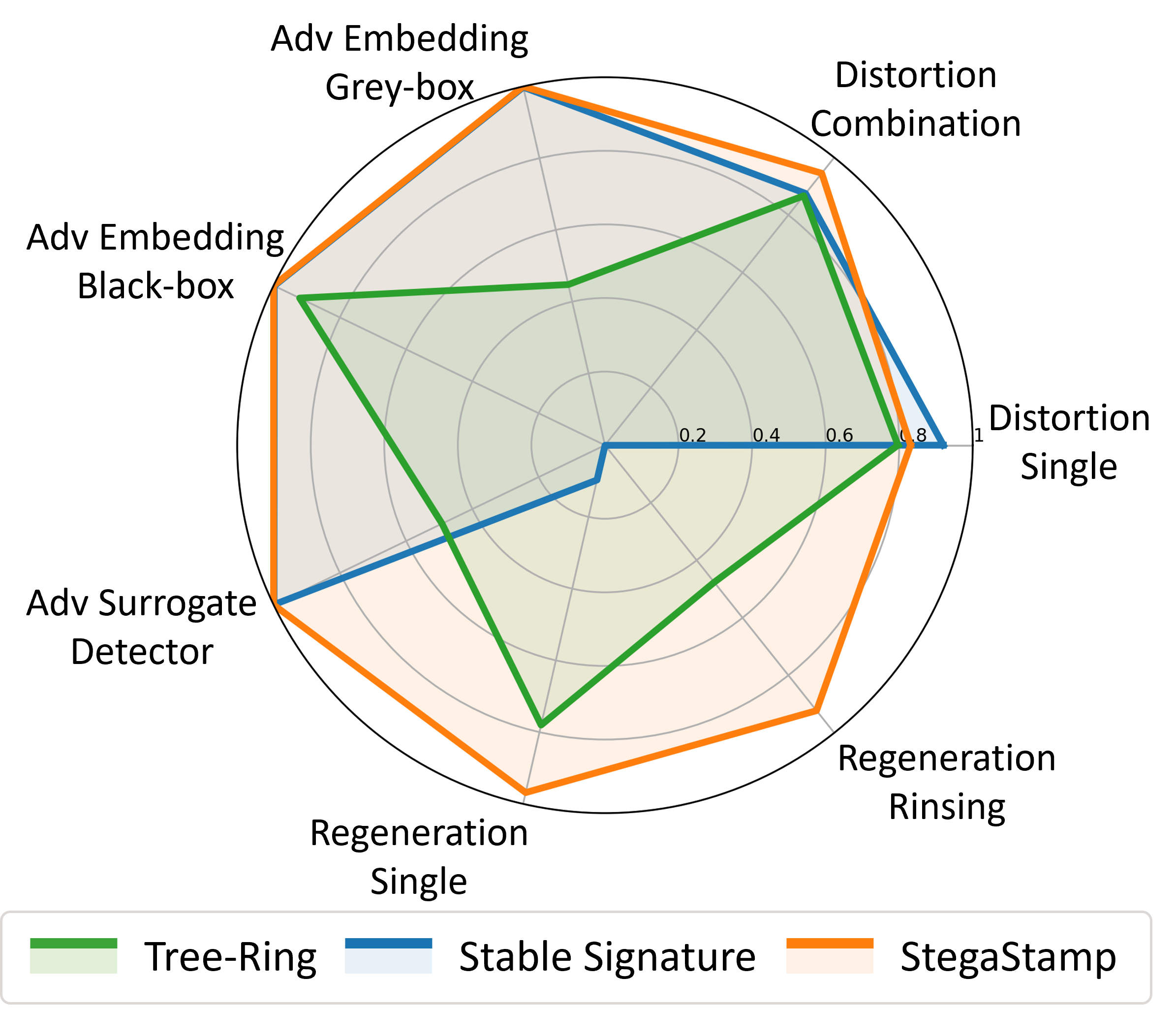}
    \caption{Average TPR@0.1\%FPR under different types of attacks.}
    \label{subfig:radar_detect}
\end{subfigure}
\hfill
\begin{subfigure}{0.43\textwidth}
\centering
    \includegraphics[width=\linewidth]{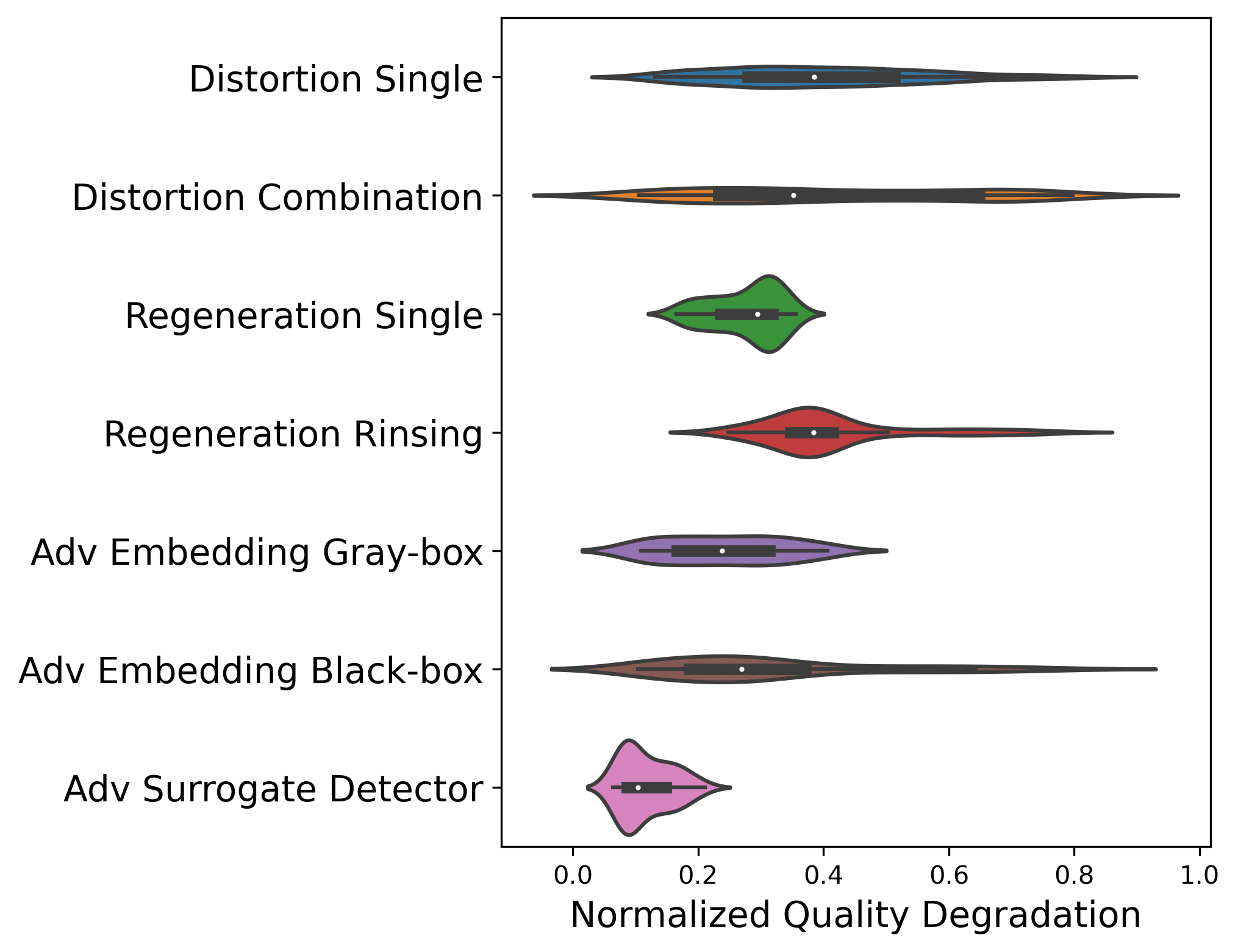}
    \caption{Distributions of quality degradation}
    \label{subfig:radar_quality}
\end{subfigure}\\
\centering
\caption{(a) Detection performance of three watermarks after attacks, measured by Average TPR@0.1\%FPR with lower values (near center) indicating higher vulnerabilities. (b) The distribution of quality degradation. The lower, the better.}
\label{fig:radar}
\end{figure*}

\cref{fig:radar} provides a high-level overview of watermarks' robustness.
We categorize effective attacks into seven types (same as categories in Table~\ref{tab:summary_attacks}): \textit{Distortion Single}, \textit{Distortions Combination}, \textit{Regeneration Single}, \textit{Regeneration Rinsing}, \textit{Adv Embedding Grey-box}, \textit{Adv Embedding Black-box}, and \textit{Adv Surrogate Detector}.
Attacks considered are detailed in Appendix~\ref{app:attacks_for_wm}. 
The Average TPR@0.1\%FPR, calculated for each category across strength levels, assesses watermarking method robustness. 
\cref{fig:radar} shows the robustness of three watermarking methods where the area covered indicates the overall robustness.
\Cref{fig:radar} shows the distribution of quality degradation for each type of attack to illustrate the potential trade-off between attack effectiveness and image quality.

\textbf{\ours provides a clear comparison of watermarks' robustness and reveals undiscovered vulnerabilities.}
~\Cref{fig:radar} reveals that StegaStamp occupies the largest area, signaling its exceptional robustness. Tree-Ring follows suit with a smaller area, and Stable Signature occupies the least space. 
Interestingly, different watermarking methods exhibit vulnerabilities to different types of attacks.
Tree-Ring is particularly vulnerable to adversarial attacks introduced in this paper, with a significant vulnerability to grey-box embedding and surrogate detector attacks.
It is also vulnerable to regeneration rinsing attacks.
Stable Signature is vulnerable to almost all regeneration attacks. 
All three watermarks maintain a relative robustness against distortions. 
Furthermore, as observed in \Cref{fig:radar}, adversarial attacks generally cause less quality degradation, highlighting their potency against Tree-Ring watermarks. 
\ours offers an apple-to-apple comparison of watermarks through a multi-dimensional stress test of their robustness, enabling a nuanced and comprehensive understanding of their security in various scenarios.

\subsection{Benchmarking Attacks}\label{sec:eval_attack}

\input{table/leaderboard}
Table~\ref{tab:leaderboard_detect} features a leaderboard ranking attacks based on their impact on detection performance and image quality. We assess attacks using performance thresholds (TPR@0.1\%FPR=0.95 and TPR@0.1\%FPR=0.7) and quality degradation at these thresholds (Q@0.95P and Q@0.7P). Additionally, we evaluate average performance (Avg P) and quality degradation (Avg Q) across all strengths. These metrics are used to rank 26 attacks for each watermarking method, with details deferred to Appendix~\ref{app:bench_attacks}.

\textbf{Attack effectiveness varies among watermarks.}
Table~\ref{tab:leaderboard_detect} shows variability in attack efficiency across watermarking methods. Metrics like Q@0.95P and Q@0.7P provide nuanced comparisons, while Avg P and Avg Q offer insights into overall attack potency and image quality impact. Our analysis identifies each watermark's specific weaknesses to certain attacks. For instance, AdvCls-UnWM\&WM, AdvCls-WM1\&WM2, and AdvEmbG-KLVAE8 are notably effective against Tree-Ring, whereas Regen-Diff and Regen-DiffP are more potent against Stable Signature.
Regeneration attacks impact StegaStamp but do not greatly affect its average detection performance; in contrast, certain distortion attacks significantly lower detection performance, at the cost of quality degradation.
No single attack excels across all watermarking methods, yet regeneration attacks exhibit some level of consistent effectiveness. This significant variation in attack effectiveness emphasizes the imperative for diverse and watermark-tailored defensive strategies.

\begin{figure}[b]
    \centering
    \vspace{-2em}
    \includegraphics[trim={0 20px 0 0}, width=\linewidth]{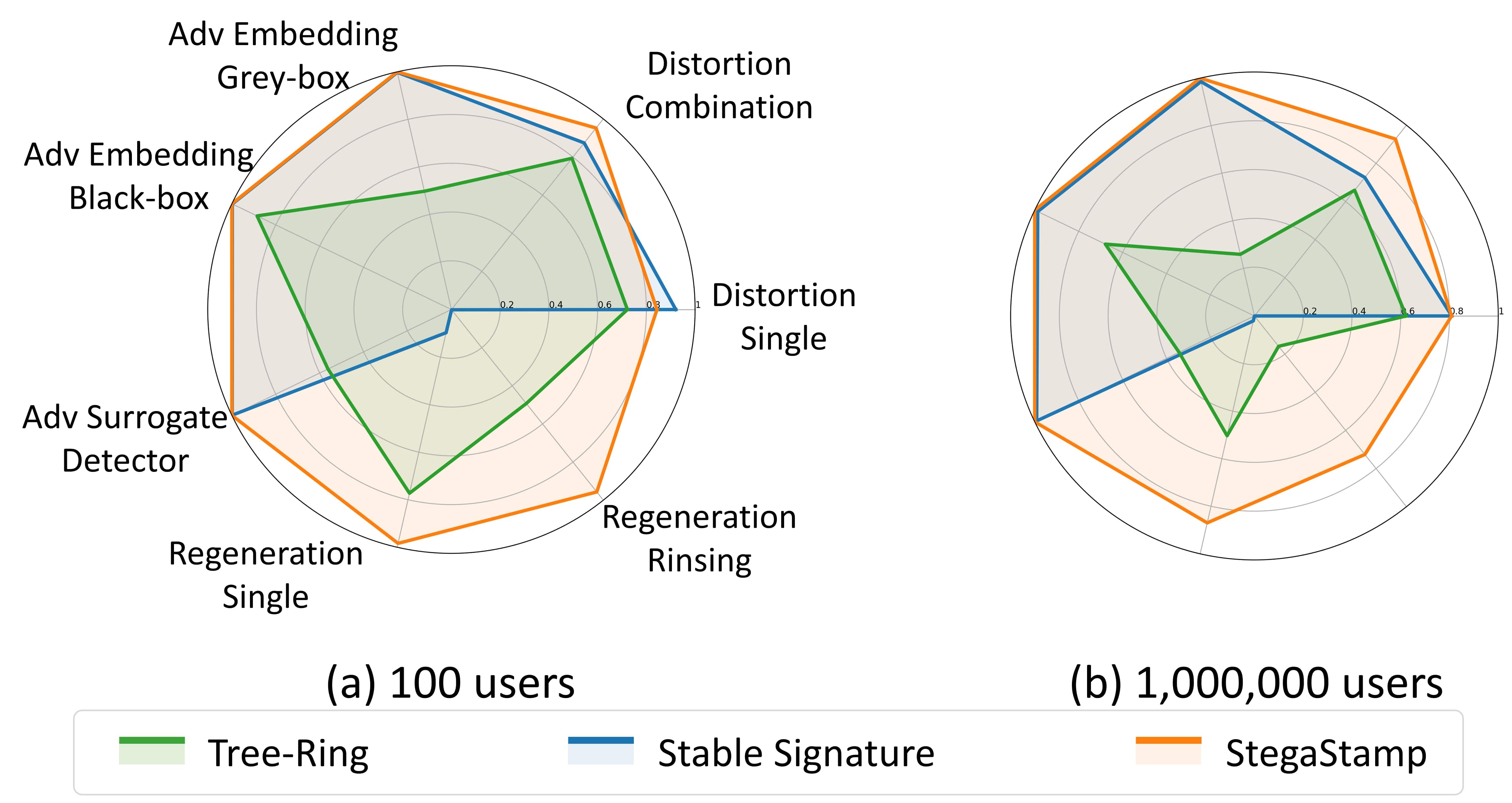}
    \vspace{-1em}
    \caption{Identification accuracy of three watermarks after attacks.}
    \label{fig:radar_id}
    \vspace{-1em}
\end{figure}

\subsection{Benchmarking Results for User Identification}
\label{apd:more_result_identification}
We detail the user identification results, following the evaluation method from Section~\ref{subsec:std_eval}. The key distinction here is the use of identification accuracy as the performance metric. Our study includes scenarios with 100, and 1 million users, reflecting a range of real-world conditions. Utilizing the same evaluation approach, we generate unified Performance vs. Quality degradation 2D plots (\cref{fig:2d_plots_ident}), radar plots for watermark comparison (\cref{fig:radar_id}), and an attack leaderboard in the identification context (Table~\ref{tab:leaderboard_ident}).

\textbf{Identification results mirror findings from detection, showing similar trends in watermark robustness and attack effectiveness.}
\Cref{fig:radar_id} and Table~\ref{tab:leaderboard_ident} reveal that trends in watermark robustness and attack potency closely match those in detection, largely because both rely on precise watermark decoding. Notably, watermarks become more vulnerable as user numbers increase, a trend particularly evident in attacks that already strongly affect detection. Since identification demands more accurate decoding, its vulnerability amplifies with user growth. Thus, insights gained from detection scenarios generally apply to identification, especially when attacks are not identification-specific. 
However, novel attacks such as our AdvCls-WM1\&WM2, may target user identification. Watermarking strategies should evolve to address emerging challenges in both detection and identification.

\subsection{Discussions}
\textbf{Understanding watermark vulnerabilities.} 
Tree-Ring is particularly vulnerable to adversarial attacks likely due to its unique watermark detection process. 
The detection first encodes an image into a latent space using a VAE encoder, then reverses the diffusion process to extract the initial noise vector and compares it with a key.
Consequently, the detection hinges on the integrity of the latent feature space, and thus disturbances inside this domain significantly hinder watermark recovery.
Embedding attacks, especially the grey-box setting, effectively disrupt the latent features without altering the perceptual appearance of the image, making them highly effective against Tree-Ring. 
We also observe a similar phenomenon for surrogate detector attacks (\cref{fig:visual_unwm_wm}, \cref{fig:visual_wm1_wm2}), which also successfully disturb latent features, including those related to the watermark.
Stable Signature is vulnerable to regeneration attacks due to its unique watermarking protocol. 
Recall that latent diffusion models first perform diffusion in the latent space, and then map back to the image space through a VAE decoder. 
To embed watermarks, Stable Signature roots the watermark in the VAE decoder by training. 
However, regeneration attacks circumvent this special decoder by using an alternate VAE or diffusion model with a different decoder. As a result, the regenerated images are stripped of the original watermarks.

\textbf{Limitations of attacks.} 
As shown in Table~\ref{tab:attacker_know}, we focus on realistic attacks where attackers have very limited knowledge, unaware of the watermarking algorithm in all scenarios. 
Distortion, regeneration, and adversarial embedding attacks (except for the grey-box setting) are universal attacks that do not use any watermark or model information.
Therefore, their effectiveness may vary.
Adversarial surrogate detector attacks target a watermark by training a surrogate detector on watermarked images. 
However, we found that they do not always work due to the transferability problem. 
That is, since the attackers do not know the true detector, the architecture of the surrogate detector (e.g., ResNet18 in this paper) may differ significantly from the true one. 
Additionally, there might be many features that can distinguish non-watermarked and watermarked images.
Hence, despite achieving high classification accuracy, the surrogate may rely on features different from those of the true detector, leading to unsuccessful transfer of attacks.
Enhanced attacker knowledge, such as the watermarking algorithm, could facilitate more effective adversarial attacks, as explored in \cite{lukas2023leveraging}.

\textbf{Potential strategies to improve robustness.}
 Although we reveal many vulnerabilities of existing watermarks, there are potential ways to improve them. For watermarks which rely on image perturbations for encoder/decoder training (Stegastamp, Stable Signature), including more types of transformations may improve robustness. For example, we have observed in internal testing that training Stable Signature's extractor with blur and rotation transformations as data augmentations improves its robustness to these transformations but also marginally reduces the encoded image quality. Similar to blur and rotation, we can add other transformations such as adversarial perturbations and regeneration as data augmentations to improve robustness towards them.

 There is also ample opportunity to improve the algorithmic frameworks themselves. For example, Tree-Ring relies on DDIM inversion, which we found is not accurate even without attack, directly affecting the watermark detection accuracy. Future work can improve it by incorporating cutting-edge techniques on more accurate DDIM inversion. For watermarks such as Tree-Ring, one may also insert a trainable U-Net which restores the watermark before it is extracted. Such a strategy may degrade the image to enhance the signal of the message, but this is irrelevant from the perspective of the image owner whose only goal is to simply detect their watermark.

For more agnostic strategies: (1) Incorporating redundant bits. This technique, known as error correction coding, can help reconstruct the original message even when parts of the watermark are corrupted. (2) A hybrid approach. Since different watermarks have varied vulnerabilities, one can try to combine different watermarks, leveraging their strengths to defend a wider range of attacks.

\subsection{Summary of Takeaway Messages}

\textbf{\ours provides a standardized framework for benchmarking watermark robustness and attack potency.} 
\ours evaluates both detection and identification tasks.
It unifies the quality metrics and assesses attack potency against both performance degradation and quality degradation. 
The Performance vs. Quality 2D plots allow for a comprehensive analysis of various watermarks in one unified framework. 
With over twenty attacks tested, \ours exposes new vulnerabilities in popular watermarking techniques.

\textbf{Different watermarking methods have different vulnerabilities.}
Our analysis reveals significant differences in watermark vulnerabilities against attacks. Specifically, Tree-Ring is more vulnerable to adversarial attacks, which generally cause less quality degradation, while Stable Signature is susceptible to most regeneration attacks. 
This diversity in vulnerabilities highlights the imperative for watermarking methods to identify and strengthen their specific weak areas.

\textbf{Avoid using publicly available VAEs.} 
\ours demonstrates the risks of using publicly available VAEs in watermarked diffusion models. An adversarial embedding attack using the same VAE easily compromises Tree-Ring by altering latent features with little visual change. 
Stable Signature's design renders it vulnerable to regeneration attacks that use a VAE with an encoder identical to the victim model's VAE encoder, while coupled with a different decoder.
Today's proprietary generators, like DALL$\cdot 3$, typically train the latent diffusion model themselves but use a publicly available VAE. 
This practice, especially with Tree-Ring or Stable Signature watermarking, increases vulnerability, pointing to a critical security concern in those popular AI services.

\textbf{The robustness of StegaStamp potentially illuminates a path for future robust watermarks.}
The StegaStamp watermark \cite{tancik2020stegastamp} stands out in our evaluation for its robustness. 
Designed for physical-world use which requires high robustness, StegaStamp is trained with a series of distortions that mimic real-world scenarios, significantly enhancing its robustness. 
However, it's important to recognize the potential trade-off between watermark robustness and quality. 
As a post-processing method, the original paper finds that StegaStamp may introduce artifacts. 
In contrast, this might not pose a problem for in-processing watermarks. Therefore, in-processing watermarks could still benefit from incorporating augmentation or adversarial training.

%% file: table/leaderboard.tex
\begin{table*}[!htbp]
\label{attack-ranking}
\vspace{-1em}
\caption{\label{tab:leaderboard_detect}\textbf{Comparison of attacks across three watermarking methods in detection setup.} Q denotes the normalized quality degradation, and P denotes the performance as derived from \cref{fig:2d_plots}. Q@0.95P measures quality degradation at a 0.95 performance threshold where "inf" denotes cases where all tested attack strengths yield performance above 0.95, and "-inf" where all are below. A similar notation applies to Q@0.7P. Avg P and Avg Q are the average performance and quality over all the attack strengths. The lower the performance and the smaller the quality degradation, the stronger the attack is. For each watermarking method, we rank attacks by Q@0.95P, Q@0.7P, Avg P, Avg Q, in that order, with lower values ($\downarrow$) indicating stronger attacks. The top 5 attacks of each watermarking method are highlighted in red.}
\begin{adjustbox}{max width=\textwidth}
\renewcommand{\arraystretch}{0.8}
\begin{tabular}{cccccccccccccccccc}
\toprule
\multirow{2}{*}{Attack}      & \multicolumn{5}{c}{Tree-Ring} &  & \multicolumn{5}{c}{Stable Signature} &  & \multicolumn{5}{c}{StegaStamp} \\ \cline{2-6} \cline{8-12} \cline{14-18} 
                             & Rank& Q@0.95P& Q@0.7P & Avg P & Avg Q &  & Rank& Q@0.95P& Q@0.7P  & Avg P   & Avg Q    &  & Rank& Q@0.95P& Q@0.7P  & Avg P  & Avg Q  \\
\midrule
Dist-Rotation   & 11 & 0.464 & 0.521 & 0.375 & 0.648 &  & 12 & 0.624 & 0.702 & 0.594 & 0.650 &  & \textcolor{red}{5}  & 0.423 & 0.498 & 0.357 & 0.616 \\
Dist-RCrop      & 18 & 0.592 & 0.592 & 0.332 & 0.463 &  & 24 & inf   & inf   & 0.995 & 0.461 &  & 6  & 0.602 & 0.602 & 0.540 & 0.451 \\
Dist-Erase      & 26 & inf   & inf   & 1.000 & 0.490 &  & 25 & inf   & inf   & 0.998 & 0.489 &  & 25 & inf   & inf   & 1.000 & 0.483 \\
Dist-Bright     & 25 & inf   & inf   & 0.997 & 0.304 &  & 23 & inf   & inf   & 0.998 & 0.305 &  & 22 & inf   & inf   & 0.998 & 0.317 \\
Dist-Contrast   & 22 & inf   & inf   & 0.998 & 0.243 &  & 20 & inf   & inf   & 0.998 & 0.243 &  & 17 & inf   & inf   & 0.998 & 0.231 \\
Dist-Blur       & 20 & 0.861 & 1.112 & 0.563 & 1.221 &  & \textcolor{red}{5}  & -inf  & -inf  & 0.000 & 1.204 &  & 9  & 0.848 & 0.962 & 0.414 & 1.198 \\
Dist-Noise      & 16 & 0.548 & inf   & 0.980 & 0.395 &  & 8  & 0.402 & 0.520 & 0.870 & 0.390 &  & 24 & inf   & inf   & 1.000 & 0.360 \\
Dist-JPEG       & 12 & 0.499 & 0.499 & 0.929 & 0.284 &  & 9  & 0.485 & 0.485 & 0.793 & 0.284 &  & 21 & inf   & inf   & 0.998 & 0.263 \\
DistCom-Geo     & 13 & 0.525 & 0.593 & 0.277 & 0.768 &  & 13 & 0.850 & inf   & 0.937 & 0.767 &  & 7  & 0.663 & 0.693 & 0.396 & 0.733 \\
DistCom-Photo   & 22 & inf   & inf   & 0.998 & 0.242 &  & 20 & inf   & inf   & 0.998 & 0.243 &  & 17 & inf   & inf   & 0.998 & 0.239 \\
DistCom-Deg     & 19 & 0.620 & inf   & 0.892 & 0.694 &  & 7  & 0.206 & 0.369 & 0.300 & 0.679 &  & 8  & 0.826 & 0.975 & 0.852 & 0.664 \\
DistCom-All     & 14 & 0.539 & 0.751 & 0.403 & 0.908 &  & 11 & 0.538 & 0.691 & 0.334 & 0.900 &  & 10 & 0.945 & 1.101 & 0.795 & 0.870 \\ \midrule
Regen-Diff      & \textcolor{red}{5}  & -inf  & 0.307 & 0.612 & 0.323 &  & \textcolor{red}{1}  & -inf  & -inf  & 0.001 & 0.300 &  & \textcolor{red}{1}  & 0.331 & inf   & 0.943 & 0.327 \\
Regen-DiffP     & \textcolor{red}{4}  & -inf  & 0.307 & 0.601 & 0.327 &  & \textcolor{red}{1}  & -inf  & -inf  & 0.001 & 0.303 &  & \textcolor{red}{1}  & 0.333 & inf   & 0.940 & 0.329 \\
Regen-VAE       & 17 & 0.578 & 0.578 & 0.832 & 0.348 &  & 10 & 0.545 & 0.545 & 0.516 & 0.339 &  & 23 & inf   & inf   & 1.000 & 0.343 \\
Regen-KLVAE     & 22 & inf   & inf   & 0.990 & 0.233 &  & 6  & -inf  & 0.176 & 0.217 & 0.206 &  & 17 & inf   & inf   & 1.000 & 0.240 \\
Rinse-2xDiff    & 6  & -inf  & 0.333 & 0.510 & 0.357 &  & \textcolor{red}{3}  & -inf  & -inf  & 0.001 & 0.332 &  & \textcolor{red}{4}  & 0.391 & inf   & 0.941 & 0.366 \\
Rinse-4xDiff    & 7  & -inf  & 0.355 & 0.443 & 0.466 &  & \textcolor{red}{4}  & -inf  & -inf  & 0.000 & 0.438 &  & \textcolor{red}{3}  & 0.388 & inf   & 0.909 & 0.477 \\ \midrule
AdvEmbG-KLVAE8  & \textcolor{red}{3}  & -inf  & 0.164 & 0.448 & 0.253 &  & 20 & inf   & inf   & 0.998 & 0.249 &  & 17 & inf   & inf   & 1.000 & 0.232 \\
AdvEmbB-RN18    & 10 & 0.241 & inf   & 0.953 & 0.218 &  & 17 & inf   & inf   & 0.999 & 0.212 &  & 14 & inf   & inf   & 1.000 & 0.196 \\
AdvEmbB-CLIP    & 15 & 0.541 & inf   & 0.932 & 0.549 &  & 26 & inf   & inf   & 0.999 & 0.541 &  & 25 & inf   & inf   & 1.000 & 0.488 \\
AdvEmbB-KLVAE16 & 8  & 0.195 & inf   & 0.888 & 0.238 &  & 19 & inf   & inf   & 0.997 & 0.233 &  & 14 & inf   & inf   & 1.000 & 0.206 \\
AdvEmbB-SdxlVAE & 9  & 0.222 & inf   & 0.934 & 0.221 &  & 17 & inf   & inf   & 0.998 & 0.219 &  & 14 & inf   & inf   & 1.000 & 0.204 \\
AdvCls-UnWM\&WM & \textcolor{red}{1}  & -inf  & 0.102 & 0.499 & 0.145 &  & 14 & inf   & inf   & 0.999 & 0.101 &  & 11 & inf   & inf   & 1.000 & 0.101 \\
AdvCls-Real\&WM & 21 & inf   & inf   & 1.000 & 0.047 &  & 14 & inf   & inf   & 0.998 & 0.092 &  & 11 & inf   & inf   & 1.000 & 0.106 \\
AdvCls-WM1\&WM2 & \textcolor{red}{1}  & -inf  & 0.101 & 0.492 & 0.139 &  & 14 & inf   & inf   & 0.999 & 0.084 &  & 13 & inf   & inf   & 1.000 & 0.129 \\
\bottomrule
\end{tabular}
\end{adjustbox}
\vspace{-1em}
\end{table*}

%% file: icml_sec/7_impact.tex
\section*{Acknowledgements}
We thank Souradip Chakraborty and Amrit Singh Bedi for insightful discussions.

 An, Ding, Rabbani, Xu, Deng, Zhu, and Huang are supported by DARPA Transfer from Imprecise and Abstract Models to Autonomous Technologies (TIAMAT) 80321, National Science Foundation NSF-IIS-2147276 FAI, DOD-ONR-Office of Naval Research under award number N00014-22-1-2335, DOD-AFOSR-Air Force Office of Scientific Research under award number FA9550-23-1-0048, DOD-DARPA-Defense Advanced Research Projects Agency Guaranteeing AI Robustness against Deception (GARD) HR00112020007, Adobe, Capital One and JP Morgan faculty fellowships.

Wen and Goldstein are supported by the ONR MURI program, the AFOSR MURI program, the National Science
Foundation (IIS-2212182), the NSF TRAILS Institute (2229885), Capital One Bank, the Amazon Research Award program, and Open Philanthropy.

\section*{Impact Statement}

This work contains research that could be used to remove watermarks from images. However, our research is focused on uncovering vulnerabilities in watermarking systems to guide the development of more robust designs. As publicly available generative imaging services like OpenAI's DALL$\cdot$E, MidJourney, and Bing Image Creator become more popular, the demand for effective watermarks is intensifying. We test and contribute a large collection of distortion, regeneration, and adversarial attacks, setting a benchmark for evaluating and enhancing watermark strength.

As the legal status of AI-generated content evolves, robust watermarking will become increasingly crucial for protecting creative ownership and preventing the misrepresentation of AI-generated content as real. Our research not only contributes to identifying weaknesses in watermarks but also advances the detection capabilities of AI-generated content, supporting the development of this significant aspect of digital watermarking technology.

%% file: icml_sec/X_suppl.tex
\section{A Mini Survey of Image Watermarks} \label{app:related}
\begin{figure*}[!htbp]
    \centering
\includegraphics[width=\linewidth]{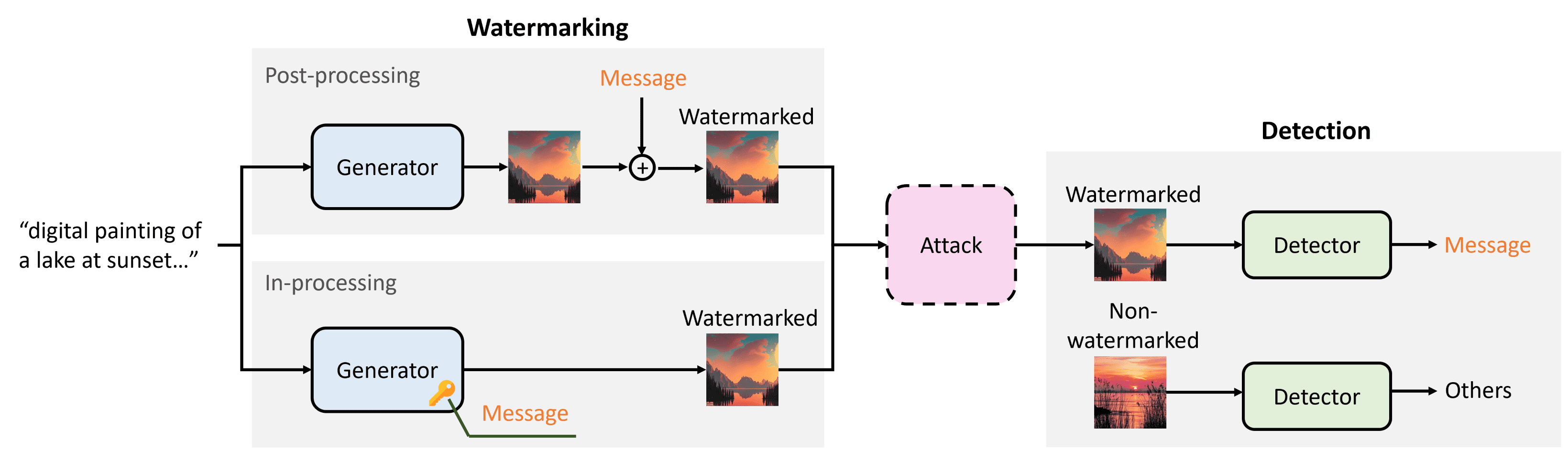}
    \caption{\textbf{An illustration of a robust watermarking workflow.} An AI company provides two services: (1) generate watermarked images, i.e., embed invisible messages, and (2) detect these messages when shown any of their watermarked images. There is an attack stage between the watermarking and detection stages. The watermarked images may experience natural distortions (e.g., compression, re-scaling)  or manipulated by malicious users attempting to remove the watermarks. A robust watermarking method should still be able to detect the original message after an attack.}
    \label{fig:workflow}
\end{figure*}

In this section, we detail the existing landscape of watermarking approaches in the era of AI-Generated Content (AIGC) everywhere. \cref{fig:workflow} depicts our scenario of interest. First, an AI company/owner embeds a watermark into its generated images. Then, if the owner is shown one of their watermarked images at a later point in time, they can identify ownership of it by recovering the watermark message. 
Commonly, users might modify watermarked images for legitimate personal purposes. There are also instances where users attempt to erase a watermark for malicious reasons, such as disseminating fake information or infringing upon copyright. For simplicity, we term any image manipulation as an ``attack.'' 

\paragraph{Watermarking AI-generated Images.}
Imprinting invisible watermarks into digital images has a long and rich history. From conventional steganography to recent generative model-based methods, we categorize popular watermarking techniques into two categories: post-processing methods and in-processing methods. 

\textbf{Post-processing} approaches embed post-hoc watermarks into images. When watermarking AI-generated images, we apply such methods \textit{after} the generation process. 
Post-processing watermarks are model-agnostic and applicable to any image. However, they sometimes introduce human-visible artifacts, compromising image quality.
We review popular post-processing methods.

\textbf{P1) Frequency-domain methods}. 
These methods manipulate the representation of an image in some transform domain~\citep{o1996watermarking, cox1996secure, o1997rotation}. 
The image transform can be a Discrete Wavelet Transform (DWT), Discrete Cosine Transform (DCT)~\citep{cox2007digital}, or SVD decomposition~\citep{chang2005svd}.
These transformations have a range of invariance properties that make them robust to translation and resizing. 
The commercial implementation of Stable Diffusion~\citep{rombach2022high} uses DWTDCT~\citep{al2007combined} to watermark its generated images.
However, many studies have shown that these watermarks are vulnerable to common image manipulations \citep{zhao2023invisible}.  

\textbf{P2) Deep encoder-decoder methods}. 
These methods rely on trained networks for embedding and decoding the watermark  \citep{hayes2017generating}. 
Methods such as HiDDeN~\citep{zhu2018hidden} and RivaGAN~\citep{zhang2019robust} learn an encoder to imprint a hidden message inside an image and a decoder (also called a detector) to extract the message. 
To train robust watermarks, RedMark~\citep{ahmadi2020redmark} integrates differentiable attack layers between the encoder and decoder in the end-to-end training process;
RivaGAN~\citep{zhang2019robust} employs an adversarial network to remove the watermark during  training;
StegaStamp~\citep{tancik2020stegastamp} adds a series of strong image perturbations between the encoder and decoder during training, resulting in watermarks which are robust to real-world distortions caused by photographing an image as it appears on a display.

\textbf{P3) Others}. There are other varieties of post-processing methods that do not fall into \textbf{P1} or \textbf{P2}. SSL~\citep{fernandez2022watermarking} embeds watermarks in self-supervised-latent spaces by shifting the image's features into a designated region. DeepSigns~\citep{rouhani2018deepsigns} and DeepMarks~\citep{chen2019deepmarks} embed target watermarks into the probability density functions of weights and activation maps. Entangled watermarks \citep{jia2021entangled} designs a reinforced watermark based on a target watermark and the task data. 

\textbf{In-processing} methods adapt generative models to directly embed watermarks as part of the image generation process, substantially reducing or eliminating visible artifacts. 
With diffusion models presently dominating the field of image generation, a surge of in-processing approaches specific to these models has recently emerged. 
We categorize current work into three categories.

\textbf{I1) Model modification.} 
\textit{The entire model.} This line of work inherits the encoder-decoder idea and bakes the encoder into the entire generative model. 
This is usually accomplished by watermarking training images with a pre-trained watermark encoder and decoder, then training or fine-tuning the generative model on these watermarked images \citep{yu2021artificial,zeng2023securing, lukas2023ptw}. 
This type of method has been shown to work well on small models like guided diffusion, but suffers from the expensive training of large text-to-image generation models \citep{zhao2023recipe}, making it inapplicable in practice.

\textit{Parts of the model.}
Stable Signature~\citep{fernandez2023stable} follows the above two-stage training pipeline while only fine-tuning the decoder of the latent-diffusion model (LDM) \citep{rombach2022high}, leaving the diffusion component unchanged. This type of watermarker is much more efficient to train. By fine-tuning multiple latent decoders, the model can embed different messages into images. 

The robustness of these two types of model modification critically relies on the robustness of the pre-trained encoder and decoder.

\textbf{I2) Modification of a random seed.} Tree-Ring~\citep{wen2023tree}, different from all the above methods, embeds a pattern into the initial noise vector used by a diffusion model for sampling. 
The pattern can be retrieved at detection time by inverting the diffusion process using DDIM~\citep{song2020denoising} as the sampler. 
This method does not require any training, can easily embed different watermarks, and is robust to many simple distortions and attacks. 
The robustness of Tree-Ring relies on the accuracy of the DDIM inversion.

\paragraph{Removing Watermarks}
Robustness is an essential property of watermarks. 
Evaluations of robustness in existing literature focus on simple image distortions like rotation, Gaussian blur, etc.
Recently, inspired by adversarial purification \cite{nie2022diffusion}, \citet{zhao2023invisible} and \citet{saberi2023robustness} both find that regenerating images by noising and denoising images through a diffusion model or a VAE can effectively remove some watermarks. 
\citet{saberi2023robustness} propose adversarial attacks based on a trained surrogate watermark detector. 
\citet{lukas2023leveraging} also introduces adversarial attacks but requires the knowledge of the watermarking algorithm and a similar surrogate generative model.
\citet{jiang2023evading} studies white-box attacks and black-box query-based attacks.
Some attacks are not possible in realistic scenarios where the attacker has only API access.  Furthermore, existing evaluations use differing quality/performance metrics, making it difficult to compare the effectiveness between watermarking methods and between attacks.

\paragraph{Benchmarks for Image Watermarks.} 
Before the advent of AIGC, there were significant benchmarks introduced that greatly accelerated the progress of watermark standardization \cite{kutter1999fair, tao2014robust, petitcolas2000watermarking}. However, with the development of AIGC, the need to watermark images generated by AI has become urgent, as previous methods were weak in robustness and could not meet current requirements. Nowadays, more and more methods for watermarking images generated by AI have been proposed, but they all use different methods to evaluate robustness. Therefore, this paper proposes a benchmark for the AIGC era.

\section{Formalism of Watermark Detection and Identification}
\label{formalism}
Invisible image watermarks, which are inspired by classical watermarks to protect the intellectual properties of creators, are now applied for a wider range of application scenarios. With the vast development of AI generative models, most current research focuses on applying invisible watermarks to (1) identify AI-generated images (AI Detection)~\citep{saberi2023robustness}, and (2) identify the user who generated the image for source tracking (User Identification)~\citep{fernandez2023stable}.

To fairly evaluate the different watermark methods for different applications, we start from formulating a general, message-based watermarking protocol, partially adopting the notation of~\citep{lukas2023leveraging}, which generalizes most of the existing setups. Let $\theta_G$ denote an image generator, $\mathcal{M}$ the space of watermark messages, and $\mathcal{X}$ the domain of images. We assume $\mathcal{M}$ is a metric space with distance function $D(\cdot,\cdot)$. The choice of message space $\mathcal{M}$ can be very different depending on the watermarking algorithm: for Tree-Ring, messages are random complex Gaussians, while for the Stable Signature and StegaStamp, each message is a length-$d$ binary string, where $d$ denotes the length of the message. For watermarking algorithms following the encoder-decoder training approach, like Stable Signature and StegaStamp, the choice of message length $d$ is fixed after training. Some methods, such as Tree-Ring, enjoy flexible message length at the time of injecting watermarks.

In addition to classifying images as watermarked or non-watermarked, a good detector will often provide a \textit{p-value} for the watermark detection, which measures the probability that the level of watermark strength observed in an image could occur by random chance. The Tree-Ring watermark also includes an image location parameter $\tau$ to embed a message $m\in\mathcal{M}$, but we subsume this under the parameters of $\theta_G$. We now introduce several important watermarking operations: 
\begin{itemize}
\item $\textsf{EMBED}: \theta_G\times \mathcal{M}\rightarrow \mathcal{X}$ is the generative procedure that creates a watermarked image given user-defined parameters of $\theta_G$ (such as prompt, guidance scale, etc. for a diffusion model) and a target message $m\in\mathcal{M}$. 
\item $\textsf{DECODE}: \mathcal{X}\rightarrow \mathcal{M}$ is a recovery procedure of a message $m$ embedded within a watermarked image $x=\mathrm{EMBED}(\theta_G,m)$. In particular, the recovery $m'=\mathrm{DECODE}(x)$ may be imperfect, i.e., $m'\neq m$. 
\item $\textsf{VERIFY}_\alpha:\mathcal{M}\times\mathcal{M}\rightarrow \{0,1\}$ is conducted by the model owner to decide whether $x$ was watermarked by inspecting $m'=\textsf{DECODE}(x)$, where $x=\textsf{EMBED}(\theta_G,m)$. For a decoded message $m'$, we consider the following $p$-value (further discussed in Section \ref{pval}) for evaluating whether the image could have been watermarked using $m$. which is defined as
\begin{equation*}
    p = \textrm{P}_m\bigl(D(\omega,m')< D(m,m') \mid H_0\bigr),
\end{equation*}
where, $D(\omega,m')$ is the similarity between an arbitrary message $\omega \sim \mathcal{M}$ (drawn uniformly at random) and $m'$, and $D(m,m')$ is the similarity between the ground truth message $m$ and the recovered message $m'$. $H_0$ denotes the null hypothesis that the image was generated without knowledge of the watermark (and therefore the recovered message is random). $\textsf{VERIFY}_{\alpha}(m',m)$ returns $1$ if $p<\alpha$, and $0$ otherwise.  In our experiments, we set $\alpha=0.001$.
\end{itemize}

To establish a comprehensive evaluation toolbox, we consider two distinct problems that naturally arise during watermark analysis: detection and identification.  Let $\mathcal{A}:\mathcal{X}\rightarrow\mathcal{X}$ represent an image attack function and denote by $Q$ a fixed subset of messages independently drawn from $\mathcal{M}$ used by $\theta_G$. Further, assume that the owner of $\theta_G$ will only embed messages contained within a finite subset $Q$ drawn randomly from $\mathcal{M}$. 

\subsection{Detection}
In the \textit{watermark detection problem}, given $x=\textrm{EMBED}(\theta_G,m)$, and an attack $x'=\mathcal{A}(x)$, the model owner is tasked with producing \textsf{EMBED} and DECODE protocols which satisfy the following,\\

\textit{(1)} If $x=\textsf{EMBED}(\theta_G,m)$ is a watermarked image, then $\textsf{VERIFY}_\alpha(\textsf{DECODE}(x'))=1$.\\
\textit{(2)} If $x=\textsf{EMBED}(\theta_G,\textsf{NULL})$ is an unwatermarked image, then $\textrm{VERIFY}_\alpha(\textsf{DECODE}(x'))=0$.

For both conditions, a comparison of the extracted message $m'=\textsf{DECODE}(x)$ is performed against all messages in $Q$. Failure of the above conditions is referred to as Type II and Type I errors, respectively. Exploration of the tradeoff between minimization of both error types is an interesting research topic in its own right \cite{zhao2023invisible,saberi2023robustness}.  

\subsection{Identification}
While watermark detection requires only that $\textsf{VERIFY}(\theta_G,x')=1,$ the \textit{watermark identification problem} further requires that one can accurately determine which message from $Q$ is embedded in the image. Rigorously, given $x=\textsf{EMBED}(\theta_G,m)$, an attack $x'=\mathcal{A}(x)$, and $m'=\textsf{DECODE}(\theta_G, x')$, the user requires the \textsf{EMBED} and \textsf{DECODE} to satisfy

\begin{equation*}
    \underset{m'\in Q}{\arg \min \hspace{0.5em}} \textrm{P}\bigl(D(\omega,m)< D(m',m) \mid H_0\bigr) = m,
\end{equation*}
for randomly drawn $\omega \sim \mathcal{M}$ if $x$.

The identification problem is useful in the scenario where the model owner wishes to identify the user who created an image (e.g., a user of DALL$\cdot$E). Note that as $|Q|\rightarrow \infty $, the identification problem becomes difficult as $Q$ will resemble $\mathcal{M}$ in distribution.

\section{Details on Performance Metrics}
\label{pval}
% Use this section as a 'rebuttal' 
\subsection{Clarifications on \texorpdfstring{$p$}{p}-Value}
Here, we clarify the definition of the $p$-value as follows.

Watermark injection and evaluation are often done by encoding a message $m$ into the image, and later recovering the message $m'$, which may be an imperfect recovery. In addition to classifying images as watermarked or non-watermarked, a good detector will often provide a \textit{$p$-value} for the watermark detection, which measures the probability that the level of watermark strength observed in an image could happen by random chance. Rigorously, we have
\begin{equation*}
    p = \textrm{P}_m\bigl(D(\omega,m')< D(m,m') \mid H_0\bigr),
\end{equation*}
where $D(\omega,m')$ is a dissimilarity metric between an arbitrary message $\omega \sim \mathcal{M}$ (selected uniformly at random) and recovered message $m'$ from the image by the detector, and $D(m,m')$ denotes dissimilarity between the ground truth message $m$ and the recovered message $m'$.  $H_0$ denotes the null hypothesis that the image was generated without knowledge of the watermark (and therefore, the recovered message is random). The same hypothesis testing can also be applied to user identification.

As in some prior work~\citep{fernandez2023stable}, one may set a threshold on the estimated $p$-value to determine the detection result. However, this approach makes it difficult to compare different watermark methods fairly. Even if we set the same $p$-value threshold on all watermark methods, the distinct choice of message space $\mathcal{M}$, message distribution $\textrm{P}_m$, and hypothesis test may differ. Therefore, we seek to evaluate watermark methods mainly using metrics that are independent of the choice of $p$-value threshold and statistical test.

\subsection{Performance Metrics for User Identification}
\label{apd:identification_metrics}
For user identification, we also focus on metrics that do not depend on statistical testing and hyperparameters like $p$-value thresholds.

The user detection issue involving $K$ users is aptly conceptualized as a $K$-way classification task. This can be reframed into a binary classification problem by designating the positive class as the correct user and the negative class as all other users. From this perspective, the TPR@$x\%$FPR metric becomes applicable, defined for a specific FPR threshold and user count. In our study, we focus on TPR@0.1\%FPR for a scenario involving 1,000 users. The identification performance results are shown in~\Cref{apd:more_result_identification}.

\subsection{Other Performance Metrics}
While this paper primarily focuses on the TPR@0.1\%FPR metric, it's important to acknowledge other common metrics such as $p$-values, AUROC scores, mean accuracies, and bit accuracies.

However, we do not report $p$-values since their absolute values depend heavily on the chosen statistical test, making them less comparable across different watermark methods.

AUROC scores, although independent of the choice of $p$-value threshold and statistical test, have limitations used as a metric for evaluating watermark detection. 
In AI-generated image applications, labeling non-watermarked images as watermarked (false positive) are particularly detrimental. 
As a result, strict control of false positive rate (FPR) is crucial.
However, a high AUROC does not guarantee a high true positive rate (TPR) at low false positive rate (FPR) levels.

Using message distances such as bit accuracy as a metric for evaluating watermarks' performance has several limitations:\\
\textbf{(1)} Insensitivity to error distribution: bit accuracy measures the proportion of correctly identified bits in the watermark but does not account for the distribution of errors. This means it treats all errors equally, regardless of their impact or pattern. In watermarking, certain types of errors (like clustered errors) might be more detrimental than others.\\
\textbf{(2)} Lack of contextual insight: bit accuracy alone doesn't provide insights into the types of errors (false positives or false negatives). In watermark detection, understanding the nature of errors is crucial, especially in differentiating between missing a watermark and incorrectly identifying one.\\
\textbf{(3)} Threshold dependency: the effectiveness of bit accuracy is dependent on the threshold chosen for determining a bit's value. Different thresholds can yield significantly different bit accuracies, making the metric somewhat arbitrary and less reliable for comparing different watermarking schemes.\\
\textbf{(4)} Non-representation of overall system performance: bit accuracy focuses narrowly on the correctness of individual bits, neglecting the broader context of the watermarking system's performance, such as its robustness against attacks, computational efficiency, or impact on image quality.\\
\textbf{(5)} Potential misleading results in imbalanced cases: in scenarios where the watermark bits are not evenly distributed (e.g., more 0s than 1s or vice versa), bit accuracy might give a skewed view of the system's performance. It could show high accuracy even if the system is only good at detecting the majority class. For these reasons, it's often more effective to use a combination of metrics that can provide a holistic view of the watermarking system's performance, considering aspects like error distribution, false positives/negatives, and overall impact on the media.

Although these metrics are not included in the paper, they are incorporated in the benchmark software and available for future research use.

\section{Design Choices of \ours}
\subsection{Dataset Preparation}\label{app:dataset}
We utilize three datasets for the non-watermarked reference images in our evaluation: \textbf{DiffusionDB}, \textbf{MS-COCO}, and \textbf{DALL$\cdot$E3}, each comprising 5000 reference images and prompts. \textbf{DiffusionDB} represents a diverse collection from the DiffusionDB dataset~\citep{wang2022diffusiondb}, focusing on images generated from the Stable Diffusion~\citep{rombach2022high} models. \textbf{MS-COCO} is derived from the well-known Microsoft COCO detection challenge~\citep{lin2014microsoft}, featuring a wide range of everyday scenes and objects. \textbf{DALL$\cdot$E3}\footnote{The DALL$\cdot$E3 dataset is hosted at \url{https://huggingface.co/datasets/laion/dalle-3-dataset}.} includes images from the DALL$\cdot$E3 model, showcasing another popular diffusion model trained on substantially different data. These datasets provide a comprehensive range of image types and contexts, ideal for robust watermark evaluation.

The three datasets are filtered subsets of the corresponding source dataset using the same filtering algorithm. The source dataset information is listed below.
\begin{itemize}
    \item \textit{DiffusionDB}: the 2m\_random\_100k split of DiffusionDB dataset~\citep{wang2022diffusiondb}, \href{https://huggingface.co/datasets/poloclub/diffusiondb/viewer/2m_random_100k}{link}.
    \item \textit{MS-COCO}: the validation split of the 2017 Microsoft COCO detection challenge~\citep{lin2014microsoft}, \href{http://images.cocodataset.org/zips/val2017.zip}{link}.
    \item \textit{DALL$\cdot$E3}: the train split of the \textit{dalle-3-dataset} repository on HuggingFace, collected from the LAION share-dalle-3 discord channel, \href{https://huggingface.co/datasets/laion/dalle-3-dataset}{link}.
\end{itemize}

The filtering algorithm considers the following rules to subsample the 5,000 image subset:
\begin{itemize}
    \item \textit{Remove columns}: Remove irrelevant columns and only keep the reference images and prompt strings.
    \item \textit{Filter prompts}: Tokenize the prompt strings by the Open Clip's tokenizer, and filter out samples with no tokens and more than 75 tokens. This is because Stable Diffusion~\citep{rombach2022high} truncates prompts at 75 tokens~\citep{wang2022diffusiondb}.
    \item \textit{Rank images}: Rank the images by their aesthetics score, as defined by~\cite{xu2023imagereward}, in descending order. We then select the top 5,000 images, along with their corresponding prompt strings. This approach is adopted because the DiffusionDB and DALL$\cdot$E3 datasets, sourced from chat-bots, contain some lower-quality images. We posit that watermarking holds greater utility for high-quality AI-generated images, as the copyright protection of low-quality generated images is less meaningful and practical.
\end{itemize}

\begin{figure}[!htbp]
  \begin{subfigure}{0.3\textwidth}
    \centering
    \includegraphics[width=\linewidth]{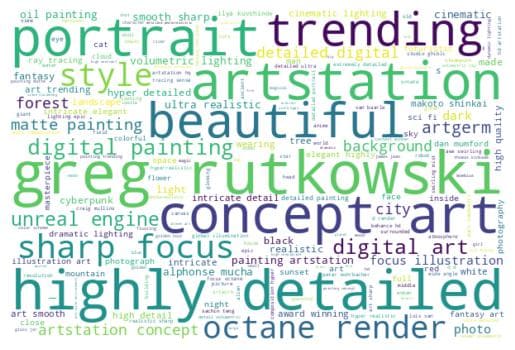}
    \caption{DiffusionDB prompts}
    \label{subfig:dataset_diffusiondb_wordcloud}
  \end{subfigure}%
  \hfill
  \begin{subfigure}{0.3\textwidth}
    \centering
    \includegraphics[width=\linewidth]{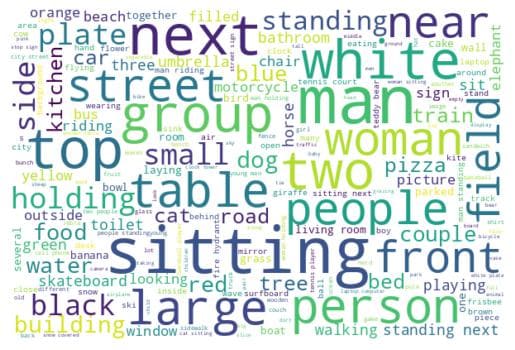}
    \caption{MS-COCO prompts}
    \label{subfig:dataset_MS-COCO_wordcloud}
  \end{subfigure}%
  \hfill
  \begin{subfigure}{0.3\textwidth}
    \centering
    \includegraphics[width=\linewidth]{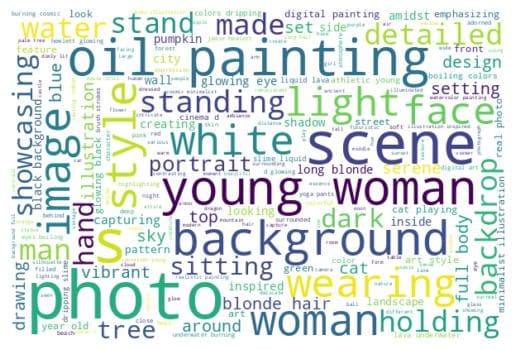}
    \caption{DALL$\cdot$E3 prompts}
    \label{subfig:dataset_dalle3_wordcloud}
  \end{subfigure}
  \caption{Word clouds of DiffusionDB, MS-COCO, and DALL$\cdot$E3 prompts.}
  \label{fig:wordclouds}
\end{figure}

In our study, we examined three distinct datasets—DiffusionDB, MS-COCO, and DALL$\cdot$E3—each characterized by a unique distribution of prompt words. As illustrated in the word-cloud plots (\Cref{fig:wordclouds}), we observe notable differences. DiffusionDB predominantly features prompt words that emphasize the desired quality of the generated images, such as ``beautiful'' and ``highly detailed.'' In contrast, MS-COCO's prompts mainly focus on describing the objects within the images. Meanwhile, DALL$\cdot$E3's prompts show a tendency towards describing aspects of fine arts.

\begin{figure}[!htbp]
\hfill
    \begin{subfigure}{0.3\textwidth}
    \centering
    \fbox{
    \includegraphics[width=\linewidth]{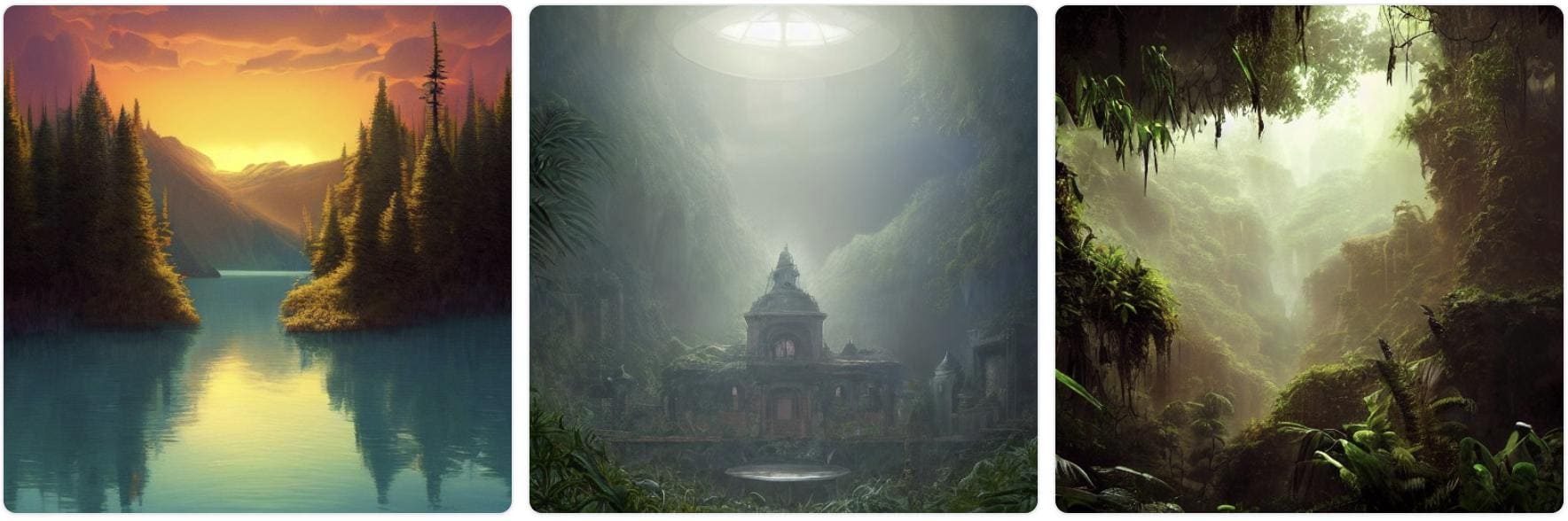}
    }
    \caption{DiffusionDB}
    \label{subfig:dataset_diffusiondb_examples}
  \end{subfigure}%
\hfill
  \begin{subfigure}{0.3\textwidth}
    \centering
    \fbox{
    \includegraphics[width=\linewidth]{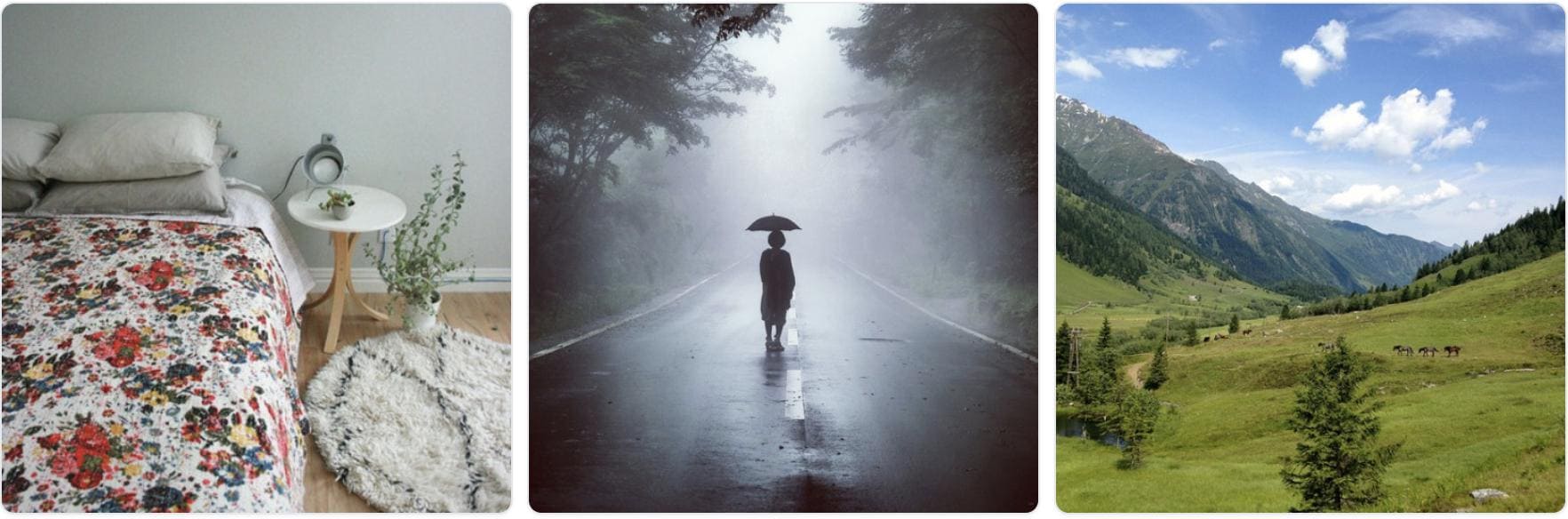}
    }
    \caption{MS-COCO}
    \label{subfig:dataset_MS-COCO_examples}
  \end{subfigure}%
\hfill
  \begin{subfigure}{0.3\textwidth}
    \centering
    \fbox{
    \includegraphics[width=\linewidth]{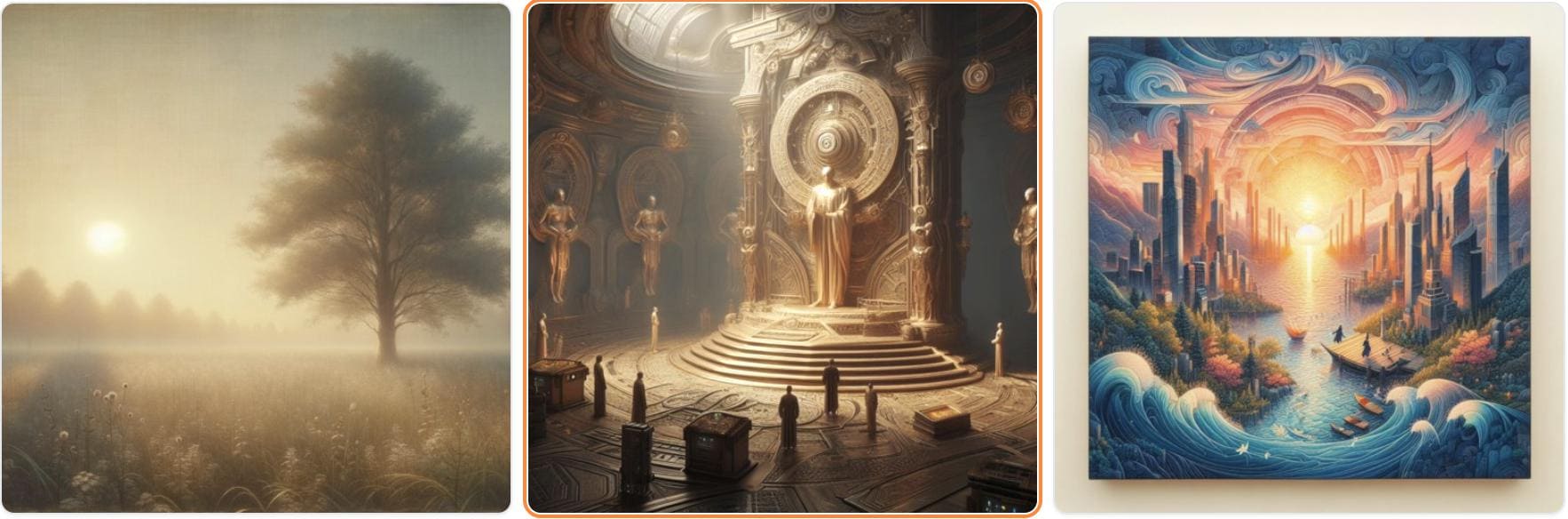}
    }
    \caption{DALL$\cdot$E3}
    \label{subfig:dataset_dalle3_examples}
  \end{subfigure}
\hfill
  \caption{Image examples of DiffusionDB, MS-COCO, and DALL$\cdot$E3.}
  \label{fig:imageexamples}
\end{figure}

Image examples from the three datasets are illustrated in~\Cref{fig:imageexamples}. The reference images for DiffusionDB are produced by Stable Diffusion, MS-COCO includes real-world photographs, and DALL$\cdot$E3 contains images generated by the DALL$\cdot$E3 model. This choice of datasets effectively covers two popular generative models and the real-world scenario, highlighting their relevance in practical watermarking applications.

\subsection{Selection of Watermark Representatives} \label{app:selection}

\begin{table}[htbp]
  \centering
  % \small
    \caption{A list of alternative watermarking algorithms not tested by \ours in this work.}
    \resizebox{\textwidth}{!}{%
  \begin{tabular}{|c|c|}
    \hline
    \textbf{Method} & \textbf{Known Weakness(es)} \\
    \hline
    DwtDct \citep{al2007combined} & Distortion \citep{wen2023tree}, Purification \citep{saberi2023robustness}  \\
    DwtDctSvd \citep{al2007combined} & Distortion \citep{zhao2023invisible,wen2023tree}, Purification \citep{saberi2023robustness}, Regeneration \citep{zhao2023invisible} \\
    RivaGan \citep{dong2023robust} & Regeneration \citep{zhao2023invisible}, Purification \citep{saberi2023robustness}  \\
    SSL \citep{fernandez2022watermarking}& Distortion\citep{zhao2023invisible}, Regeneration \citep{zhao2023invisible} \\
    WatermarkDM \citep{zhao2023recipe}& Purification \citep{saberi2023robustness}  \\
    \hline
  \end{tabular}
}
  \label{tab:other-watermarks}
  \vspace{-1em}
\end{table}

Our \ours framework can be used to stress-test the robustness of any watermark. In this work, however, we focus on three methods: the \textit{Stable Signature}, \textit{Tree-Ring}, and \textit{Stegastamp}. This is due to existing and extensive studies \citep{zhao2023invisible,saberi2023robustness,wen2023tree} indicating these three methods are far more robust to simple off-the-shelf attacks than alternative watermarking algorithms listed in \Cref{app:related}. We list these competitors along with their documented vulnerabilities in ~\Cref{tab:other-watermarks}.

\section{Evaluation Details}\label{app:eval_details}
In this section, we provide more details on the evaluation scheme of \ours.

\subsection{Watermarking Protocol and Evaluation Workflow.}\label{app:protocol_workflow}
In-depth information on the applications of invisible image watermarks is provided, focusing on AI detection and user identification. We delve into the evolution of watermarks from classical copyright protection tools to their modern uses in AI scenarios. The appendix discusses the specific roles of AI detection in distinguishing AI-created images and user identification in tracing image origins, citing studies like~\citep{saberi2023robustness, fernandez2023stable}.

The formulation of our watermarking protocol is detailed, explaining the use of an image generator $\theta_G$, a metric space of watermark messages $\mathcal{M}$, and an image domain $\mathcal{X}$. We elaborate on the variations in the choice of message space $\mathcal{M}$ across different watermark methods. For example, Tree-Ring uses random complex Gaussians, whereas Stable Signature and StegaStamp use binary strings. The implications of these choices on the flexibility and effectiveness of watermark methods are discussed.

An extensive analysis of the trade-off between watermark performance and image quality in the context of watermark attacks is provided. This includes the rationale for using Performance vs. Quality 2D plots for attack comparisons, highlighting the comprehensive perspective this offers over traditional performance-focused analyses. The methodology of our evaluation process is laid out in detail, describing how we compare watermarked images from model $\theta_G$ with a mixed set of real and AI-generated images to achieve a robust and unbiased assessment. This section also covers the specific metrics used, including TPR@0.1\%FPR and various image quality metrics, and how they are integrated into a consolidated performance vs. quality analysis.

\subsection{Performance Evaluation Metrics}\label{app:perf_metrics}
The evaluation approach in \ours addresses the challenges of using $p$-values for fair watermark method comparison. The diversity in message spaces $\mathcal{M}$, distributions $\textrm{P}_m$, and hypothesis tests can lead to biased results when traditional $p$-value thresholds are used. Our metrics, designed to be independent of these thresholds and tests, offer a balanced and thorough evaluation of watermark methods, focusing on their inherent strengths in encoding and recovering messages.

Emphasizing TPR@$x\%$FPR, particularly at the low FPR of $0.1\%$, sets \ours apart in evaluating watermark methods. This novel approach, inspired by studies like \citet{wen2023tree,fernandez2023stable}, challenges watermark methods beyond typical benchmarks such as TPR@$1\%$FPR. Applied to a broader image dataset, it provides a more comprehensive evaluation of their effectiveness. In user identification, \ours's multi-class classification approach assesses watermark methods' efficacy in correctly attributing users. The appendices detail the methodology's implementation and present additional results, demonstrating the effectiveness and accuracy of our approach in various user identification scenarios.

We treat the user identification problem as a multi-class classification task, as outlined in Section~\ref{subsec:std_eval}. This involves defining a set of ground-truth messages, each corresponding to a unique user. To avoid the exhaustive evaluation process (watermark encoding, attacking, and decoding) for varying numbers of users, we consistently watermark images with the same message, the ground-truth message of the first user, and generate a random set of ground-truth messages for the remaining users at the time of evaluation. This approach is feasible since the ground-truth messages for users other than the first do not influence the watermarking or attack phases. We conduct the identification assessment ten times with ten distinct random sets of ground-truth messages for the other users, and we report the mean multi-class classification accuracy.

\subsection{Processing Results}\label{app:process_results}
\textbf{A set of Performance vs. Quality 2D plots show the detailed evaluation results.} We evaluate 3 watermarking methods under the 26 attacks, and report results across 3 datasets in \Cref{fig:all_diffusiondb_1} to \Cref{fig:all_dalle_2}. The quality of images post-attack is evaluated using 8 metrics and the detection performance of 3 methods is measured by TPR@0.1\%FPR. 

\begin{figure}[!htbp]
    \vspace{-0.5em}
     \centering
     \includegraphics[width=0.5\textwidth]{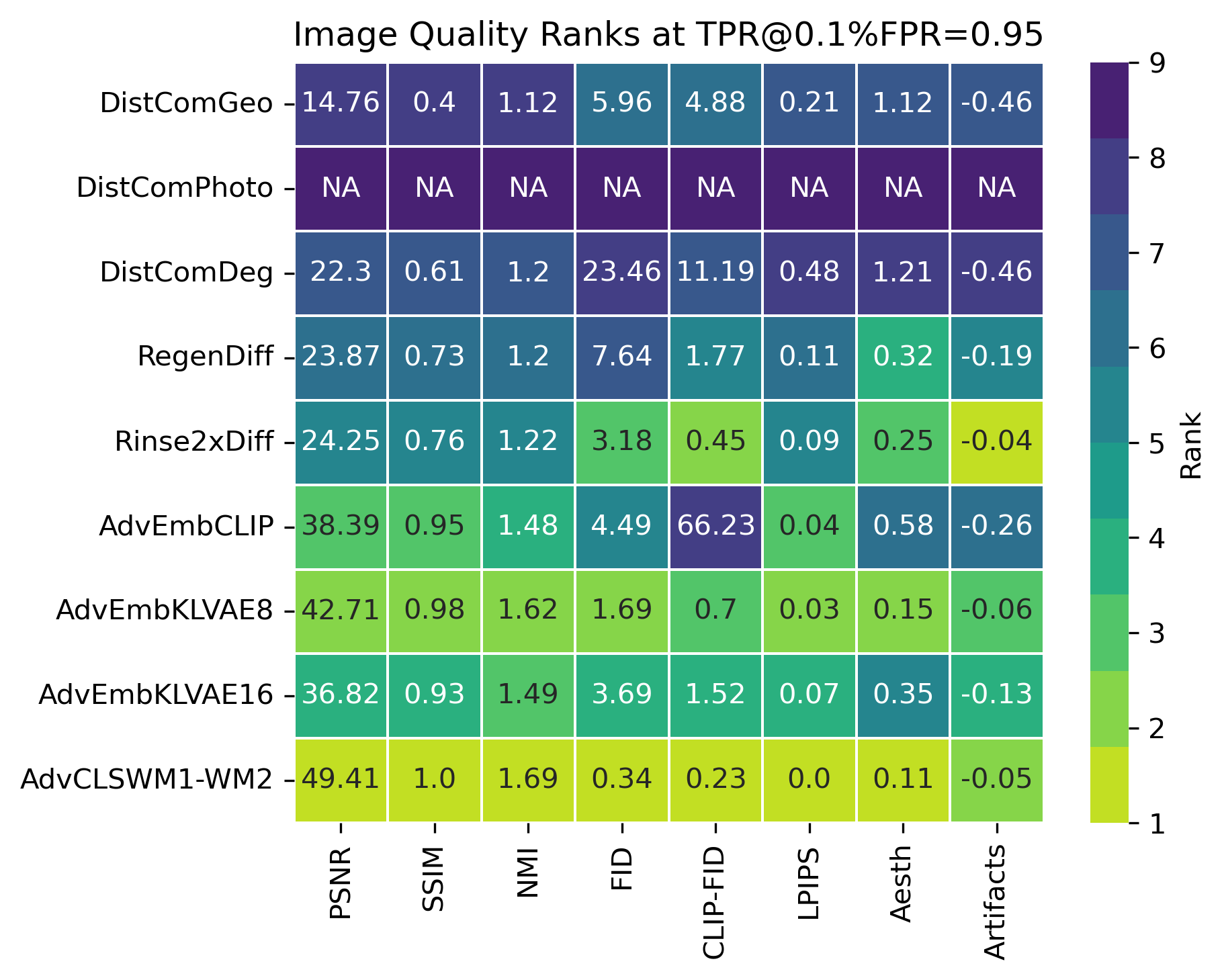}
     \caption{\textbf{Ranking attacks with different quality metrics} on DiffusionDB images watermarked by Tree-Ring. Attack potency is ranked by image quality at 0.95 TPR@0.1\%FPR. Colors indicate the ranks (1=best, 9=worst), and values show the measured quality. We use 'NA' to label an attack if its attack curve lies entirely above TPR=0.95; the attack is automatically ranked last. 
     }
     % \vspace{-2em}s
     \label{fig:tree-ring-heat}
 \end{figure}
\textbf{Different quality metrics yield similar ranking of attacks.} Despite measuring different aspects of image quality, we observe that eight quality metrics consistently produce similar rankings for attacks, as illustrated in \Cref{fig:tree-ring-heat}. 
Since a strong attack should remove the watermark without sacrificing the image quality, we rank attack potency by ranking the post-attack quality, from best to worst, at a frozen performance threshold (e.g., TPR@0.1\%FPR=0.95).
Upon comparing the rankings derived from different quality metrics, we find that the variations in rank order are minimal. 
Consequently, we aggregate these metrics into a single, unified quality metric.

\textbf{Unified Performance vs. Quality degradation 2D plots.} 
We first set the ``standardized'' 0.1 and 0.9 points for each metric according to the distribution of measured values (as depicted in \cref{fig:quality_metric_cdf_normalize_range}).
Subsequently, every metric's value is normalized to predominantly fall within the $[0.1, 0.9]$ range of the normalized quality metric (the detailed methodology is provided in Appendix~\ref{app:normal_quality}). 
We average these normalized quality scores to derive the \textit{Normalized Quality Degradation}, with lower scores indicating lesser quality degradation caused by attacks, which is preferred. 
Furthermore, we aggregate the results across three distinct datasets.
The Performance vs. Quality degradation 2D plots, as shown in \cref{fig:2d_plots}, visualize
the unified evaluation results for each watermarking method. 
We use unified Performance vs. Quality degradation 2D plots to benchmark watermarks and attacks in the following sections.

\subsection{Normalization and Aggregation of Quality Metrics}\label{app:normal_quality}
\begin{figure}
    \centering
    \includegraphics[width=\textwidth]{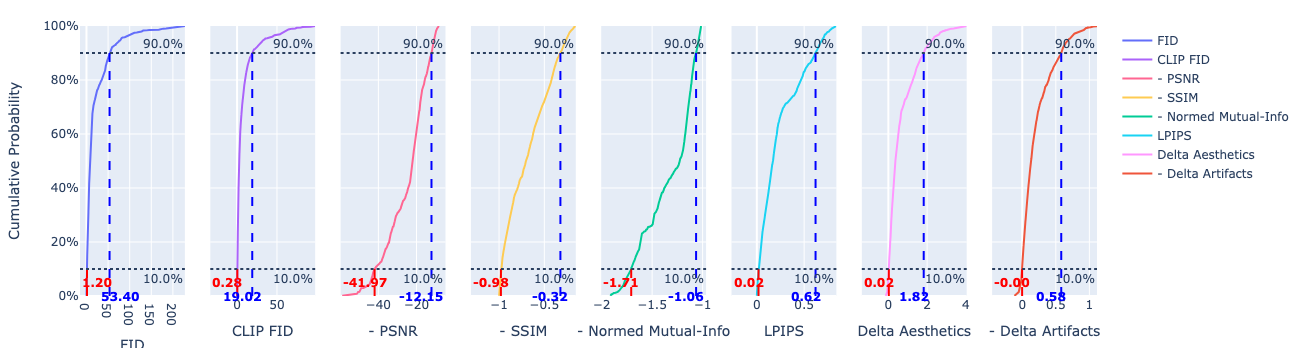}
    \caption{Cumulative distribution functions (CDFs) for eight image quality metrics across all attacked watermarked images. The horizontal dashed lines mark the 10\% and 90\% quantiles, and the intersecting vertical dashed lines delineate the bounds of the normalization intervals. Values at the lower bound are normalized to 0.1, and those at the upper bound to 0.9.}
    \label{fig:quality_metric_cdf_normalize_range}
\end{figure}

The eight quality metrics in \ours exhibit unique range characteristics. To synthesize these into a single metric, we normalize each metric into a common interval, assigning the 10\% quantile of all attacked images as the 0.1 point, and the 90\% quantile as the 0.9 point. This normalization is based on a comprehensive dataset covering 26 attack methods, three watermark methods, and three datasets. Our focus is on specific applications, particularly attacking invisible image watermarks. The normalization process is informed by the cumulative distribution functions (CDFs) of these metrics, which exhibit a roughly linear distribution between the 10\% and 90\% quantiles, but a non-linear pattern outside this range. This observation is particularly evident in metrics like PSNR. The normalization method ensures values carry equivalent significance across different metrics. Figure~\ref{fig:quality_metric_cdf_normalize_range} in this appendix provides a visual representation of the CDFs across all metrics. After normalization, metrics are aggregated by averaging to form the comprehensive quality metric, utilized in Section~\ref{sec:results} for Performance vs Quality plots, watermark radar plots, and attack leaderboards. This section elaborates on the normalization and aggregation process, providing a foundation for understanding the metric's application and significance.

In Figure~\ref{fig:quality_metric_cdf_normalize_range}, the cumulative distribution functions (CDFs) for eight image quality metrics over all attacked watermarked images are presented. This illustration includes the metric values at the 10\% and 90\% quantiles, which are used as the boundaries for normalizing the metric values within the range of $[0.1, 0.9]$. Such normalization ensures that all normalized metrics exhibit a comparable statistical distribution over attacked watermarked images, facilitating an unbiased aggregated evaluation. To consolidate these normalized metrics, we first calculate the average within each of the four defined categories (image similarities, distribution distances, perception-based metrics, and image quality assessments) as delineated in Section~\ref{subsec:std_eval}. Subsequently, the average of these category averages is calculated to yield a single, consolidated normalized, and aggregated quality metric.

\subsection{Details of Benchmarking Watermarks}\label{app:attacks_for_wm}
When benchmarking watermark robustness in \cref{fig:radar} and \cref{fig:radar_id}, we consider the following effective attacks. We select 21 attacks from 26 attacks. 
We include all distortion attacks. 
We select the two most effective single regeneration attacks and two rinsing attacks.
For adversarial attacks, we do not include AdvEmbB-RN18, and AdvCls-Real\&WM since they basically do not work. We also eliminate AdvCls-UnWM\&WM and only use AdvCls-WM1\&WM2 to represent surrogate detector attacks since AdvCls-UnWM\&WM is based on an unrealistic assumption. 
For each type of attack, we compute Average TPR@0.1\%FPR across all practical strength levels that cause quality degradation less than 0.8, and across all attacks in each category. 
\begin{itemize}
    \item \textit{Distortion Single}: Dist-Rotation, Dist-RCrop, Dist-Erase, Dist-Bright, Dist-Contrast, Dist-Blur, Dist-Noise, Dist-JPEG.
    \item \textit{Distortions Combination}: DistCom-Geo, DistCom-Photo, DistCom-Deg, DistCom-All.
    \item \textit{Regeneration Single}: Regen-Diff, Regen-KLVAE.
    \item \textit{Regeneration Rinsing}: Regen-2xDiff, Regen-4xDiff.
    \item \textit{Adv Embedding Grey-box}: AdvEmbG-KLVAE8.
    \item \textit{Adv Embedding Black-box}: AdvEmbB-CLIP, AdvEmbB-SdxlVAE, AdvEmbB-KLVAE16.
    \item \textit{Adv Surrogate Detector}: AdvCls-WM1\&WM2.
\end{itemize}

\subsection{Details of Benchmarking Attacks}\label{app:bench_attacks}
In addition to benchmarking watermarks, \ours also facilitates the analysis from the perspective of attacks.
Table~\ref{tab:leaderboard_detect} provides a leaderboard of individual attacks. 
A strong attack should result in low post-attack detection performance while simultaneously preserving image quality for practical uses.
Therefore, we benchmark attacks according to both performance and quality degradation. 
Based on three Performance vs. Quality 2D plots in \cref{fig:2d_plots}, we first select two performance thresholds, TPR@0.1\%FPR=0.95 and TPR@0.1\%FPR=0.7, ensuring intersections with most attack curves. 
Then, we calculate the quality degradation for each attack at these two performance thresholds, denoted as Q@0.95P and Q@0.7P. 
Given that some attack curves do not intersect with either threshold, we also compute each attack's average performance and quality degradation across all strengths, termed as Avg P and Avg Q. 
We report these metrics — Q@0.95P, Q@0.7P, Avg P, and Avg Q — for attack comparison. 
Based on them, we also provide a ranking of 26 attacks for each watermarking method for reference.
During this ranking process, we incorporate a 0.01 buffer for both P and Q, meaning that if the difference between any two values is less than 0.01, they are considered a tie in terms of ranking.

\section{Details of Attacks}

\input{table/attacker_know}

\subsection{Distortion Attacks}\label{app:distortion}
For single distortions, we consider, as described in~\Cref{sec:attack}, eight types: rotation, resized-crop, random erasing, brightness adjustment, contrast adjustment,  Gaussian blur, Gaussian noise, and JPEG compression. 
For each distortion, we consider five evenly distributed distortion strengths between minimum and maximum; the minimums and maximums are listed as follows.

\begin{itemize}
    \item \textit{Rotation}: rotate 9$^\circ$ to 45$^\circ$ clock-wise.
    \item \textit{Resized-crop}: crop 10\% to 50\% of the image area.
    \item \textit{Random erasing}: erase 5\% to 25\% of the image area and fill with gray color.
    \item \textit{Brightness adjustment}: increase image brightness by 20\% to 100\%.
    \item \textit{Contrast adjustment}: increase image contrast by 20\% to 100\%.
    \item \textit{Gaussian blur}: blur with kernel size from 4 to 20 pixels.
    \item \textit{Gaussian noise}: add Gaussian random noise with standard deviation from 0.02 to 0.1 (when pixel values normalized to [0, 1]).
    \item \textit{JPEG compression}: compress with JPEG quality score from 90 to 10.
\end{itemize}

It is worth noting that our strength selections are more conservative than most of the watermark papers, such as~\citep{wen2023tree,fernandez2023stable}. This is because we want to keep the image quality after distortion within a reasonable interval compared to the other attacks. While some watermark papers intentionally select unreasonably large distortion strength (for example, cropping 90\% of image area in~\citep{fernandez2023stable}, or Gaussian blurring with kernel size 40~\citep{wen2023tree}) to demonstrate their robustness under some distortions. We implement the distortions following the standard image augmentations in the \textit{torchvision} library.

For combinations of distortions (also called combo distortions in paper for short), we apply each single distortion with the same relative strength, where the relative strength is between 0 and 1, normalized with respect to the minimum and maximum strengths above. For combinations of geometric, photometric, and degradation distortions, we consider five evenly distributed normalized strengths from 0.05 to 0.45. For combinations of all distortions, we consider five evenly distributed normalized strengths from 0.05 to 0.20. The relative strengths are selected for reasonable image qualities after distortions again.

\begin{figure}[!htbp]
    \centering
    \hfil
    \begin{subfigure}{0.3\textwidth}
        \includegraphics[width=\linewidth]{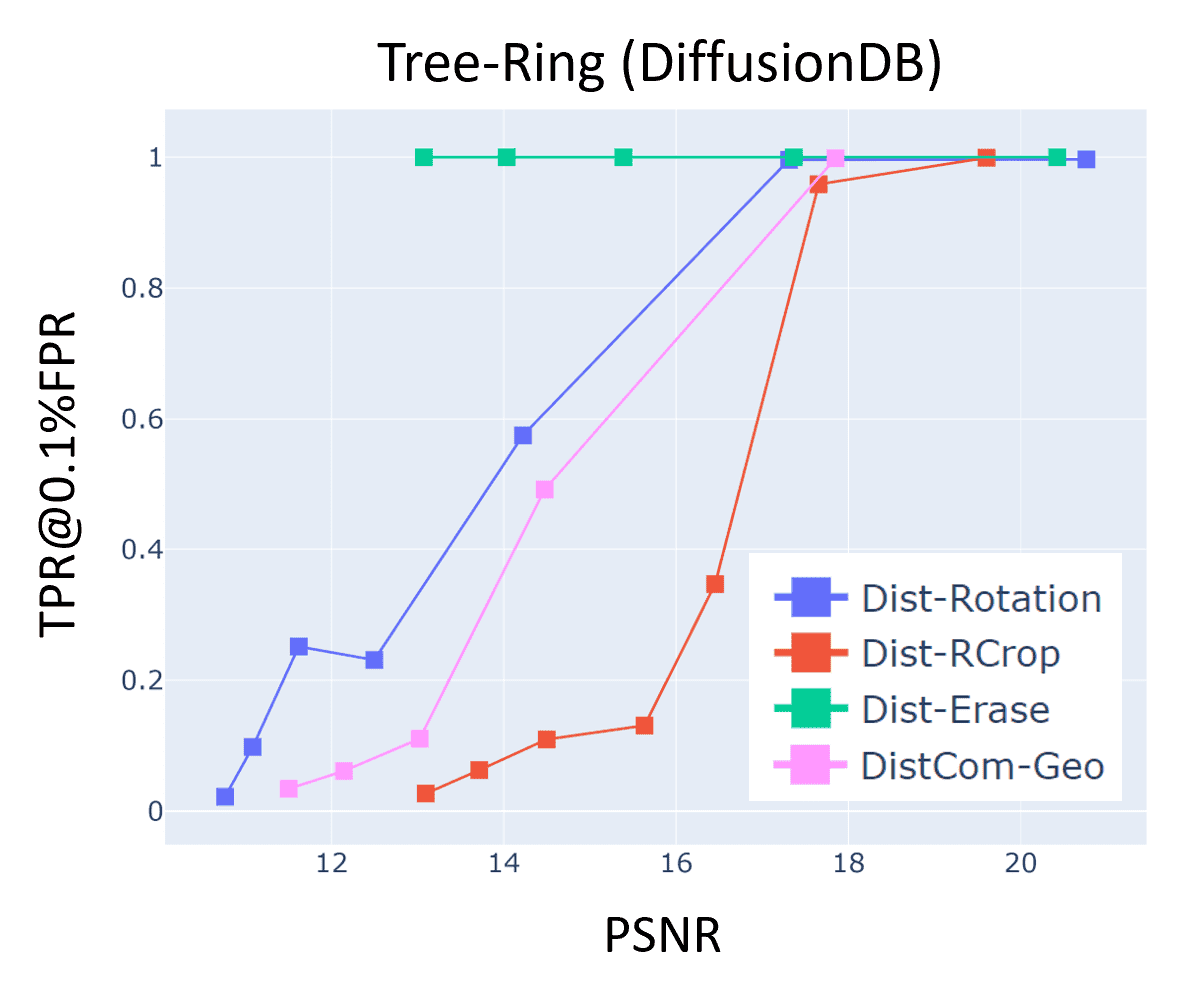}
        \caption{Geometric distortions (PSNR $\uparrow$)}
    \end{subfigure}
    \hfil
    \begin{subfigure}{0.3\textwidth}
        \includegraphics[width=\linewidth]{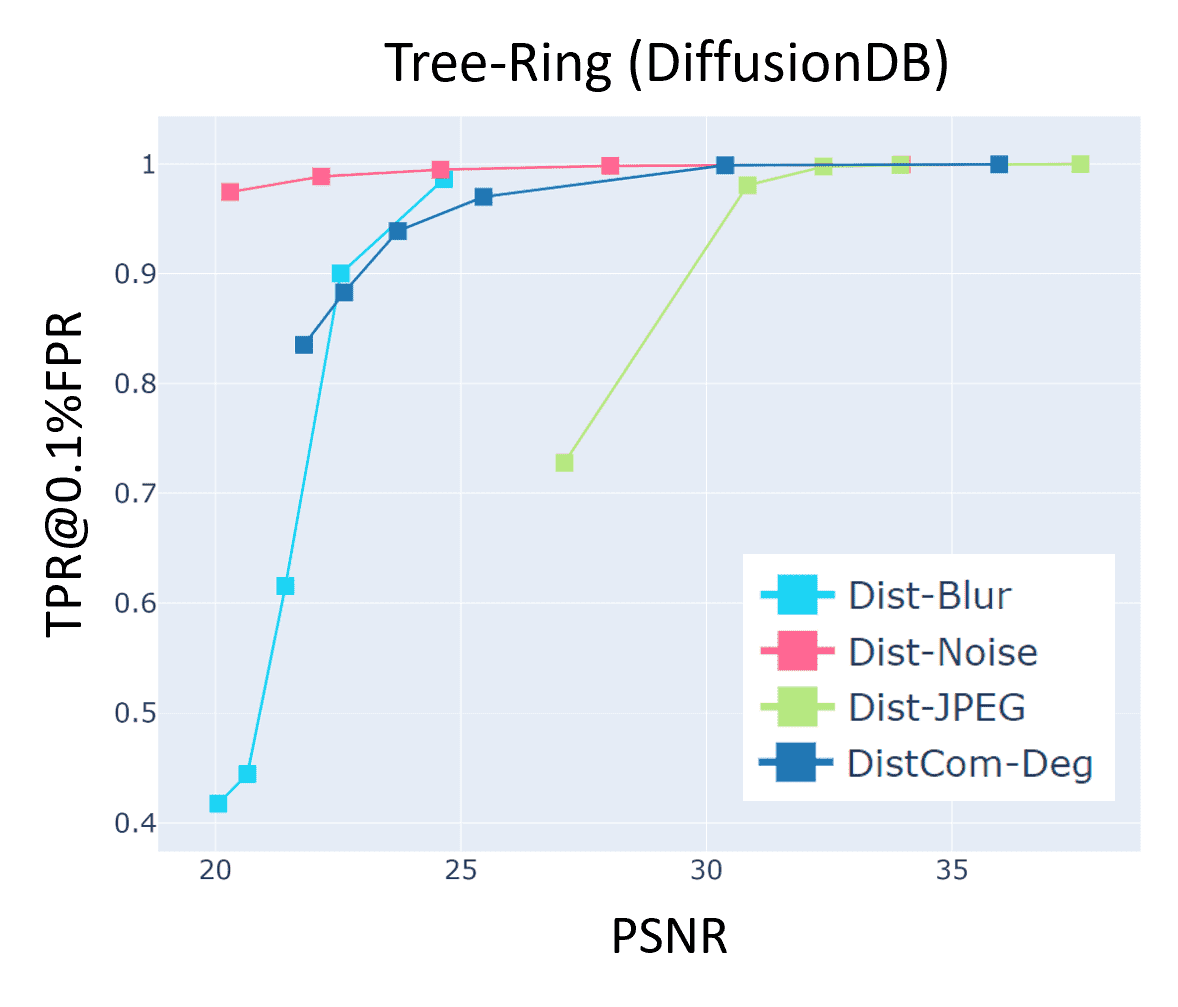}
        \caption{Degradation distortions (PSNR $\uparrow$)}
    \end{subfigure}
    \hfil
    \caption{Distortions and their combinations. We combine three types of distortions: geometric, photometric, and degradation, both individually and collectively. By comparing quality-performance plots, we see combinations of distortions do not necessarily lead to better attacks.}
    \label{fig:distortion}
\end{figure}

\subsection{Regeneration Attacks}
\label{regeneration}
Following the language of Section \ref{sec:eval-metric}, regeneration attacks \cite{zhao2023invisible} use off-the-shelf VAEs and diffusion models to transfer a target image $x\in\mathcal{X}$ to a latent representation followed by a restoration to $x'\in\mathcal{X}$ that is faithful to its original representation, i.e., $x'\approx x$. Since the chosen VAE or diffusion model will not be contained by the attacker's model of interest, the entire regeneration is likely to disrupt the latent representation of $x$, thereby damaging an embedded watermark. However, since the capacity of the attacker's regenerative model is inferior to the target model, $x'$ will likely be of reduced quality. In this work, the target model is Stable Diffusion v2.1 while the surrogate model used for regeneration is Stable Diffusion v1.4. 

Figure \ref{multi-regen} demonstrates that a long diffusion or low-quality VAE attack will significantly reduce watermark detectability but at the expense of reduced image quality, which is clear by visual inspection of the sequence of images in Figure \ref{fig:regen_diffusion_three-subfigures}. Rising regenerations achieve similar reductions in detection, although too deep of rinsing regenerations ($>30$ noising steps) significantly alter image quality as evidenced by Figure \ref{fig:4x_regen_diffusion_three-subfigures}. 
\input{figures/regen_diffusion}
\input{figures/4x_regen_diffusion}

\subsubsection{Prompted Regeneration}
We propose a simple variation on a regenerative diffusion attack: if an image is produced via a known prompt, then an attacker uses the prompt to guide the diffusion of their surrogate model. This type of attack is reasonable and realistic for users of online generative models such as DALL$\cdot$E or Midjourney. Figure \ref{multi-regen} and Tables \ref{tab:leaderboard_ident} \& \ref{tab:leaderboard_detect} indicate that this type of attack, labeled Regen-DiffP is slightly stronger than conventional Regen-Diff. 

\subsubsection{Mixed Regeneration}
\begin{figure}[!htbp]
  \centering
  \subcaptionbox{Unattacked}{\includegraphics[width=0.3\linewidth]{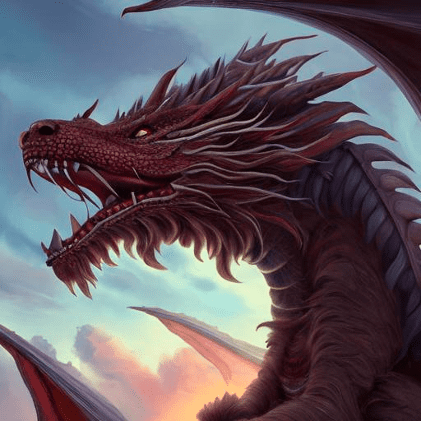}}
  \subcaptionbox{Rinse-3xDiff}{\includegraphics[width=0.3\linewidth]{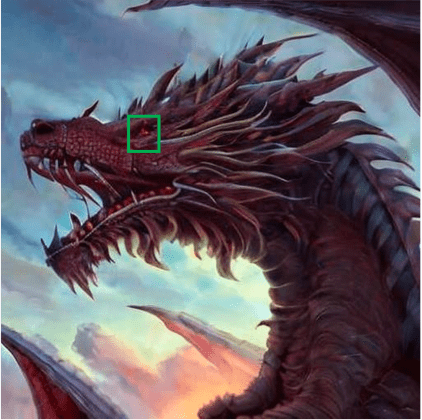}}
  \subcaptionbox{Rinse-3xDiff+VAE}
  {\includegraphics[width=0.3\linewidth]{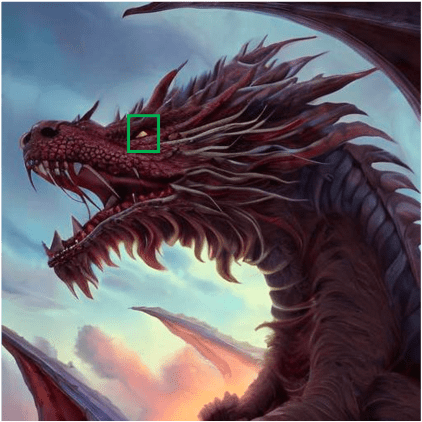}}
  \vspace{-0.5em}
  \caption{An image of a dragon attacked using a 3x rinsing regeneration. Pushing the image through a VAE restores image quality, noticeable in the eye color of the dragon (indicated by the green box). Image is drawn from the Gustavosta Stable Diffusion dataset available @ \url{https://huggingface.co/datasets/Gustavosta/Stable-Diffusion-Prompts}.} 
  \label{dragon-attack}
\end{figure}

Mixed regeneration refers to any style of attack that uses a regenerative diffusion on an image followed by VAE-style regeneration for the purposes of denoising. In Figure \ref{multi-regen}, we label examples of such attacks as RinseD-VAE and RegenD-KLVAE, which respectively denote VAE and KLVAE denoising following a 4x rinsing regeneration with 50 steps (Rinse-4xDiff-50). According to Figure \ref{multi-regen}, such a combination improves PSNR and CLIP-FID, as opposed to a Rinse-4xDiff alone. The restorative effects of mixed regeneration are visually observable for shallower (i.e., 2x or 3x) rinsing regenerations, as depicted in Figure \ref{dragon-attack}. We do not extensively study or rank such attacks in this work, but include them as a future topic of research. 

All tested regeneration attacks are summarized as follows, with five evenly divided strengths between the listed minimum and maximum unless specified otherwise:
\begin{itemize}
    \item \textit{Regeneration via diffusion}: passes an image through Stable Diffusion v1.4 with strength as the number of noise/de-noising steps timesteps, 40 to 200.
    \item \textit{Regeneration via prompted diffusion}: passes an image through Stable Diffusion v1.4 conditioned on its generative prompt with strength as the number of noise/de-noising steps timesteps, 40 to 200. 
    \item \textit{Regeneration via VAE}: Image is encoded then decoded by a pre-trained VAE (bmshj2018) \cite{DBLP:conf/iclr/BalleMSHJ18} with strength as quality level from 1 to 7.
    \item \textit{Regeneration via KL-VAE}: Image is encoded and then decoded by a pre-trained KL-regularized autoencoder with strength as bottleneck sizes 4, 8, 16, or 32.
    \item \textit{Rinsing generation 2x}: an image is noised then de-noised by Stable Diffusion v1.4 two times with strength as number of timesteps, 20-100 (per diffusion). 
    \item \textit{Rinsing generation 4x}: an image is noised then de-noised by Stable Diffusion v1.4 two times with strength as number of timesteps, 10-50 (per diffusion). 
    \item \textit{Mixed Regeneration via VAE}: an image passed through a rinsing regeneration 4x (for 50 timesteps each) and then a VAE with strength as quality level from 1-7.
    \item \textit{Mixed Regeneration via KL-VAE}: an image passed through a rinsing regeneration 4x (for 50 timesteps each) and then a KL-VAE with strength as bottleneck sizes 4, 8, 16, or 32.
\end{itemize}

\subsection{Adversarial Attacks}
\label{adv_attacks}

\subsubsection{Embedding Attack} \label{app:adv_emb}
The embedding attacks use off-the-shelf encoders and perform untargeted attacks. 
We use the Projected Gradient Descent (PGD) algorithm \citep{madry2017towards} to optimize the adversarial examples. 
We conduct the attack using a range of perturbation budgets $\epsilon$, specifically \{2/255, 4/255, 6/255, 8/255\}. 
All the attacks are configured with a step size of $\alpha=0.05*\epsilon$ and the number of total iterations of 200. 
The attacks are on the watermarked images, aiming to remove the watermarks by perturbing their latent representations.

\subsubsection{Surrogate Detector Attack} \label{app:adv_cls}
Figure~\ref{fig:illu_adv} illustrates the three settings of training the surrogate detectors. 
In all three settings, we train the surrogate detectors by fine-tuning the ResNet18\footnote{https://pytorch.org/vision/main/models/generated/torchvision.models.resnet18.html} for 10 epochs with a learning rate of 0.001 and a batch size of 128. 
The training images are either generated by the victim generator with the ImageNet text prompts "A photo of a \{\textit{ImageNet class name}\}," or real ImageNet images.
We randomly shuffle those images and build the binary training set according to each setting.
In the AdvCls-UnWM\&WM setting, we train the surrogate detector with 3000 images (1500 images per class) since we find a larger training set might have the overfitting problem. 
In the AdvCls-Real\&WM and AdvCls-WM1\&WM2 settings, we train the surrogate detector with 15000 images (7500 images per class). 
The watermarked images in AdvCls-WM1\&WM2 are embedded with two distinct messages. One message is the one used in the test watermarked images. The other one is randomly generated.
In all three settings, we use 5000 images (2500 images per class) for validation (derived from the same source as the training set), and the training yields nearly 100\% validation accuracy in all cases.

After completing the training phase, the adversary executes a Projected Gradient Descent (PGD) attack on the surrogate detector using the testing data (DiffusionDB, MS-COCO, DALL$\cdot$E3). 
In all three settings, we conduct the attack using a range of perturbation budgets $\epsilon$, specifically \{2/255, 4/255, 6/255, 8/255\}. The attack is configured with a step size of $\alpha=0.01*\epsilon$ and the number of total iterations of 50. By flipping the label, the adversary can either try to remove the watermarks or add the watermarks. The analyses of results appear in Appendix~\ref{app:adv_results}.

\section{Additional Results}
\subsection{More Results for Identification}\label{app:more_result_identification}
\begin{figure*}[!htbp]
    \centering
    \includegraphics[width=\textwidth]{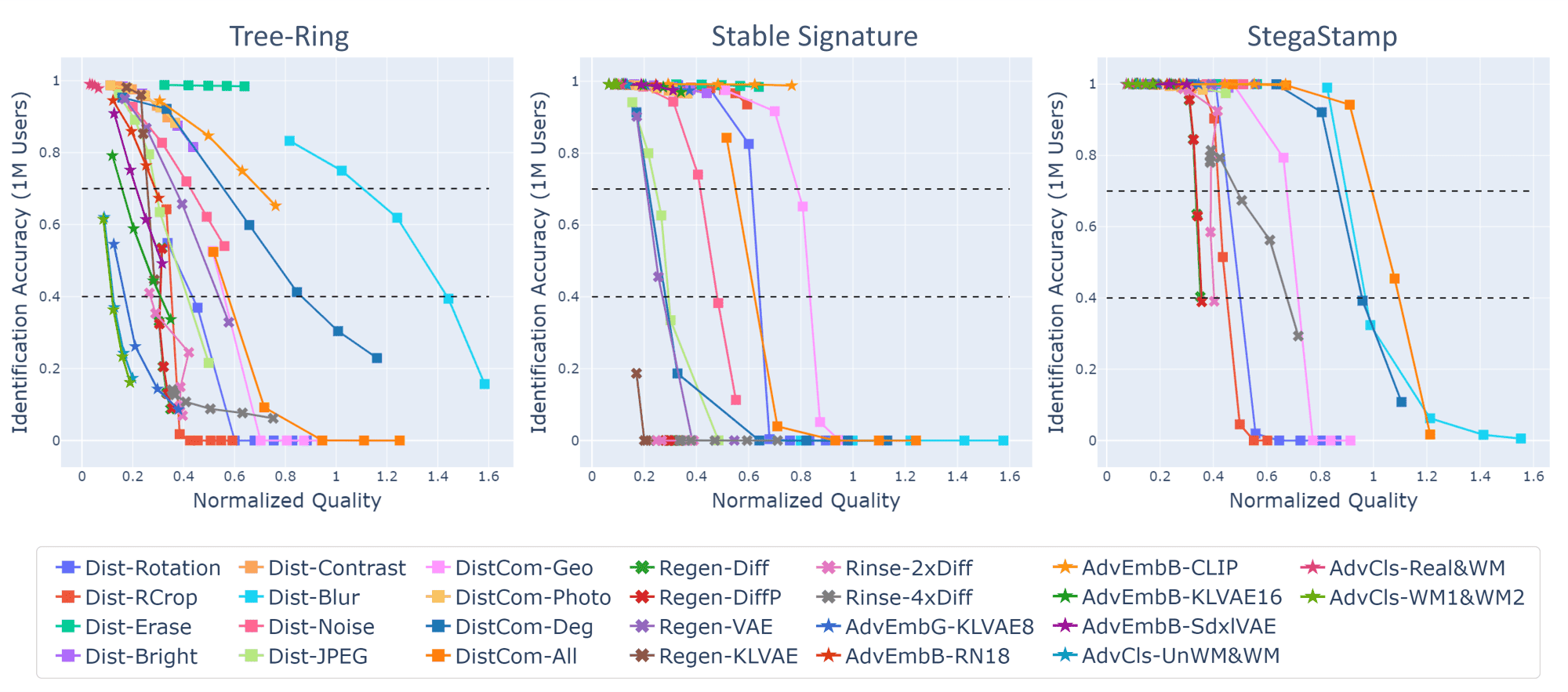}
    \caption{\textbf{Aggregated performance vs. quality degradation 2D plots under identification setup (one million users).} We evaluate each watermarking method under various attacks. Two dashed lines show to thresholds used for ranking attacks.}
    \label{fig:2d_plots_ident}
\end{figure*}

\cref{fig:2d_plots_ident} shows the Performance vs. Quality degradation plots under the user identification setting. Table~\ref{tab:leaderboard_ident} presents the ranking of attacks in the identification setup. 
\input{table/leaderboard_ident}

\subsection{More Analyses on Surrogate Detector Attacks}\label{app:adv_results}

\begin{figure}
\centering
    \includegraphics[width=0.4\linewidth]{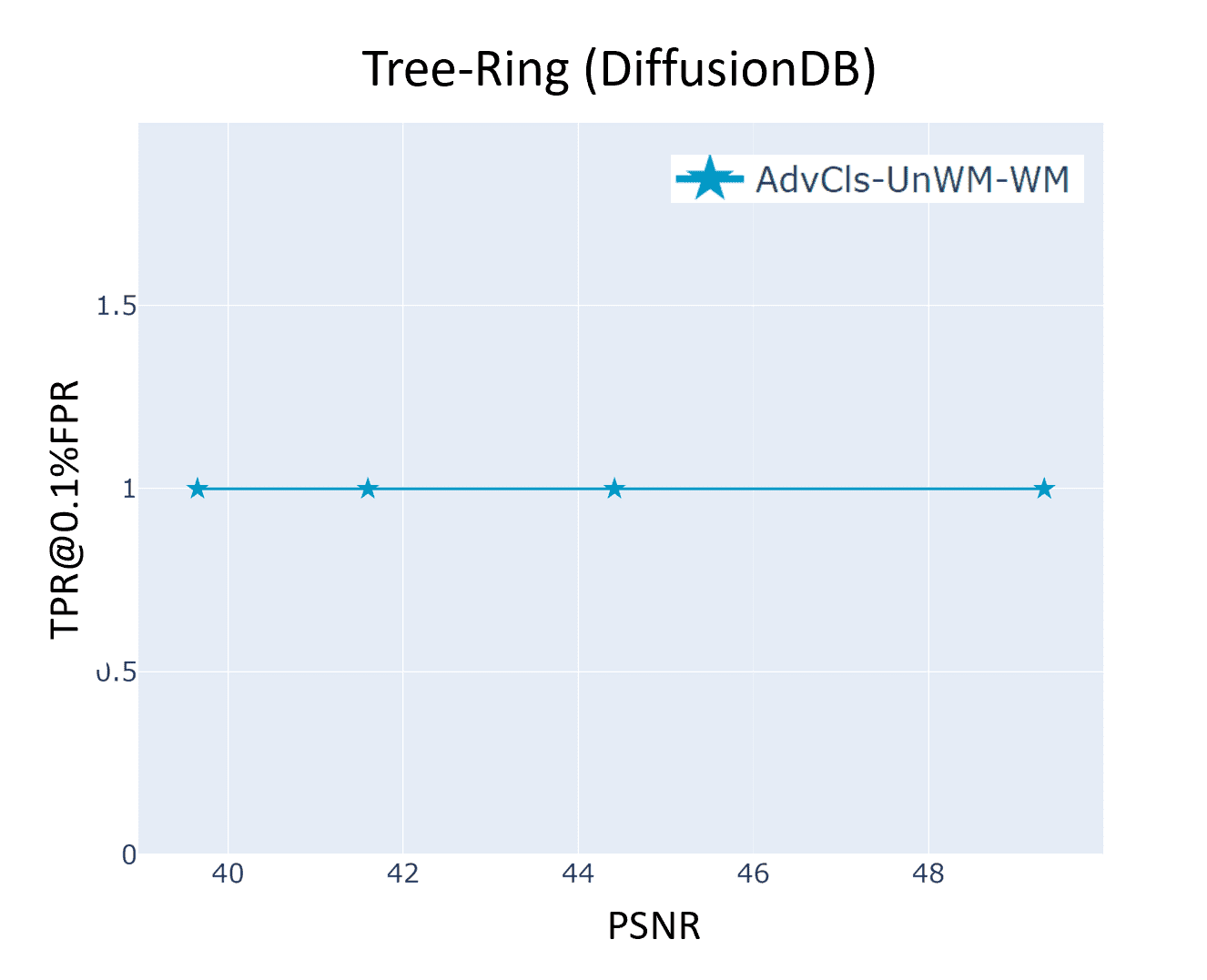}
    \vspace{-1em}
    \caption{The spoofing attack fails for AdvCls-UnWM\&WM. }
    \vspace{-1em}
    \label{fig:adv_spoof}
\end{figure}

The AdvCls-UnWM\&WM attack leverages a surrogate model to distinguish between images that are watermarked and those that are not. 
As demonstrated in Figure~\ref{fig:adv_su}, the PGD attack is effective in removing watermarks by flipping the label of watermarked images. 
This raises a question: Is it possible to similarly `add' watermarks to clean images by flipping their labels? 
This process, commonly referred to as a spoofing attack, which demonstrates a false detection of watermarks in clean images, is explored in our study.

\begin{figure}
    \centering
    \includegraphics[width=0.6\textwidth]{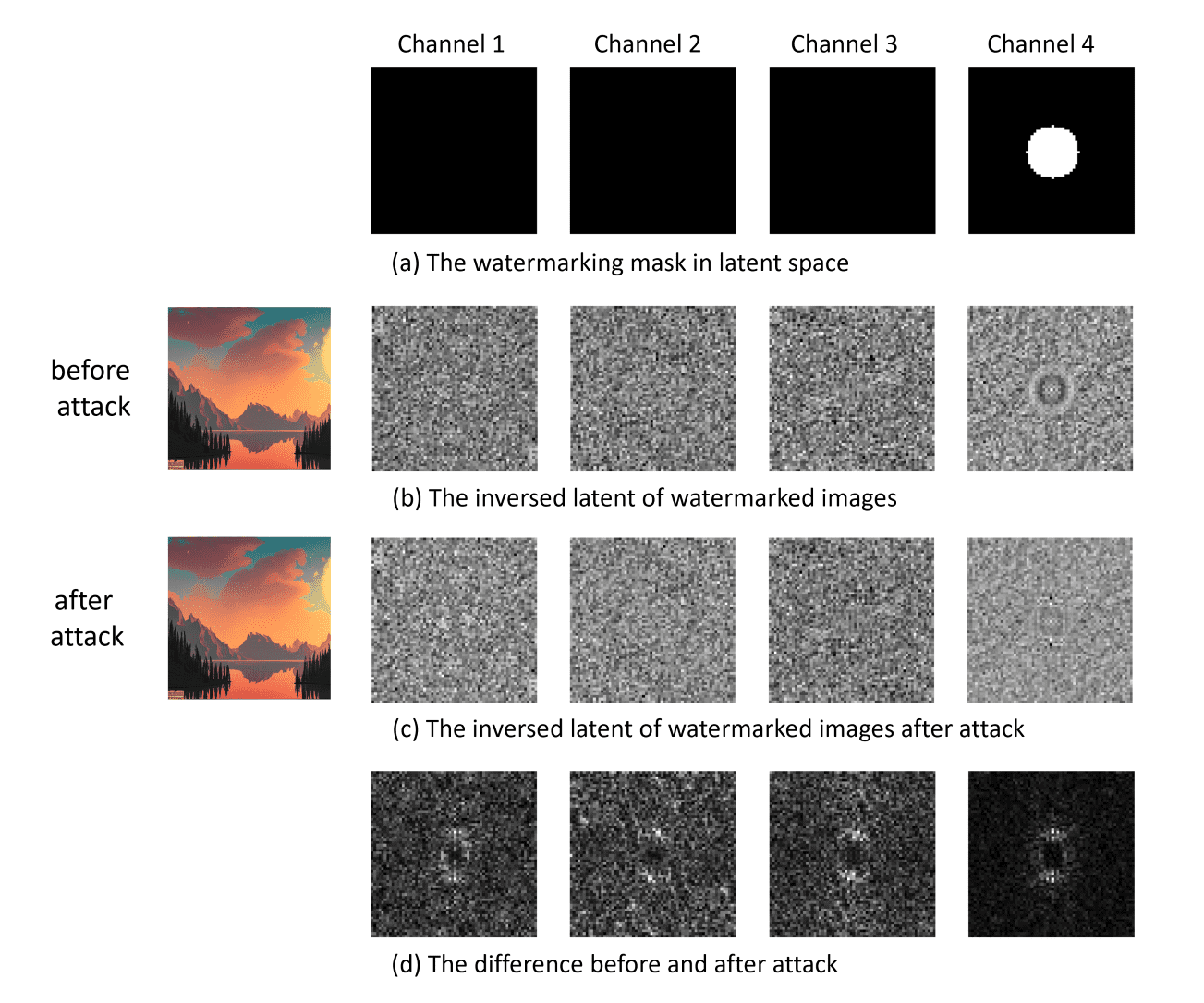}
    \vspace{-1em}\caption{Visualization of AdvCls-UnWM\&WM attack. (a) shows the watermarking mask of Tree-Ring where there are four channels, and we only watermark the last channel. The watermark message is the rings, which contain ten complex numbers that are not shown in the figure. (b) and (c) show the inversed latent before and after the attack in the Fourier space. We only show the real part of the latent. Clearly, the rings exist before the attack and vanish after the attack. (d) shows the magnitude of the element-wise difference before and after the attack. The attack not only perturbs the watermark part but also other features. The average magnitude change of the watermark-part and non-watermark-part is around 2:1. The attack successfully disturbs the watermark, albeit in an imprecise manner. }
    \label{fig:visual_unwm_wm}
\end{figure}

However, as illustrated in \Cref{fig:adv_spoof}, our attempts to add watermarks to clean images by simply flipping the labels were unsuccessful. 
In this experiment, detailed in \Cref{fig:adv_spoof}, we focus exclusively on unwatermarked images, aiming to introduce watermarks, while leaving already watermarked images untouched. 
Despite employing the most intensive perturbations, we were unable to artificially add watermarks to these images.
This outcome leads to an intriguing inquiry: Why is the technique effective in removing watermarks but not in adding them? We delve into the underlying reasons for this asymmetry in \Cref{fig:visual_unwm_wm}.

The insights from \Cref{fig:visual_unwm_wm}
reveal that the surrogate model does not exactly remove the watermark. 
Instead, it perturbs the watermark along with other features within the latent space.
The disturbance alone is sufficient to confuse the detector, making it challenging to recognize the watermark.
In contrast, successfully adding watermarks requires precise modifications in the latent space, rather than mere perturbations, which proves to be a more challenging task.
The relative imprecision of this attack may stem from the `transferable gap' between the surrogate model and the ground-truth detector. 
Notably, for the purpose of watermark removal, perturbing the latent space proves to be adequately effective.

\begin{figure}
    \centering
    \includegraphics[width=0.6\textwidth]{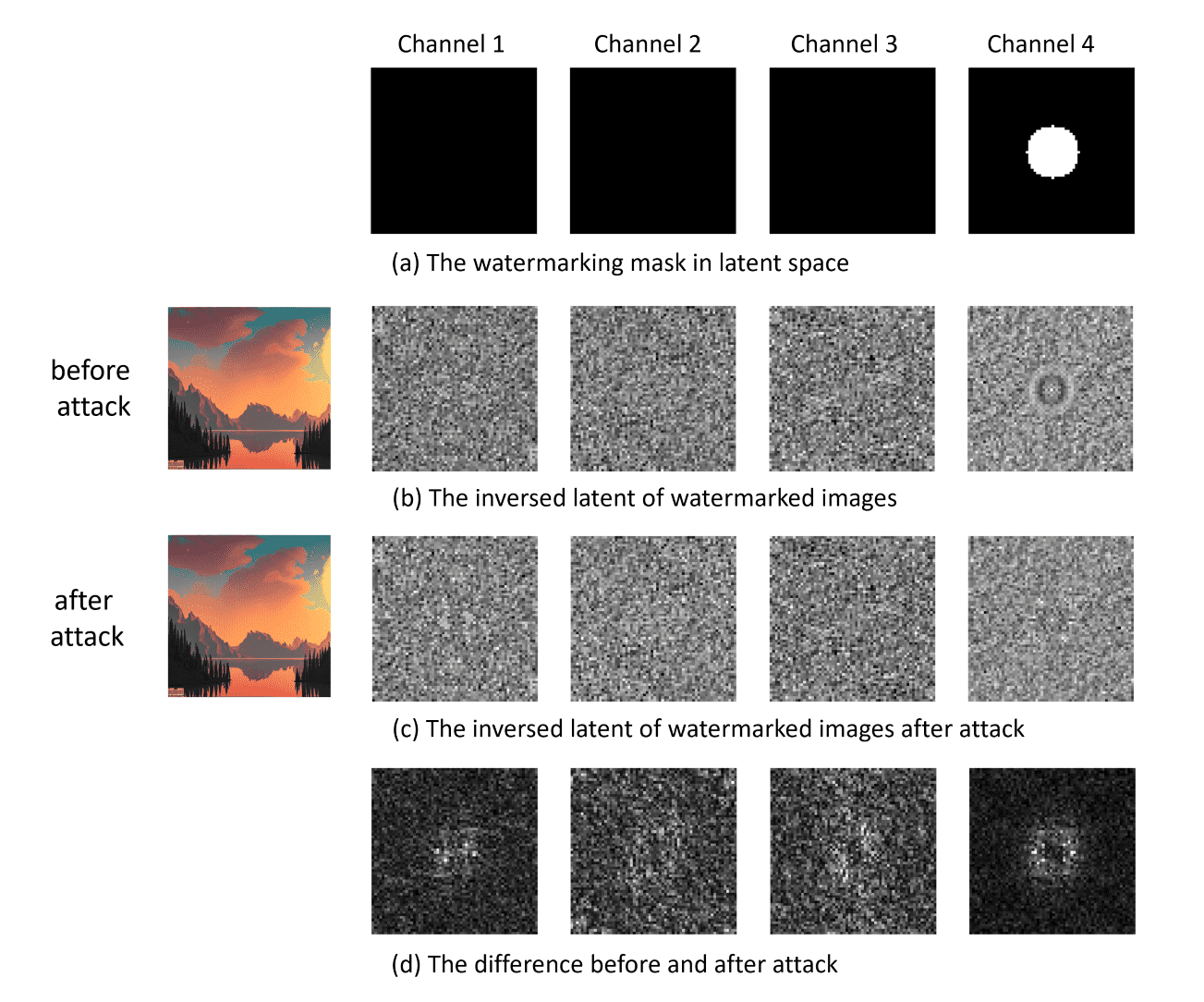}
   \vspace{-1em} \caption{Visualization of AdvCls-WM1\&WM2 attack. (a) and (b) are the same as that in Figure~\ref{fig:visual_unwm_wm}. (c) shows the inversed latent after the attack, where the watermark vanishes instead of changing to another watermark. (d) shows the magnitude of the element-wise difference before and after the attack. The attack not only perturbs the watermark part but also other features. The average magnitude change of the watermark-part and non-watermark-part is also around 2:1. Although the surrogate detector is trained to classify two different watermark messages. The attack based on it cannot change the watermark message from one to another but can effectively disturb the watermark.}
    \label{fig:visual_wm1_wm2}
\end{figure}
\begin{figure}[!hbtp]
    \centering
    \includegraphics[width=0.7\textwidth]{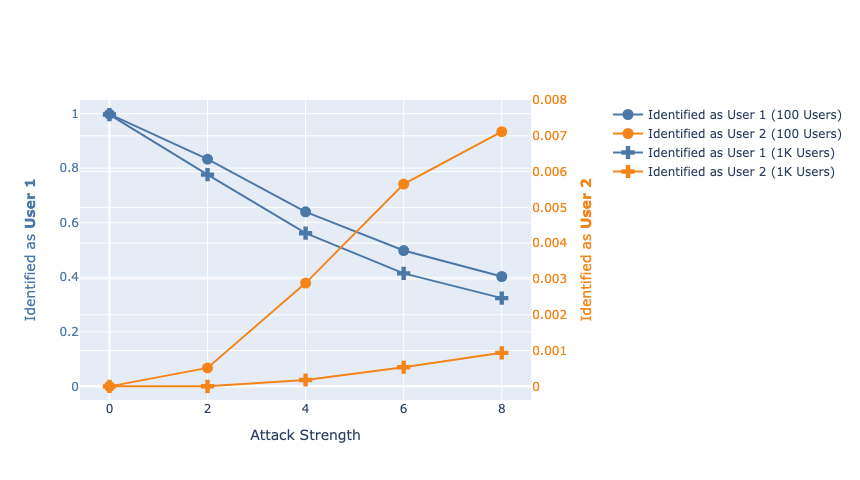}
    \caption{The user identification results for Tree-Ring under AdvCls-WM1\&WM2 attacks. The original watermarked images are embedded with User1's message. AdvCls-WM1\&WM2 tries to disrupt the latent feature of those images so that they can be misidentified as User2 generated. 
    We simulate two settings: 100 users and 1000 users in total. 
    The blue curves represent the proportion of images correctly identified as belonging to User1, while the orange curves show those misidentified as User2's.
    Note that, the axes for blue and orange curves have different ranges in the figure.
    With increasing attack strengths, the likelihood of correctly identifying them as User1's decreases significantly under both 100 and 1K user scenarios. However, misidentification as User2's images occurs notably only when the total number of users is small (e.g., 100 users).}
    \label{fig:adv_ident}
\end{figure}
\begin{figure}[!htbp]
    \centering
    \includegraphics[width=\textwidth]{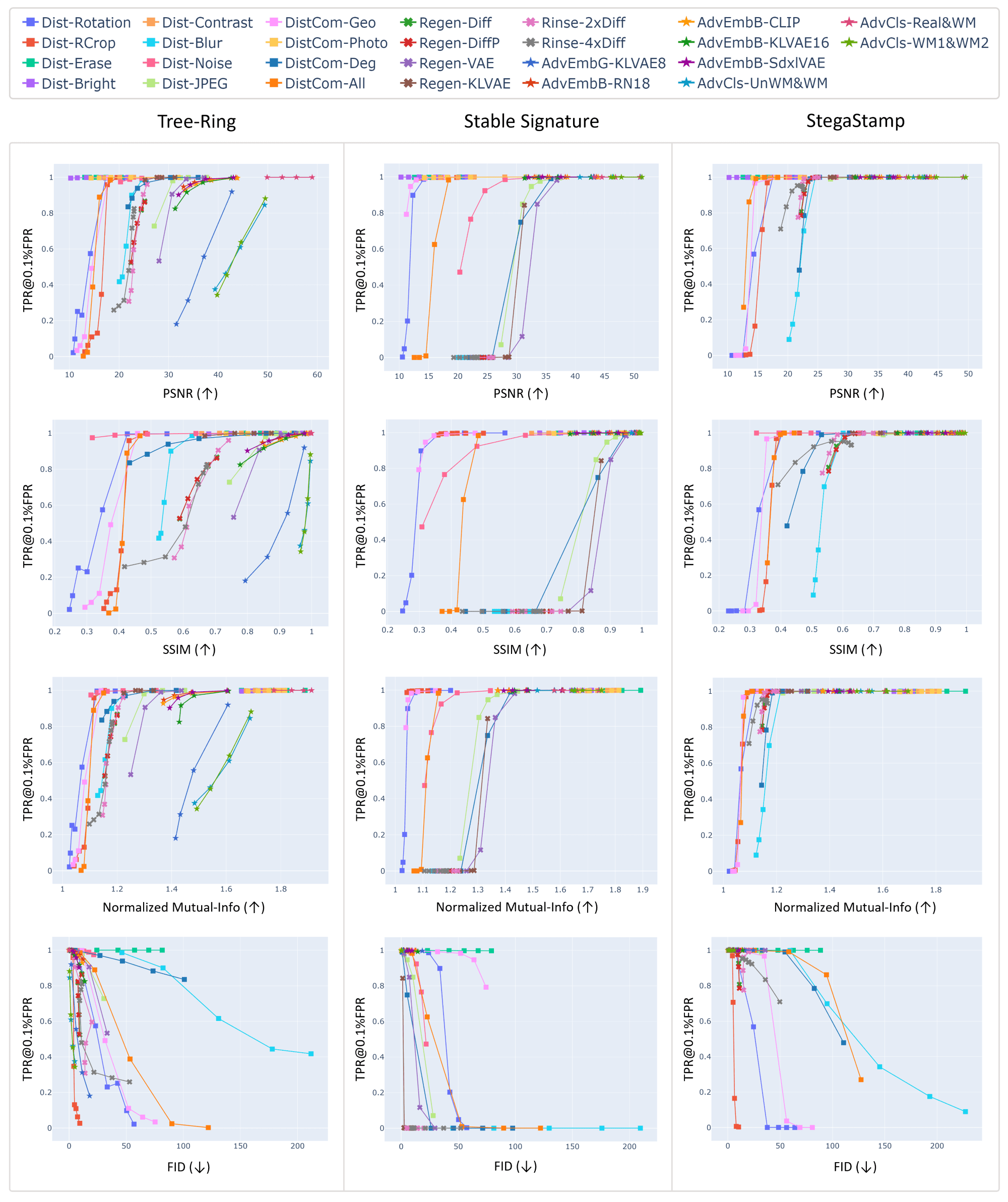}
    \caption{Evaluation on DiffusionDB dataset under the detection setup (part 1).}
    \label{fig:all_diffusiondb_1}
\end{figure}
\begin{figure*}[!htbp]
    \centering
    \includegraphics[width=\textwidth]{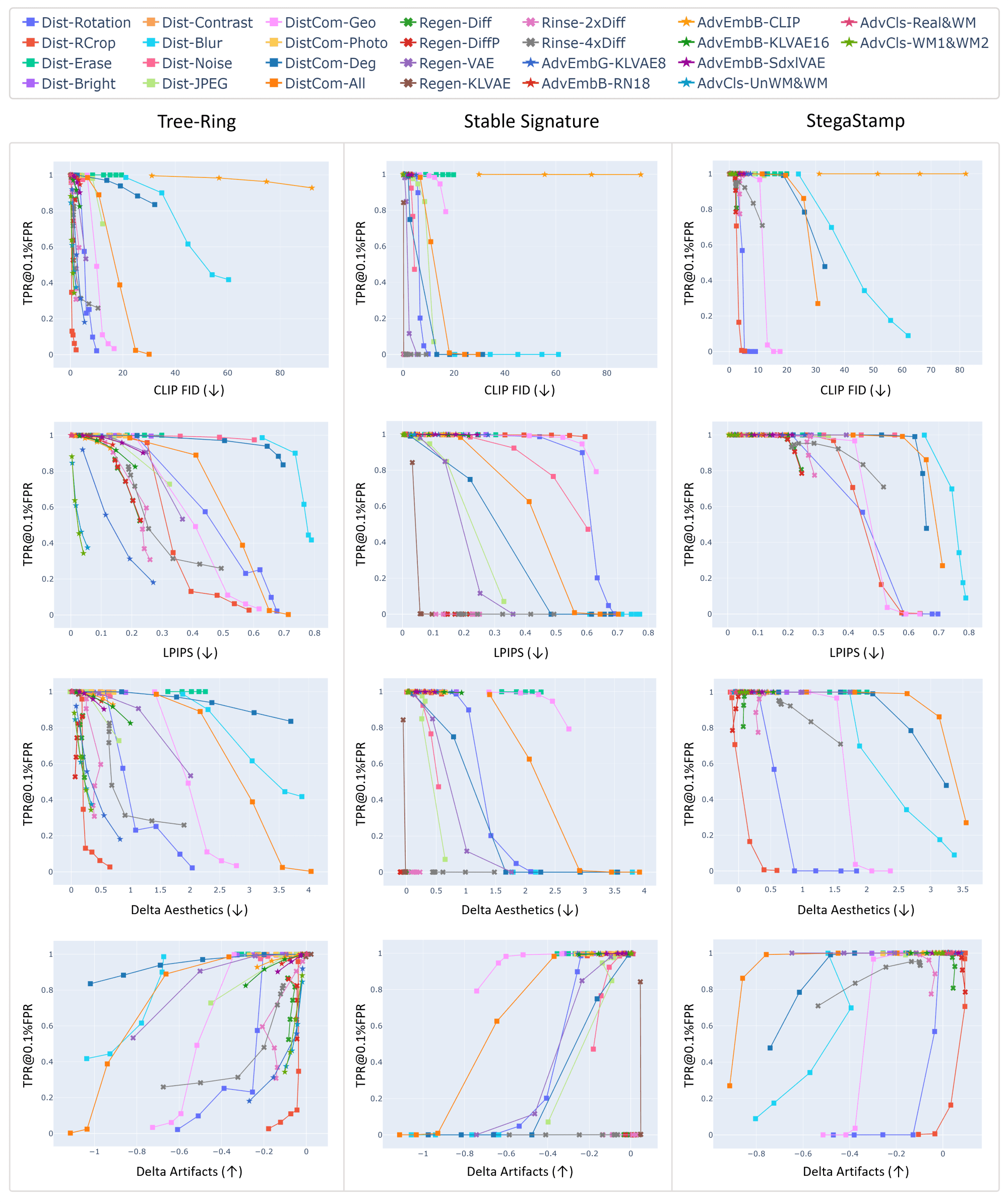}
    \caption{Evaluation on DiffusionDB dataset under the detection setup (part 2).}
    \label{fig:all_diffusiondb_2}
\end{figure*}
\begin{figure}[!htbp]
    \centering
    \includegraphics[width=\textwidth]{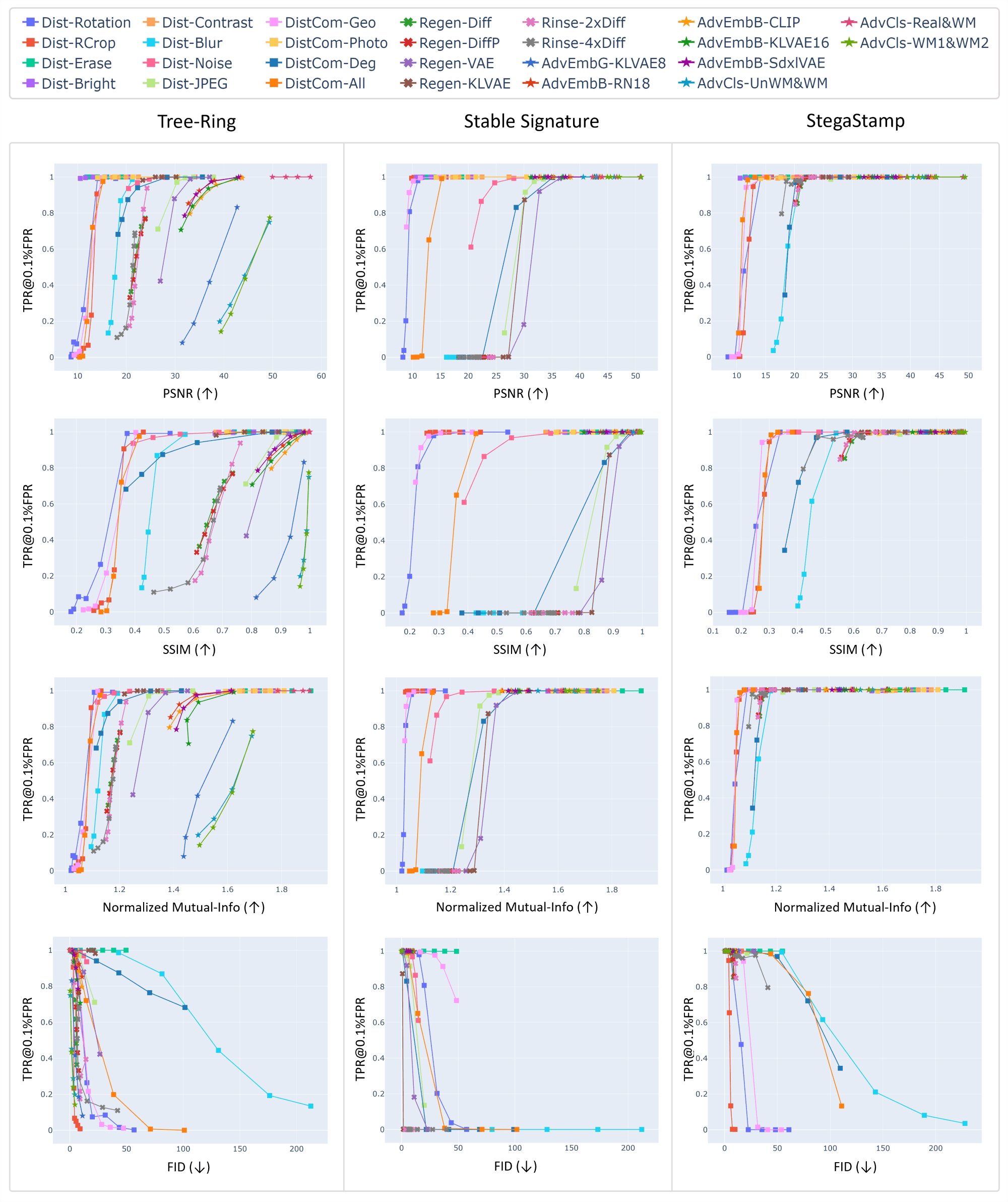}
    \caption{Evaluation on MS-COCO dataset under the detection setup (part 1).}
    \label{fig:all_coco_1}
\end{figure}
\begin{figure*}[!htbp]
    \centering
    \includegraphics[width=\textwidth]{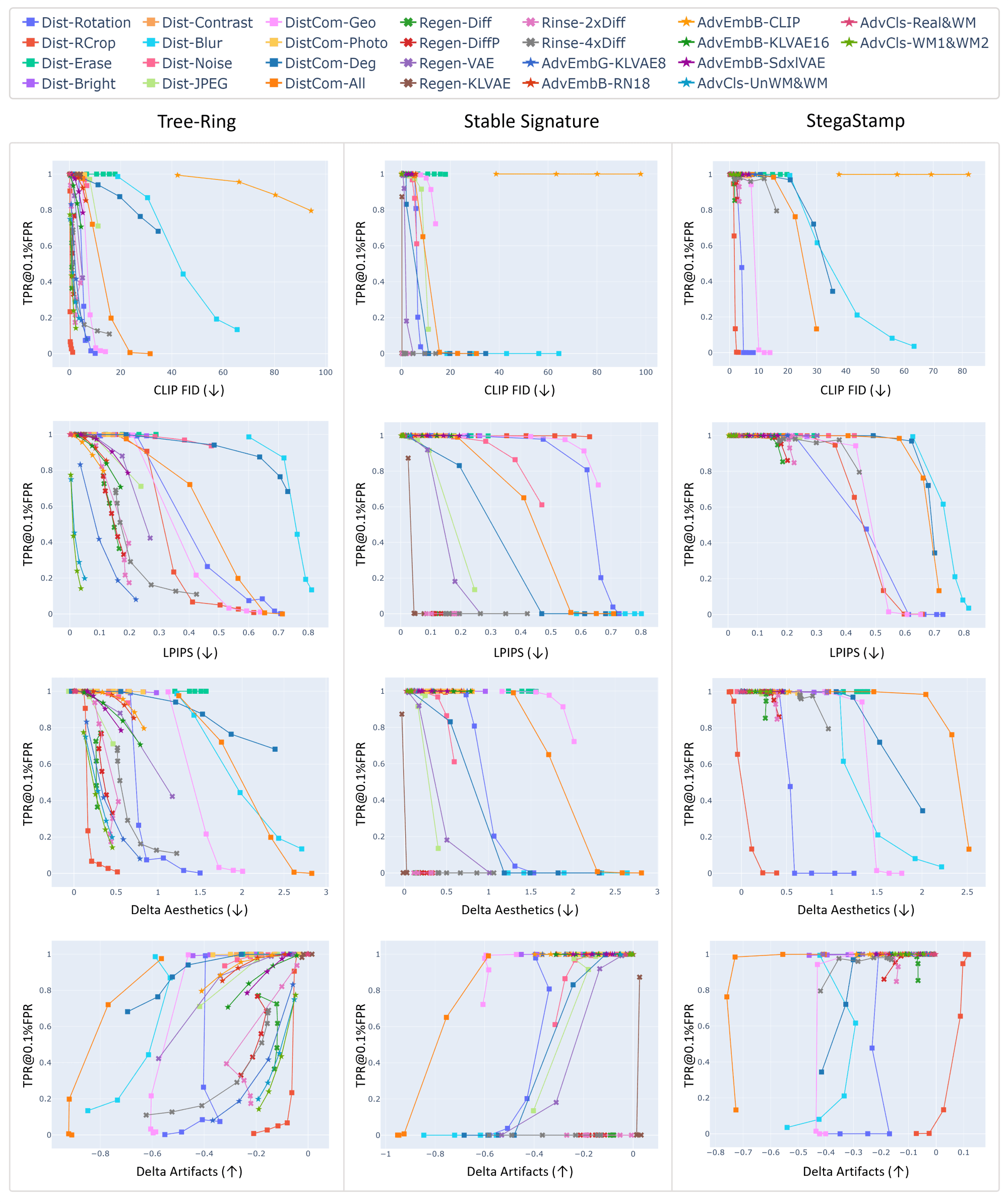}
    \caption{Evaluation on MS-COCO dataset under the detection setup (part 2).}
    \label{fig:all_coco_2}
\end{figure*}
\begin{figure}[!htbp]
    \centering
    \includegraphics[width=\textwidth]{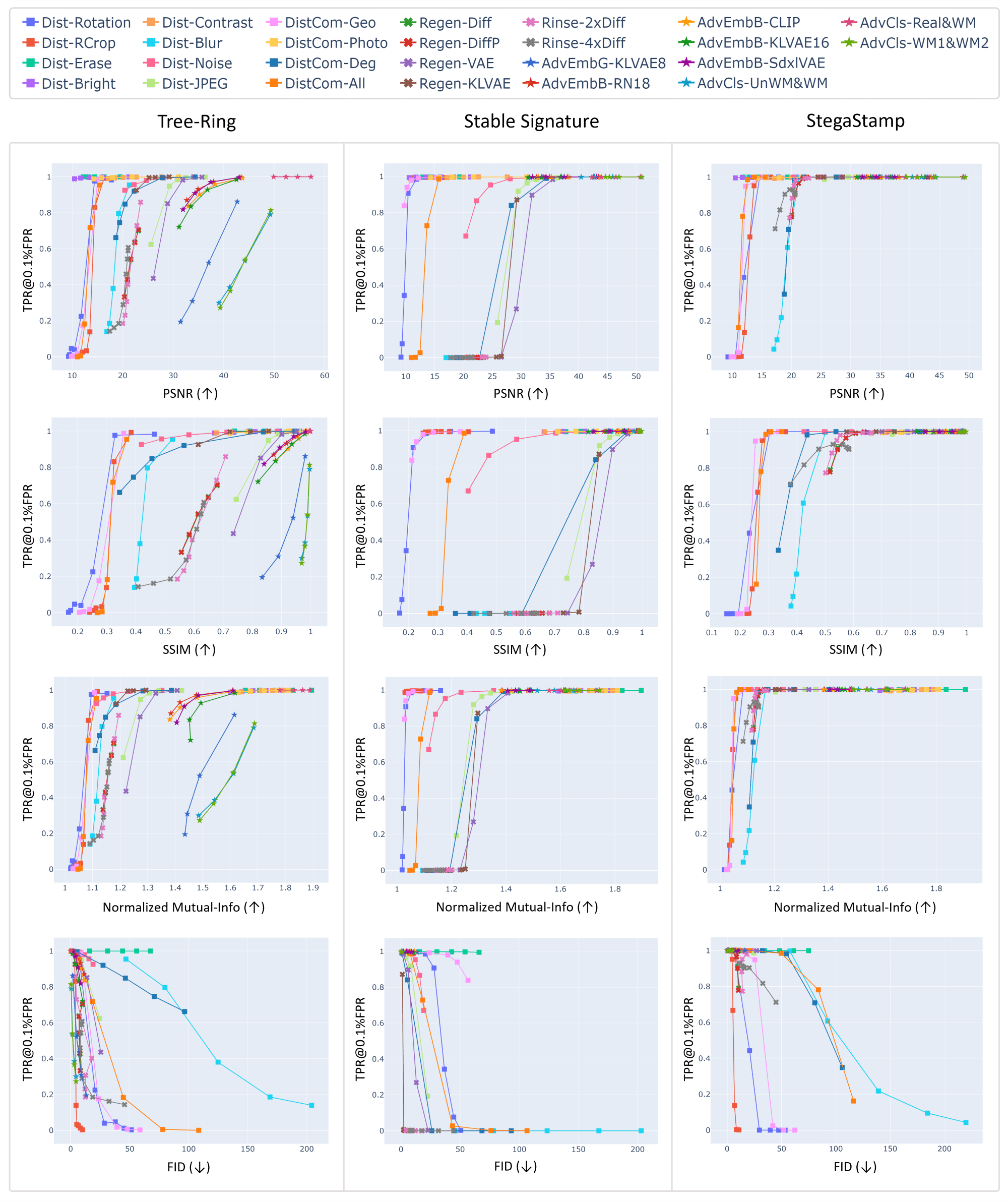}
    \caption{Evaluation on DALL$\cdot$E3 dataset under the detection setup (part 1).}
    \label{fig:all_dalle_1}
\end{figure}
\begin{figure*}[!htbp]
    \centering
    \includegraphics[width=\textwidth]{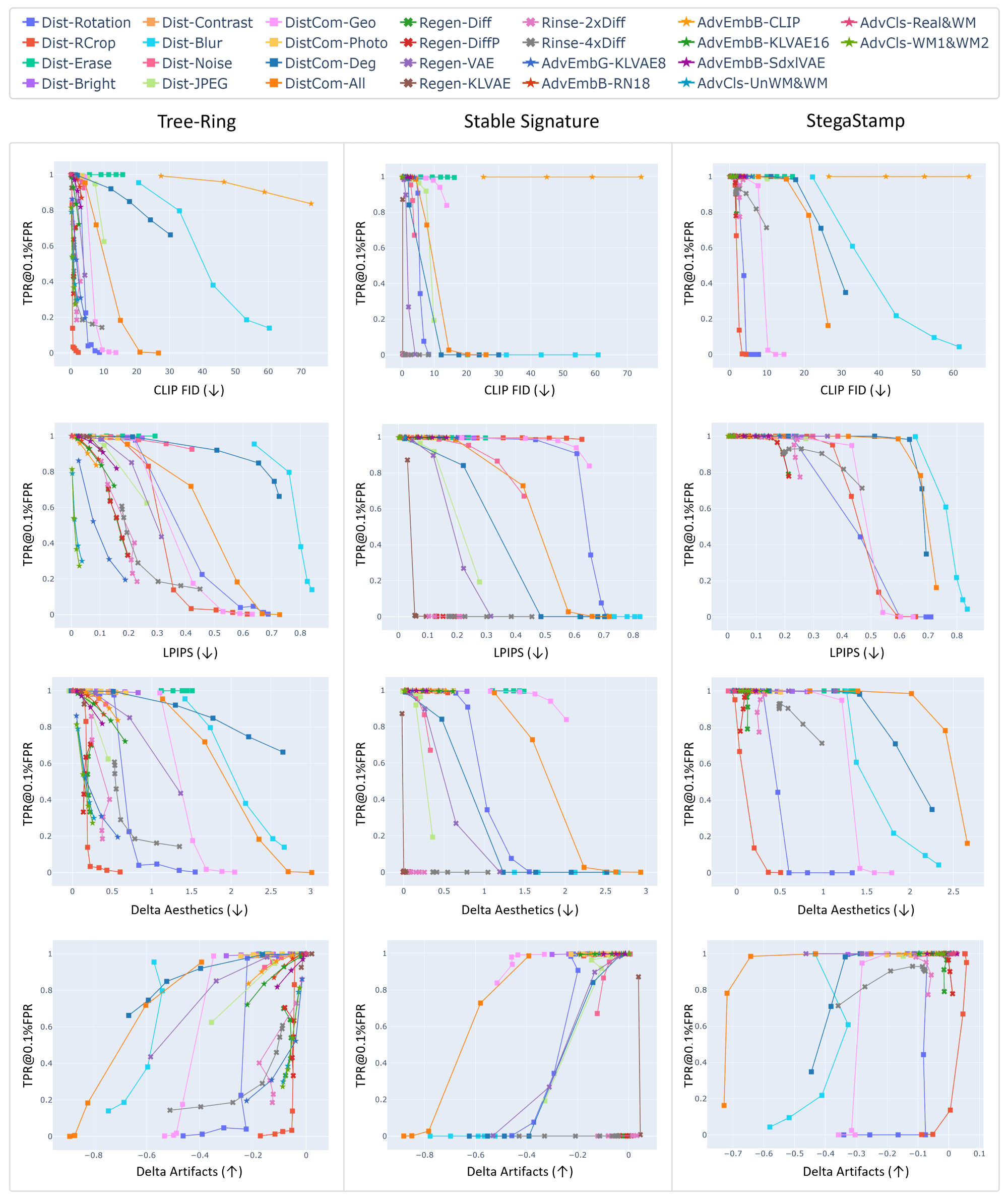}
    \caption{Evaluation on DALL$\cdot$E3 dataset under the detection setup (part 2).}
    \label{fig:all_dalle_2}
\end{figure*}

These findings have led to the development of our proposed AdvCls-WM1\&WM2 attack, which utilizes images watermarked with different messages (e.g., collected from two users, User1 and User2). 
The essential requirement for this approach is the surrogate model's ability to map images to the generator's latent space. 
This mapping allows the attacker to perturb the latent space, removing the watermark.  
In contrast to the AdvCls-UnWM\&WM approach, which uses both watermarked and non-watermarked images for training (differing only in the latent space), AdvCls-WM1\&WM2 uses two sets of images, each embedded with a distinct watermark message (differing only in the latent space as well). 
Figure~\ref{fig:visual_wm1_wm2} shows that AdvCls-WM1\&WM2 attack effectively disrupts the latent features of the images, including the watermarks. However, it lacks the precision to interchange the embedded watermark message. 
Consequently, while this attack can remove watermarks and mislead user identification—mistaking an image originally generated by User1 as belonging to another user—it cannot accurately manipulate the identification to frame User2 as desired by the attacker.  
The identification results in \cref{fig:adv_ident} also support this finding. 
Although AdvCls-WM1\&WM2 aims to misidentify images as belonging to User2, it often leads to misidentification as users other than User2.
However, in a system with fewer users, like 100 users, and under intense attack conditions (e.g., strength=8), AdvCls-WM1\&WM2 demonstrates a targeted identification success rate of 0.7\%, showing a potential direction for attacks aimed at targeted user identification.

\subsection{Visualization of Attacks}
In \cref{many-attacks}, we present visualizations of several attacks included in the \ours benchmark. Prefix indicates the attack strategy, while suffix indicates the strength.
\input{figures/visual_attack_comparison}

\subsection{Full Results on DiffusionDB, MS-COCO and \texorpdfstring{DALL$\cdot$E3}{DALL·E3}}

\subsection{Evaluation on Additional Watermarks: DWT-DCT and MBRS} \label{app:add_watermark}

To further demonstrate the utility and versatility of the WAVES benchmark, we evaluated two additional watermark methods: DWT-DCT \citep{al2007combined} and MBRS \citep{jia2021mbrs}. DWT-DCT combines Discrete Wavelet Transform (DWT) and Discrete Cosine Transform (DCT) for watermark embedding, while MBRS enhances the resilience of DNN-based watermarks to JPEG compression by incorporating real and simulated JPEG artifacts during training.

\begin{figure}[h]
    \centering
    \includegraphics[width=0.7\textwidth]{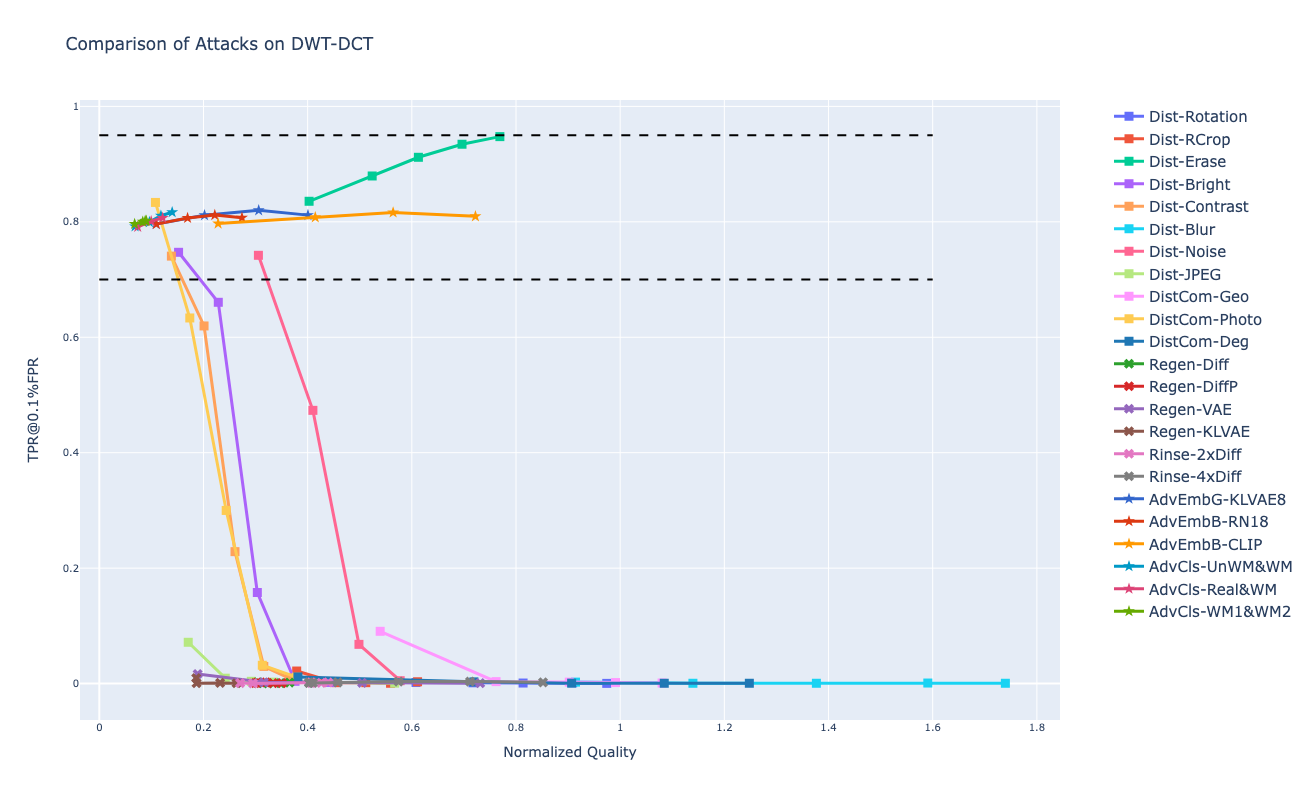}
    \caption{Stress test results for DWT-DCT. It is highly susceptible to regeneration attacks (cross markers) and most distortion attacks (square markers), but relatively robust against adversarial attacks.}
    \label{fig:dwtdct}
\end{figure}

\begin{figure}[h]
    \centering
    \includegraphics[width=0.7\textwidth]{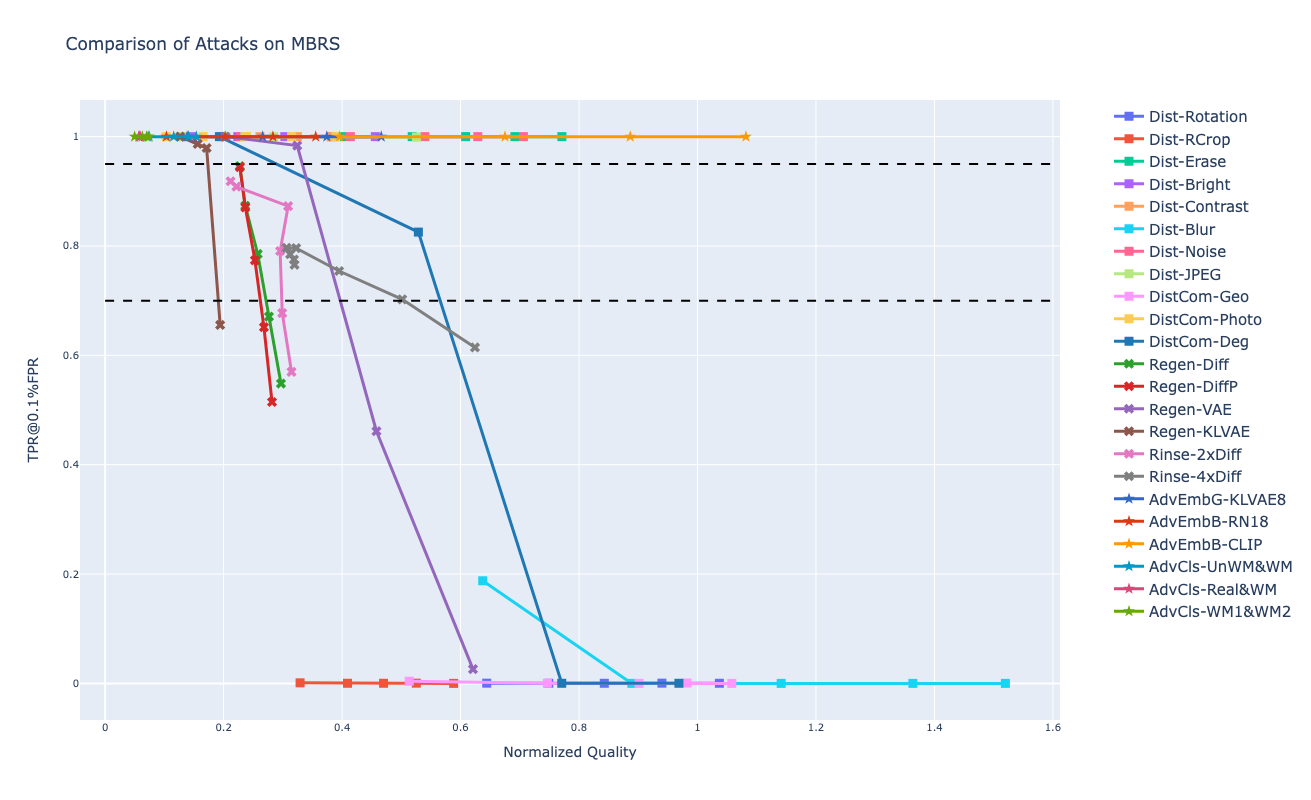}
    \caption{Stress test results for MBRS. It is vulnerable to certain distortion attacks (resized-cropping, blurring, rotation, combo distortions) and regeneration attacks, but robust against other distortions (JPEG compression, brightness/contrast, random erasing, noise) and adversarial attacks.}
    \label{fig:mbrs}
\end{figure}

Stress tests were conducted on these watermarks using all the attack methods in WAVES. Results are presented in Figures \ref{fig:dwtdct} and \ref{fig:mbrs} as performance vs. quality degradation 2D plots. Figure 7 in the main paper provides a comparison with the three existing watermarks (Tree-Ring, Stable Signature, and StegaStamp).

These findings confirm the utility of WAVES for identifying weaknesses in different watermark methods and demonstrate the ease of use and versatility of our benchmark toolkit, making it a valuable standard for the watermark research community.

\section{Limitations}
Although we have stress-tested five watermarks and 26 attacks, there could exist more watermarks and attacks that we did not include in this paper. However, we emphasize our framework is extensible to any watermarking method and attacks.
Additionally, our attack ranking method relies on author-selected TPR thresholds and image quality metrics that we believe will fairly capture attack potency based on existing literature and experimental studies. The use of other quality metrics (MSE, Watson-DFT, etc.) and differing TPR thresholds may affect attack rankings.

%% file: table/attacker_know.tex
\begin{table}[]
\caption{The knowledge of attackers}
\centering
\small
\begin{tabular}{cccccc}
\toprule
\multicolumn{2}{c}{\textbf{Attacks}}                                                                          & \textbf{\begin{tabular}[c]{@{}c@{}}Know Watermark\\ Algorithm\end{tabular}} & \textbf{\begin{tabular}[c]{@{}c@{}}Know Victim \\ Model\end{tabular}}                       & \textbf{Know Data}                                                                                                   & \textbf{Training} \\
\midrule
Distortion                                                                                   & All            & \xmark                                                                       & \xmark                                                                                       & \xmark                                                                                                           & \xmark             \\
\midrule
\multirow{3}{*}{Regeneration}                                                                & Regen-DiffP    & \xmark                                                                       & \xmark                                                                                       & user prompts                                                                                              & \xmark             \\ \cline{2-6}
                                                                                             & Regen-KLVAE    & \xmark                                                                       & \begin{tabular}[c]{@{}c@{}}VAE encoder (only\\ bottleneck size 8)\end{tabular} & \xmark                                                                                                           & \xmark             \\ \cline{2-6}
                                                                                             & Others         & \xmark                                                                       & \xmark                                                                                       & \xmark                                                                                                           & \xmark             \\
                                                                                             \midrule
\multirow{2}{*}{\begin{tabular}[c]{@{}c@{}}Adversarial\\ Embedding\end{tabular}}             & AdvEmbG-KLVAE8 & \xmark                                                                       & VAE encoder                                                                   & \xmark                                                                                                           & \xmark             \\ \cline{2-6}
                                                                                             & Others         & \xmark                                                                       & \xmark                                                                                       & \xmark                                                                                                           & \xmark             \\
                                                                                             \midrule
\multirow{3}{*}{\begin{tabular}[c]{@{}c@{}}Adversarial \\ Surrogate\\ Detector\end{tabular}} & AdvCLS-UnWM\&WM  & \xmark                                                                       & \xmark                                                                                       & \begin{tabular}[c]{@{}c@{}}watermarked and\\ non-watermarked images\\ from the victim model\end{tabular} & \cmark               \\ \cline{2-6}
                                                                                             & AdvCLS-Real\&WM  & \xmark                                                                       & \xmark                                                                                       & watermarked images                                                                                              & \cmark               \\ \cline{2-6}
                                                                                             & AdvCLS-WM1\&WM2  & \xmark                                                                       & \xmark                                                                                       & \begin{tabular}[c]{@{}c@{}}watermarked images \\ from two users\end{tabular}                                    & \cmark    \\     
                                                                                             \bottomrule
\end{tabular} \label{tab:attacker_know}
\end{table}

%% file: figures/regen_diffusion.tex
\begin{figure}[!htbp]
  \centering
  \subcaptionbox{Regen-Diff-40}{\includegraphics[width=0.22\linewidth]{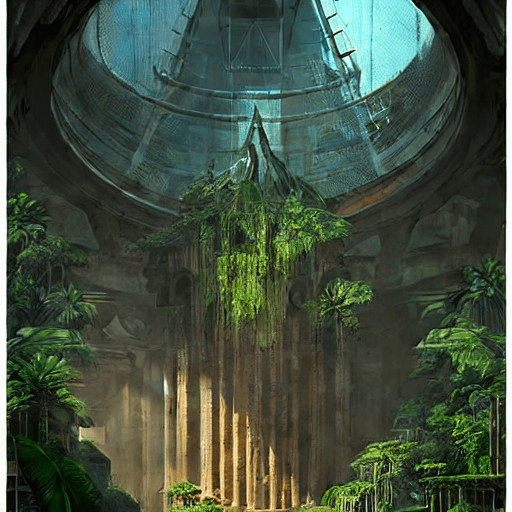}}
  \subcaptionbox{Regen-Diff-120}{\includegraphics[width=0.22\linewidth]{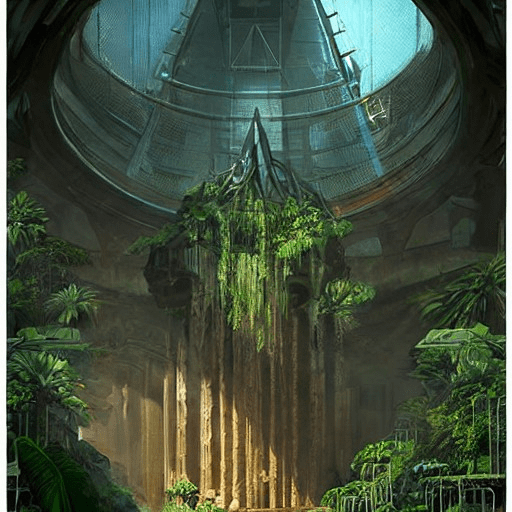}}
  \subcaptionbox{Regen-Diff-200}{\includegraphics[width=0.22\linewidth]{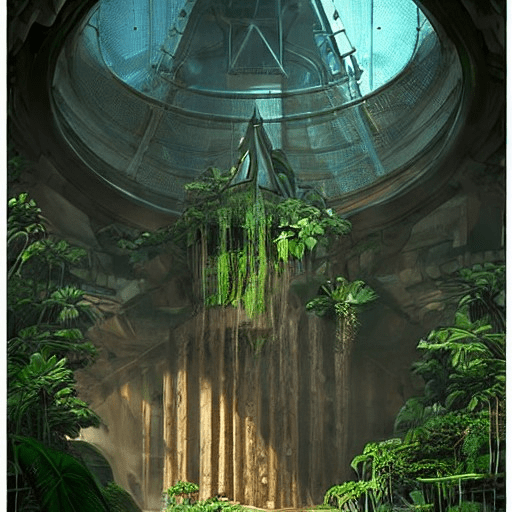}}
  \subcaptionbox{Regen-VAE-1}{\includegraphics[width=0.22\linewidth]{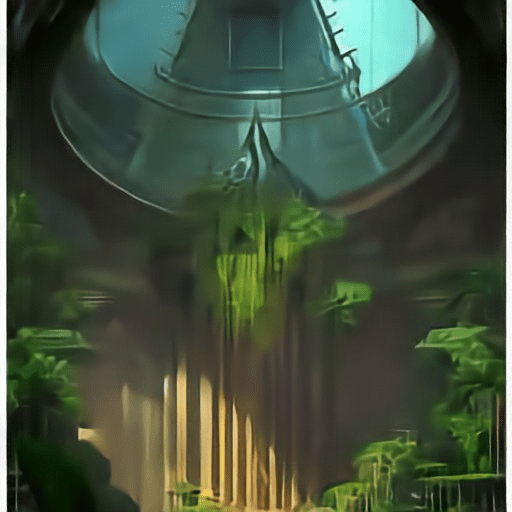}}
  \caption{Regenerative diffusion with varying depth of noising steps and a VAE regeneration with a low quality factor.}
  \label{fig:regen_diffusion_three-subfigures}
\end{figure}

%% file: figures/4x_regen_diffusion.tex
\begin{figure}[!htbp]
  \centering
  \subcaptionbox{Rinse-4xDiff-10}{\includegraphics[width=0.22\linewidth]{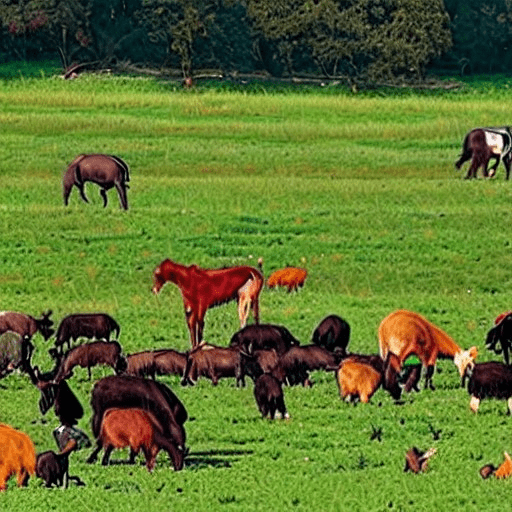}}
  \subcaptionbox{Regen-4xDiff-30}{\includegraphics[width=0.22\linewidth]{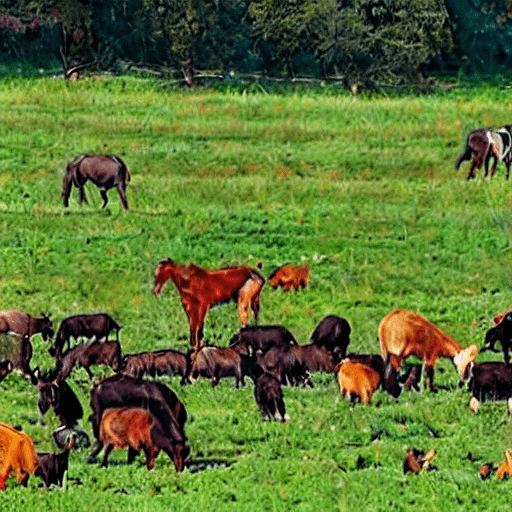}}
  \subcaptionbox{Regen-4xDiff-50}
  {\includegraphics[width=0.22\linewidth]{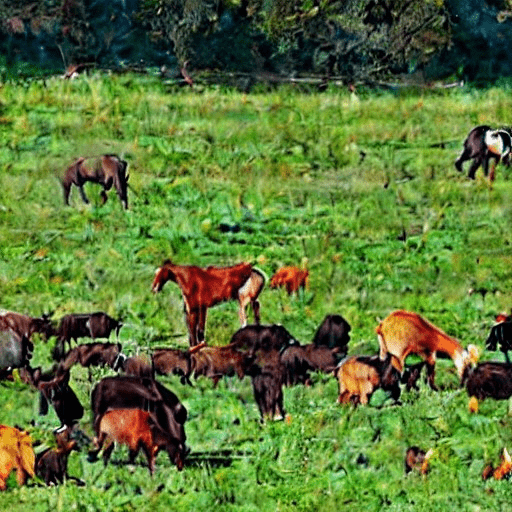}}
  \vspace{-0.5em}
  \caption{4x rinsing regeneration with varying depth of noising steps per diffusion.} \label{fig:4x_regen_diffusion_three-subfigures}
\end{figure}

%% file: table/leaderboard_ident.tex
\begin{table}[!htpb]
\caption{\textbf{Comparison of attacks across three watermarking methods under the identification setup (one million users).} Q denotes the normalized quality degradation and P denotes the performance as derived from aggregated 2D plots. Q@0.7P measures quality degradation at a 0.7 performance threshold where "inf" denotes cases where all tested attack strengths yield performance above 0.7, and "-inf" where all are below. Q@0.4P is defined analogously. Avg P and Avg Q are the average performance and quality over all the attack strengths. The lower the performance and the smaller the quality degradation, the stronger the attack. For each watermarking method, we rank attacks by Q@0.7P, Q@0.4P, Avg P, Avg Q, in that order, with lower values ($\downarrow$) indicating stronger attacks. The top 5 attack of each watermarking method are highlighted in red.}
\resizebox{\textwidth}{!}{%
\begin{tabular}{cccccccccccccccccc}
\toprule
\multirow{2}{*}{Attack}      & \multicolumn{5}{c}{Tree-Ring} &  & \multicolumn{5}{c}{Stable Signature} &  & \multicolumn{5}{c}{StegaStamp} \\ \cline{2-6} \cline{8-12} \cline{14-18} 
                             & Rank& Q@0.7P& Q@0.4P & Avg P & Avg Q &  & Rank& Q@0.7P& Q@0.4P  & Avg P   & Avg Q    &  & Rank& Q@0.7P& Q@0.4P  & Avg P  & Avg Q  \\
\midrule
Dist-Rotation   & 8  & -inf  & 0.434 & 0.131 & 0.648 &  & 12 & 0.613 & 0.642 & 0.400 & 0.650 &  & \textcolor{red}{4}  & 0.454 & 0.500 & 0.288 & 0.616 \\
Dist-RCrop      & 11 & -inf  & 0.592 & 0.094 & 0.463 &  & 24 & inf   & inf   & 0.972 & 0.461 &  & 6  & 0.602 & 0.602 & 0.494 & 0.451 \\
Dist-Erase      & 26 & inf   & inf   & 0.986 & 0.490 &  & 25 & inf   & inf   & 0.988 & 0.489 &  & 25 & inf   & inf   & 1.000 & 0.483 \\
Dist-Bright     & 22 & inf   & inf   & 0.913 & 0.304 &  & 23 & inf   & inf   & 0.982 & 0.305 &  & 22 & inf   & inf   & 0.995 & 0.317 \\
Dist-Contrast   & 23 & inf   & inf   & 0.949 & 0.243 &  & 20 & inf   & inf   & 0.979 & 0.243 &  & 17 & inf   & inf   & 0.994 & 0.231 \\
Dist-Blur       & 21 & 1.105 & 1.437 & 0.551 & 1.221 &  & \textcolor{red}{5}  & -inf  & -inf  & 0.000 & 1.204 &  & 9  & 0.897 & 0.970 & 0.280 & 1.198 \\
Dist-Noise      & 16 & 0.427 & inf   & 0.728 & 0.395 &  & 8  & 0.415 & 0.480 & 0.633 & 0.390 &  & 24 & inf   & inf   & 1.000 & 0.360 \\
Dist-JPEG       & 17 & 0.499 & 0.499 & 0.700 & 0.284 &  & 9  & 0.485 & 0.485 & 0.540 & 0.284 &  & 21 & inf   & inf   & 0.995 & 0.263 \\
DistCom-Geo     & 9  & -inf  & 0.559 & 0.105 & 0.768 &  & 13 & 0.788 & 0.835 & 0.519 & 0.767 &  & 7  & 0.676 & 0.717 & 0.359 & 0.733 \\
DistCom-Photo   & 23 & inf   & inf   & 0.947 & 0.242 &  & 20 & inf   & inf   & 0.981 & 0.243 &  & 17 & inf   & inf   & 0.994 & 0.239 \\
DistCom-Deg     & 18 & 0.556 & 0.864 & 0.570 & 0.694 &  & 7  & 0.216 & 0.281 & 0.183 & 0.679 &  & 8  & 0.870 & 0.957 & 0.737 & 0.664 \\
DistCom-All     & 10 & -inf  & 0.575 & 0.123 & 0.908 &  & 11 & 0.550 & 0.623 & 0.176 & 0.900 &  & 10 & 0.995 & 1.096 & 0.682 & 0.870 \\ \midrule
Regen-Diff      & 6  & -inf  & 0.307 & 0.258 & 0.323 &  & \textcolor{red}{1}  & -inf  & -inf  & 0.000 & 0.300 &  & \textcolor{red}{2}  & 0.333 & inf   & 0.766 & 0.327 \\
Regen-DiffP     & 6  & -inf  & 0.308 & 0.256 & 0.327 &  & \textcolor{red}{1}  & -inf  & -inf  & 0.000 & 0.303 &  & \textcolor{red}{1}  & 0.336 & 0.356 & 0.763 & 0.329 \\
Regen-VAE       & 19 & 0.578 & 0.578 & 0.701 & 0.348 &  & 10 & 0.545 & 0.545 & 0.340 & 0.339 &  & 23 & inf   & inf   & 1.000 & 0.343 \\
Regen-KLVAE     & 14 & 0.257 & inf   & 0.810 & 0.233 &  & 6  & -inf  & -inf  & 0.047 & 0.206 &  & 17 & inf   & inf   & 0.999 & 0.240 \\
Rinse-2xDiff    & \textcolor{red}{5}  & -inf  & 0.270 & 0.220 & 0.357 &  & \textcolor{red}{3}  & -inf  & -inf  & 0.000 & 0.332 &  & \textcolor{red}{3}  & 0.390 & 0.402 & 0.778 & 0.366 \\
Rinse-4xDiff    & \textcolor{red}{1}  & -inf  & -inf  & 0.110 & 0.466 &  & \textcolor{red}{4}  & -inf  & -inf  & 0.000 & 0.438 &  & \textcolor{red}{5}  & 0.488 & 0.676 & 0.687 & 0.477 \\ \midrule
AdvEmbG-KLVAE8  & \textcolor{red}{4}  & -inf  & 0.168 & 0.259 & 0.253 &  & 20 & inf   & inf   & 0.985 & 0.249 &  & 17 & inf   & inf   & 1.000 & 0.232 \\
AdvEmbB-RN18    & 15 & 0.288 & inf   & 0.811 & 0.218 &  & 17 & inf   & inf   & 0.990 & 0.212 &  & 14 & inf   & inf   & 1.000 & 0.196 \\
AdvEmbB-CLIP    & 20 & 0.697 & inf   & 0.798 & 0.549 &  & 26 & inf   & inf   & 0.991 & 0.541 &  & 25 & inf   & inf   & 1.000 & 0.488 \\
AdvEmbB-KLVAE16 & 12 & 0.158 & 0.309 & 0.540 & 0.238 &  & 19 & inf   & inf   & 0.983 & 0.233 &  & 14 & inf   & inf   & 1.000 & 0.206 \\
AdvEmbB-SdxlVAE & 13 & 0.214 & inf   & 0.692 & 0.221 &  & 17 & inf   & inf   & 0.986 & 0.219 &  & 14 & inf   & inf   & 1.000 & 0.204 \\
AdvCls-UnWM\&WM & \textcolor{red}{2}  & -inf  & 0.123 & 0.352 & 0.145 &  & 14 & inf   & inf   & 0.991 & 0.101 &  & 11 & inf   & inf   & 1.000 & 0.101 \\
AdvCls-Real\&WM & 25 & inf   & inf   & 0.986 & 0.047 &  & 14 & inf   & inf   & 0.990 & 0.092 &  & 11 & inf   & inf   & 1.000 & 0.106 \\
AdvCls-WM1\&WM2 & \textcolor{red}{2}  & -inf  & 0.118 & 0.343 & 0.139 &  & 14 & inf   & inf   & 0.991 & 0.084 &  & 13 & inf   & inf   & 1.000 & 0.129 \\ 
\bottomrule
\end{tabular}\label{tab:leaderboard_ident}
}
\end{table}

%% file: figures/visual_attack_comparison.tex
\begin{figure*}[!htbp]
\centering
  \begin{subfigure}{0.24\textwidth}
    \includegraphics[width=\linewidth]{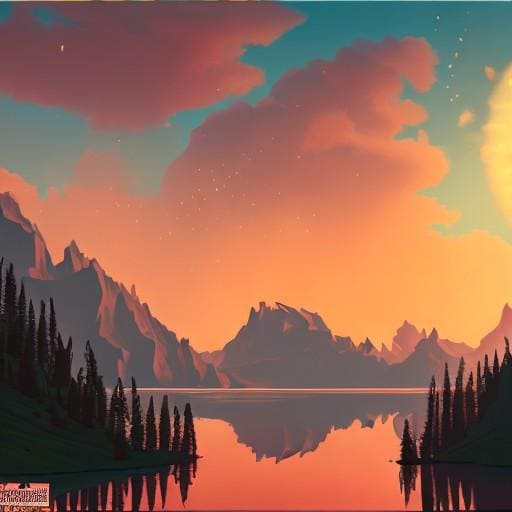}
    \caption{Tree-Ring Unattacked}
  \end{subfigure}
\hfill
  \begin{subfigure}{0.24\textwidth}
    \includegraphics[width=\linewidth]{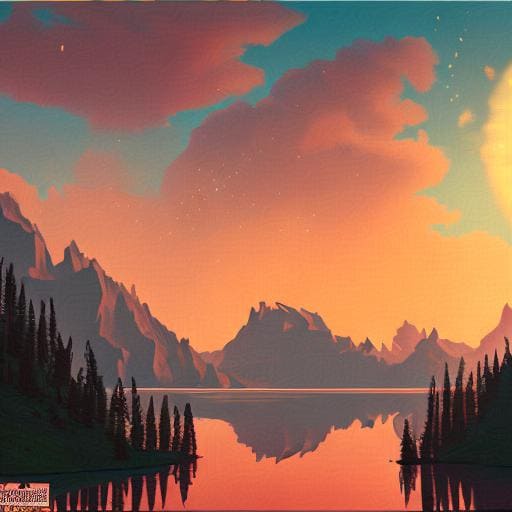}
    \caption{{\small AdvEmbG-KLVAE8-2/255}}
  \end{subfigure}
  \hfill
  \begin{subfigure}{0.24\textwidth}
    \includegraphics[width=\linewidth]{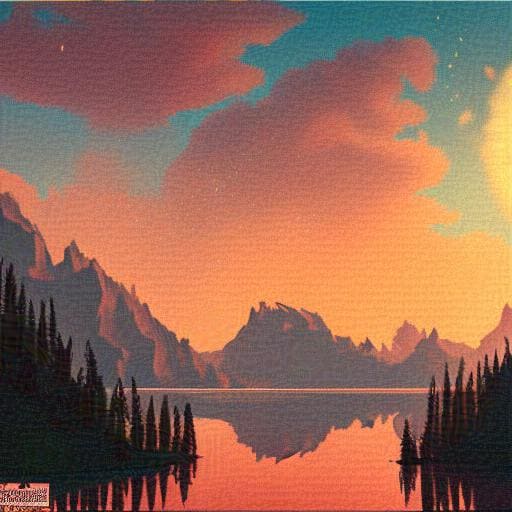}
    \caption{{\small AdvEmbG-KLVAE8-8/255}}
  \end{subfigure}
    \hfill
  \begin{subfigure}{0.24\textwidth}
    \includegraphics[width=\linewidth]{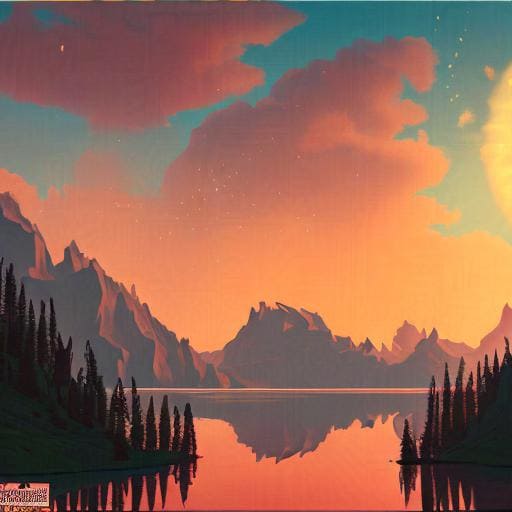}
    \caption{{\small AdvEmbB-CLIP-2/255}}
  \end{subfigure}%
  \hfill
  \begin{subfigure}{0.24\textwidth}
    \includegraphics[width=\linewidth]{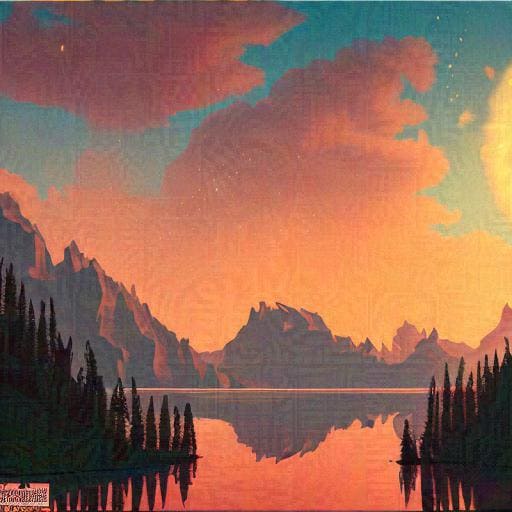}
    \caption{{\small AdvEmbB-CLIP-8/255}}
  \end{subfigure}
    \hfill
  \begin{subfigure}{0.24\textwidth}
    \includegraphics[width=\linewidth]{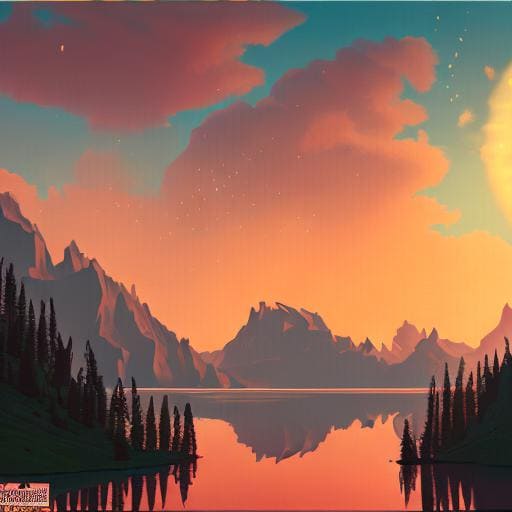}
    \caption{{\small AdvClsWM1WM2-2/255}}
  \end{subfigure}%
  \hfill
  \begin{subfigure}{0.24\textwidth}
    \includegraphics[width=\linewidth]{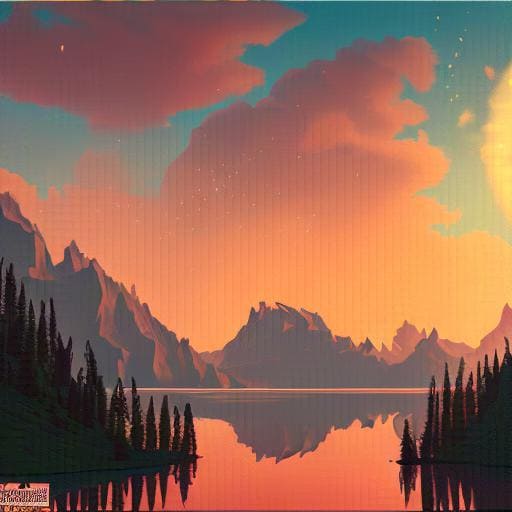}
    \caption{{\small AdvClsWM1WM28/255}}
  \end{subfigure}
 \hfill
  \begin{subfigure}{0.24\textwidth}
    \includegraphics[width=\linewidth]{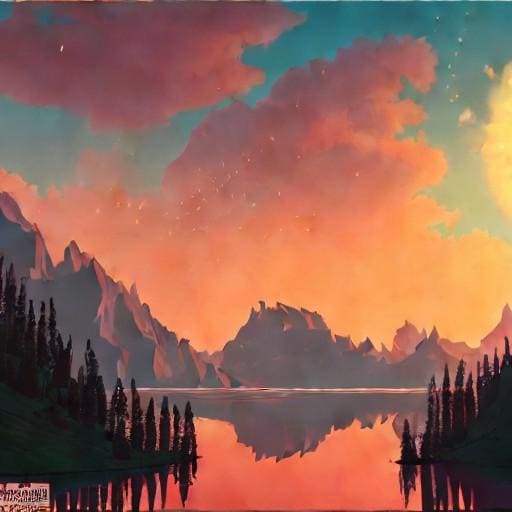}
    \caption{Regen-Diff-40}
  \end{subfigure}%
  \hfill
  \begin{subfigure}{0.24\textwidth}
    \includegraphics[width=\linewidth]{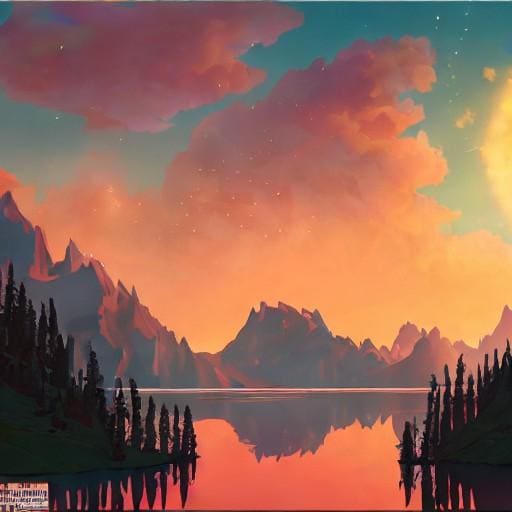}
    \caption{Regen-Diff-200}
  \end{subfigure}
\hfill
  \begin{subfigure}{0.24\textwidth}
    \includegraphics[width=\linewidth]{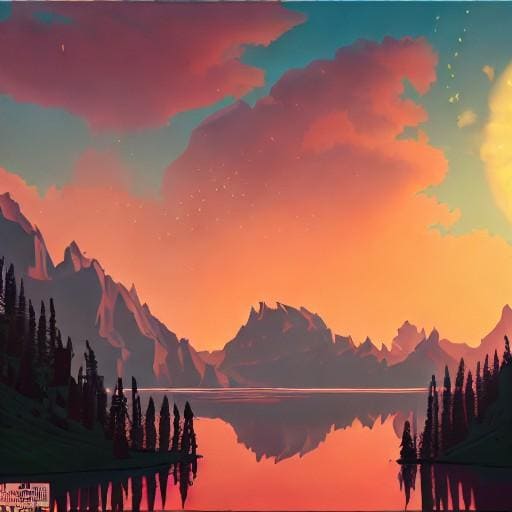}
    \caption{Rinse-2xDiff-20}
  \end{subfigure}%
  \hfill
  \begin{subfigure}{0.24\textwidth}
    \includegraphics[width=\linewidth]{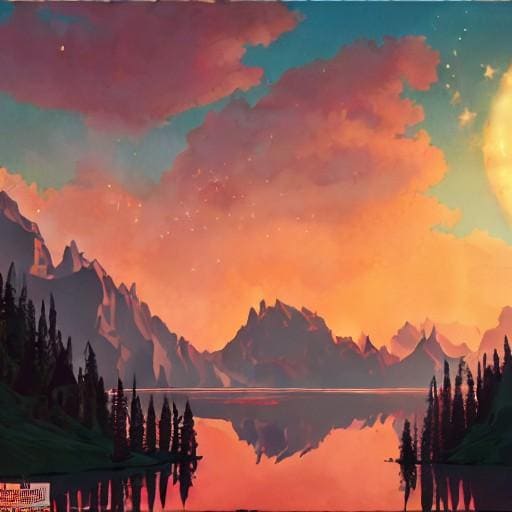}
    \caption{Rinse-2xDiff-100}
  \end{subfigure}
\hfill
  \begin{subfigure}{0.24\textwidth}
    \includegraphics[width=\linewidth]{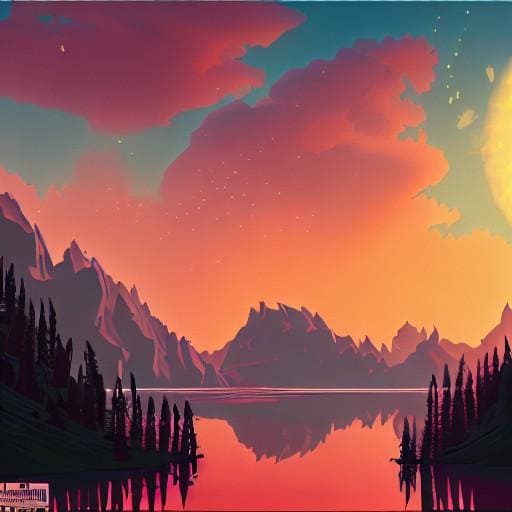}
    \caption{Rinse-4xDiff-10}
  \end{subfigure}%
  \hfill
  \begin{subfigure}{0.24\textwidth}
    \includegraphics[width=\linewidth]{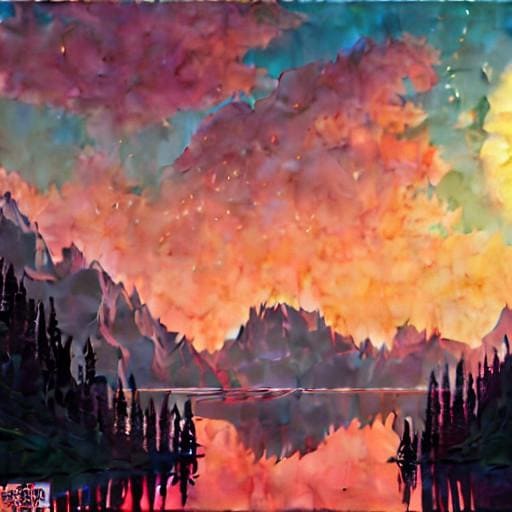}
    \caption{Rinse-4xDiff-50}
  \end{subfigure}
  \hfill
  \begin{subfigure}{0.24\textwidth}
    \includegraphics[width=\linewidth]{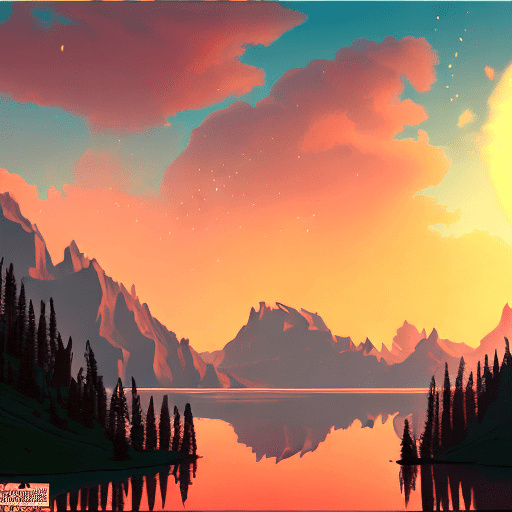}
    \caption{DistCom-Photo-0.15}
  \end{subfigure}%
  \hfill
  \begin{subfigure}{0.24\textwidth}
    \includegraphics[width=\linewidth]{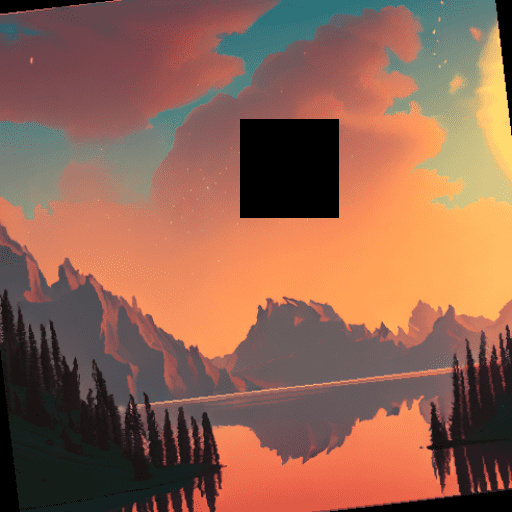}
    \caption{DistCom-Geo-0.15}
  \end{subfigure}
  \hfill
  \begin{subfigure}{0.24\textwidth}
  \label{source}
    \includegraphics[width=\linewidth]{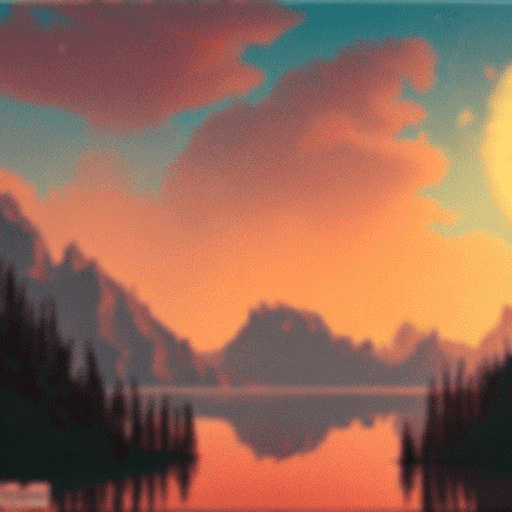}
    \caption{DistCom-Deg-0.15}
  \end{subfigure}
\caption{A visual demonstration of various adversarial, regeneration, and distortion attacks on a Tree-Ring watermarked image. \textbf{Figure (a)} is the base unattacked image. The base prompt, drawn from DiffusionDB, is ``digital painting of a lake at sunset surrounded by forests and mountains,'' along with further styling details.}
\label{many-attacks}
\end{figure*}